%% file: main.tex
\documentclass[10pt, letterpaper, journal, twoside]{IEEEtran}
\IEEEoverridecommandlockouts

\usepackage{cite}
\usepackage{mathtools,amssymb,amsfonts}
\usepackage{graphicx}
\usepackage[table]{xcolor}
\usepackage{booktabs}
\usepackage{siunitx}
\usepackage{float}
\usepackage{stfloats} 
\usepackage{algorithm}
\usepackage{algpseudocode}
\usepackage{tikz}
\usepackage{pgfplots}
\pgfplotsset{compat=1.16}
\usepackage[normalem]{ulem} 
\usepackage{comment} 
\usepackage{cancel} 
\usepackage{needspace} 
\usetikzlibrary{arrows.meta,positioning,calc,fit,shapes.geometric,shapes.misc,decorations.pathmorphing}
\usepgfplotslibrary{groupplots}

\def\BibTeX{{\rm B\kern-.05em{\sc i\kern-.025em b}\kern-.08em
    T\kern-.1667em\lower.7ex\hbox{E}\kern-.125emX}}

\usepackage{hyperref}
\hypersetup{
    colorlinks=true,
    linkcolor=blue,
    filecolor=magenta,      
    urlcolor=blue,
}

\usepackage[top=0.75in, bottom=0.9in, left=0.59in, right=0.59in]{geometry}

\newtheorem{remark}{Remark} 
\newcommand{\Ebar}{\overline{\mkern-3.5mu E\mkern-1.mu}}

\newcommand{\mapname}[1]{\texttt{#1}}

\usepackage[font=small, skip=1pt]{caption}
\usepackage[skip=-1pt]{subcaption} 

\def\subparagraph{} 
\usepackage{titlesec}

\titleformat{\subsubsection}[runin]{\itshape}{\arabic{subsubsection})}{0.5em}{}[:]

\titlespacing*{\section}
{0pt}{4pt plus 3pt}{4pt plus 1pt} 
\titlespacing*{\subsection}
{0pt}{4pt plus 3pt}{4pt plus 1pt} 
\titlespacing*{\subsubsection}
{8pt}{2pt plus 2pt}{2pt} 
\titleformat{\paragraph}[runin]{\itshape}{\alph{paragraph})}{0.5em}{}[:]
\titlespacing*{\paragraph}
{8pt}{2pt plus 2pt}{2pt} 
\newif\ifhideequation
\hideequationfalse
\ifhideequation
  \excludecomment{eqblock}
\else
  \includecomment{eqblock}
\fi

\newif\ifanonymous
\anonymousfalse

\begin{document}

\title{Search-Informed MPPI for Kinodynamic Motion Planning via Monte Carlo Tree Expansion}

\ifanonymous
\author{\smash[b]{Anonymous Authors}%
}
\else
\author{Ihab S. Mohamed and Lantao Liu%
\thanks{Authors are with the Luddy School of Informatics, Computing, and Engineering, Indiana University, Bloomington, IN 47408, USA (e-mail: {\tt\small \{mohamedi \& lantao\}@iu.edu}).}}
\fi

\maketitle

\begin{abstract}
    
    Kinodynamic motion planning in large-scale, cluttered environments remains challenging, as dynamically feasible, collision-free trajectories must be computed efficiently over long horizons.
    Monte Carlo Tree Search (MCTS) can strategically explore long-horizon alternatives but faces large branching factors in continuous control spaces and often relies on random default-policy rollouts, whereas Model Predictive Path Integral (MPPI) efficiently optimizes local trajectories yet remains susceptible to obstacle-induced local minima.
    To overcome these challenges, we propose \textit{Search-Informed MPPI (SI-MPPI)}, a unified framework that embeds MPPI trajectory optimization within MCTS-based tree search. The core concept is to fuse the expansion and default-policy simulation stages of MCTS into a single MPPI expansion stage guided by structured intermediate targets, termed \emph{baits}. 
    During each expansion, an adaptive bait-handling rule retains geometrically valid baits as local targets for GPU-batched MPPI optimizations that generate dynamically feasible candidate motion segments.
    Feasibility gating then checks these candidates for collision safety and sufficient progress, and top-$k$ selection admits the lowest-cost accepted candidates as tree edges, yielding sparse, directed tree growth; thereafter, effective sample size (ESS)-guided control adaptation warm-starts subsequent tree expansions, while cost-based backpropagation updates tree-search statistics.
    By conducting extensive simulations across four large-scale, cluttered 2D occupancy-grid maps, we demonstrate that SI-MPPI consistently produces the smallest search trees and the smoothest angular control profiles compared with three kinodynamic baselines, while maintaining competitive planning times and path lengths. Ablation results further highlight the importance of GPU batching, adaptive bait handling, and retaining multiple expansion candidates for planning efficiency and robustness.
    \ifanonymous\else
    Videos and additional results are available on the project page: \url{https://ihabmohamed.github.io/SI-MPPI/}.
    \fi
\end{abstract}

{\renewcommand{\abstractname}{Note to Practitioners}
\begin{abstract}
    Autonomous mobile robots in warehouses, factories, hospitals, and greenhouses must plan long routes through cluttered layouts while respecting how they can physically move. 
    Widely used planners involve practical trade-offs: sampling-based planners such as the asymptotically optimal Rapidly-exploring Random Tree (RRT$^{*}$) and its variants may spend considerable computation exploring regions irrelevant to the task, grid-based planners depend on map resolution, and lattice-based planners often rely on predefined motion primitives. This paper presents SI-MPPI, a planner that uses tree search to decide which routes are worth exploring and MPPI to optimize many short, dynamically feasible motion candidates in parallel on a laptop-grade GPU, retaining low-cost candidates that satisfy collision-safety and progress checks. The planner operates directly on standard 2D occupancy-grid maps, using the robot's motion model for trajectory generation and its footprint for collision checking. In simulations on four $\SI{100}{\metre}\times\SI{100}{\metre}$ cluttered maps, SI-MPPI consistently produced the smoothest steering commands under a common parameter configuration, potentially reducing actuator wear and payload disturbance. Averaged across these maps, it built search trees roughly $7$--$15\times$ smaller than those of the evaluated baselines and planned about $10$--$27\times$ faster than the RRT$^{*}$-based planners, while maintaining comparable or shorter path lengths. Practitioners should note that the proposed framework currently targets static, known maps, does not guarantee the shortest route, and produces planned trajectories with relatively low speeds; for higher-speed execution, however, a separate receding-horizon MPPI controller can track the resulting waypoints, as illustrated in the supplementary video. Future work will extend the approach to robots with more complex dynamics and to speed-aware planning.%
\end{abstract}}


\section{Introduction}
\label{sec:introduction}

\IEEEPARstart{A}{utonomous} mobile robots (AMRs) are increasingly transitioning from controlled laboratory settings to large-scale, highly structured, and cluttered environments, such as manufacturing floors, warehouses, complex hospital layouts, smart farms, and greenhouses \cite{fragapane2021planning, yepez2023mobile, keith2024review, da2021robotic}, where they must navigate efficiently while avoiding obstacles and respecting their dynamic constraints. At the heart of this autonomy lies motion planning: the problem of computing a trajectory or path from one position to another while balancing reliability, computational efficiency, and trajectory quality \cite{gasparetto2015path, gonzalez2026fast}. 
In these real-world scenarios, generating safe, dynamically feasible trajectories in real time is essential for reliable operation. Meeting this requirement through kinodynamic motion planning—wherein the robot must satisfy non-convex collision constraints alongside nonlinear dynamics—remains particularly challenging in complex, cluttered environments, as existing planners often struggle to efficiently couple high-level discrete decisions, such as selecting a narrow passage, with low-level continuous control under actuation limits~\cite{gammell2021asymptotically, sotirchos2024search}.

Conventional sampling-based algorithms, most notably Rapidly-exploring Random Trees (RRTs), have demonstrated remarkable success in geometric path planning by offering probabilistic completeness guarantees~\cite{lavalle1998rapidly}.
However, extending these methods to kinodynamic settings introduces substantial computational bottlenecks: generating dynamically feasible local connections typically requires either sequentially propagating control inputs through numerical integration~\cite{lavalle2001randomized, li2016asymptotically, butcher2016numerical}, solving a challenging two-point boundary value problem (BVP) to connect exact states~\cite{donald1993kinodynamic}, linearizing the system dynamics around local operating points~\cite{perez2012lqr, webb2013kinodynamic}, approximating steering or local path generation using learning-based methods~\cite{qureshi2019motion, li2021mpc}, or utilizing flat-output representations to derive algebraic trajectory rollouts~\cite{bascetta2017flat, duong2026ultrafast}.
Yet, despite these alternatives, the underlying randomized sampling strategies can still expend substantial computation on regions irrelevant to the global task. This untargeted exploration can slow convergence in cluttered environments containing narrow passages~\cite{gammell2014informed, wang2026asd}.
Consequently, although recent advances in learned, guided, biased, and bidirectional sampling improve exploration efficiency~\cite{wang2020neural, hu2026path, chiang2019rl, kyaw2024greedy}, the computational cost of randomized global exploration and repeated kinodynamic edge generation can still hinder real-time performance in large-scale, obstacle-dense scenarios~\cite{allen2019real, shome2021asymptotically, yang2025lqr}.

Search-based approaches offer an alternative by structuring motion planning as a search over long-horizon decision sequences and can be broadly divided into graph-search methods~\cite{lavalle2006planning} and stochastic tree-search methods such as Monte Carlo Tree Search (MCTS)~\cite{browne2012survey}. 
Within the graph-search branch, classical planners such as Dijkstra~\cite{dijkstra1959note} and A$^*$~\cite{hart1968formal}, their anytime and incremental/replanning variants~\cite{koenig2002d, likhachev2003ara, koenig2004lifelong, likhachev2005anytime}, and kinodynamic planners based on state lattices or motion primitives~\cite{pivtoraiko2009differentially, likhachev2009planning, liu2017search} reason over discrete state--action representations and can provide guarantees with respect to the chosen graph, lattice, or primitive set.
Their performance, however, depends strongly on graph resolution, lattice discretization, and primitive-set design; as the state dimension or planning horizon grows, the number of discrete states and admissible actions can increase rapidly, substantially increasing the computational burden of graph-based planning~\cite{sotirchos2024search}.
To mitigate this burden, recent methods have explored two complementary strategies: compact geometric representations of collision-free space~\cite{deits2015computing, marcucci2023motion, gonzalez2026fast} and hybrid search--optimization planners that combine discrete search with continuous trajectory refinement~\cite{natarajan2021interleaving, natarajan2024pinsat, ortiz2024idb}.
Yet such methods often rely on environment-specific structure, precomputed geometric or convex decompositions, simplified dynamics, analytical steering, or repeated trajectory optimization and repair, thereby limiting their direct applicability to large-scale, cluttered occupancy-grid environments, where planners must generate continuous, dynamically feasible motion segments under nonlinear dynamics.

Within the stochastic tree-search branch, MCTS offers a complementary mechanism for long-horizon planning by using tree-search statistics to adaptively allocate computation among promising decisions~\cite{kocsis2006bandit, browne2012survey}. 
Despite its success in large discrete decision-making domains such as Go and general game playing, its direct extension to continuous kinodynamic planning remains challenging: continuous control spaces induce large effective branching factors and rapid tree growth~\cite{riviere2024monte, kujanpaa2024continuous}; stochastic rollouts introduce variance into finite-sample value estimates~\cite{veness2011variance}; and unguided tree expansion can be inefficient for long-horizon planning and collision-aware robotic navigation~\cite{riviere2021neural, schramm2024volume, bonanni2025monte}.
Recent robotics work has addressed aspects of these challenges through specialized tree representations~\cite{xue2025monte}, task-specific formulations for constrained and multi-objective planning~\cite{xu2025contact, chen2019pareto}, optimization-coupled MCTS for manipulation and locomotion~\cite{zhu2023efficient, amatucci2022monte, dhedin2025simultaneous}, and continuous-space search mechanisms~\cite{riviere2024monte, kujanpaa2024continuous}.
In a separate line of work, Model Predictive Path Integral (MPPI) control has emerged as an effective sampling-based trajectory optimizer for nonlinear systems, leveraging massively parallel rollouts to optimize control sequences in real time~\cite{williams2017model, williams2018information}. 
MPPI naturally accommodates nonlinear dynamics, state-dependent costs, and control limits, making it attractive for real-time local trajectory optimization in agile robot navigation. 
Yet MPPI remains a local finite-horizon optimizer; without an informative nominal control sequence or reference trajectory, it can become trapped in obstacle-induced local minima or fail to make consistent global progress in cluttered, non-convex environments~\cite{mohamed2023gpmppi, tao2023rrt}.

Motivated by these complementary capabilities and limitations, we propose \textit{Search-Informed MPPI (SI-MPPI)}, a kinodynamic motion planning framework that tightly integrates MCTS-based tree search with GPU-batched MPPI optimization---an integration that, to the best of the authors' knowledge, has not previously been reported for collision-aware kinodynamic navigation in cluttered environments. The key insight is to replace the conventional expansion and default-policy simulation stages of MCTS with a unified MPPI expansion stage guided by structured intermediate targets, termed \textit{baits} (see Fig.~\ref{fig:mcts_si-mppi_diagram}). Within SI-MPPI, MCTS and MPPI serve complementary functions: MCTS adaptively allocates computation among competing long-horizon alternatives, whereas MPPI generates cost-informed local motion segments under the robot's nonlinear dynamics. 
The primary contributions of this work are as follows:
\begin{enumerate}
    \item \textit{Unified search--optimization framework:}
    Rather than relying on exact BVP steering, predefined motion primitives, or purely random control propagation, our proposed framework constructs an asymmetric search tree from admitted MPPI-optimized motion segments, thereby coupling long-horizon global search with local nonlinear trajectory optimization.

    \item \textit{Bait-guided MPPI expansion:}
    We develop a GPU-batched expansion mechanism that optimizes finite-horizon control sequences toward multiple retained baits in parallel and incorporates adaptive bait handling, feasibility gating, and top-$k$ selection, enabling computationally efficient, sparse, and directed tree growth in continuous state--action spaces.
    
    \item \textit{ESS-guided adaptation and backpropagation:}
    We leverage the effective sample size (ESS) to adapt the MPPI nominal control mean, providing informed warm starts for subsequent expansions, while cost-based backpropagation updates the tree statistics to guide global search.

    \item \textit{Comparative and ablation studies:}
    We evaluate our proposed framework through comparative simulations against three representative sampling- and search-based kinodynamic planners, complemented by an ablation study, with performance assessed in terms of mission success, search complexity, and trajectory quality in complex, large-scale, cluttered environments.
\end{enumerate}

The remainder of this paper is organized as follows. Section~\ref{sec:related_work} reviews related work, followed by the formal kinodynamic motion-planning problem formulation in Section~\ref{sec:problem} and an overview of MPPI and MCTS in Section~\ref{sec:preliminaries}. Building on these foundations, Section~\ref{sec:si-mppi} presents the proposed SI-MPPI framework, detailing Upper Confidence Bound (UCB)-guided tree selection, bait-guided GPU-batched MPPI expansion, candidate admission, and ESS-guided adaptation with cost-based backpropagation. Section~\ref{sec:results} then presents the comparative offline planning evaluation, ablation study, and discussion, followed by the conclusions and future research directions in Section~\ref{sec:conclusion}.

\section{Related Work}
\label{sec:related_work}
To contextualize our proposed planning framework, this section reviews four closely related paradigms for kinodynamic motion planning: sampling-based methods, graph-search and hybrid methods, MCTS-based tree search, and MPPI-based trajectory optimization.

\subsection{Sampling-Based Kinodynamic Motion Planning}

Sampling-based motion planning methods---most notably Probabilistic Roadmaps (PRM)~\cite{kavraki2002probabilistic}, Rapidly-exploring Random Trees (RRTs)~\cite{lavalle1998rapidly}, and asymptotically optimal variants such as RRT$^*$~\cite{karaman2011sampling}---approximate continuous configuration or state spaces through randomly generated samples and local connections, making them foundational tools for high-dimensional planning due to their probabilistic completeness and avoidance of explicit state-space discretization. In kinodynamic settings, however, the central bottleneck shifts from deciding \textit{where} to sample to determining \textit{how} to connect sampled states with dynamically feasible edges; this local connection procedure is commonly referred to as the edge-extension or steering operator.
Early kinodynamic planners addressed this challenge through random control propagation, where control inputs are sampled and the system dynamics are integrated forward over a short horizon~\cite{lavalle2001randomized}.
Later methods such as Stable Sparse-RRT (SST) and SST$^*$ retain random forward propagation while maintaining a sparse set of representative states~\cite{li2016asymptotically}.
Although random propagation avoids solving a BVP, it provides no direct control over the terminal state and may require many tree expansions to make directed progress, which can yield irregular trajectories and inefficient exploration in cluttered environments.
Conversely, exact point-to-point steering based on closed-form analytical solutions---such as Dubins curves for forward-only car-like motion~\cite{dubins1957curves} or Reeds--Shepp curves for forward-and-backward car-like motion~\cite{reeds1990optimal}---can produce highly directed connections between prescribed poses.
However, these closed-form steering functions are limited to specific vehicle kinematics, while general nonlinear systems typically require numerical BVP solvers or approximate steering methods~\cite{donald1993kinodynamic, xie2015toward}.

To handle complex nonlinear dynamics without relying on exact analytical steering or repeated numerical BVP solutions, the literature has explored several approximate steering strategies.
LQR-RRT$^*$ approximates locally optimal steering by linearizing the nonlinear dynamics around local operating points and applying a Linear Quadratic Regulator (LQR)~\cite{perez2012lqr}, whereas Kinodynamic RRT$^*$ provides exact optimal connections for systems with controllable linear dynamics~\cite{webb2013kinodynamic}.
Building upon LQR-based edge extension, LQR-CBF-RRT$^*$ incorporates Control Barrier Functions (CBFs) to enforce safety constraints~\cite{yang2025lqr}, while Visibility-Aware RRT$^*$ builds on LQR-CBF steering by introducing an additional visibility-aware CBF for perception-limited robots in partially unknown environments~\cite{kim2025visibility}.
While these approaches provide significantly more directed edge expansions than random propagation, they share critical limitations for nonlinear systems: the required linearizations are valid only locally, and repeatedly solving Riccati or related optimal-control equations, computing first-order Taylor approximations, or evaluating constraints during tree expansion can introduce a sequential computational burden that grows with tree size and state dimensionality.
An alternative line of work leverages differential flatness to transform the planning problem into a flat-output space, enabling ultrafast parallelized kinodynamic rollouts and reducing the need for expensive numerical steering~\cite{bascetta2017flat, duong2026ultrafast}. While highly efficient, this approach is limited to systems admitting a suitable flat-output representation.
Learning-based methods offer another direction, using prior experience to bias sampling, generate paths, or provide learned local planners, as demonstrated by Neural RRT$^*$ for non-uniform sampling~\cite{wang2020neural}, Motion Planning Networks (MPNet) for path generation~\cite{qureshi2019motion} and their kinodynamic extension~\cite{li2021mpc}, and RL-RRT for learned steering and reachability estimation~\cite{chiang2019rl}.
Although these methods can accelerate planning, their practical performance depends on the quality and coverage of the training data and on the ability of the learned components to generalize to unseen conditions.

In parallel, a significant line of research focuses on optimizing the global sampling distribution rather than the steering operator itself. For instance, Informed RRT$^*$ restricts sampling to an admissible ellipsoidal subset after an initial solution is found to accelerate convergence~\cite{gammell2014informed}.
Similarly, G-RRT$^*$ combines bidirectional search with greedy informed sampling~\cite{kyaw2024greedy}, whereas ASD-RRT$^*$ uses adaptive sampling and densification~\cite{wang2026asd} and HB-RRT combines hybrid sampling with space optimization~\cite{hu2026path}, thereby directing exploration toward regions more likely to improve the current solution or support collision-free progress. Despite these advances, these methods primarily improve where samples are generated rather than how dynamically feasible local motion segments are constructed.
In contrast, SI-MPPI avoids the need for exact BVP steering, local linearization, learned steering functions, or purely random control propagation by employing MPPI as a cost-aware local expansion operator that forward-simulates the robot's dynamics directly within the search tree.
Consequently, each tree edge represents an MPPI-generated, dynamically feasible motion segment that is admitted only after explicit feasibility and collision checks.

\subsection{Graph-Search and Hybrid Kinodynamic Planning}

Graph-search approaches offer an alternative paradigm by formulating planning as graph search over a discretized representation of the state--action space~\cite{lavalle2006planning, sotirchos2024search}. 
Classical methods such as Dijkstra~\cite{dijkstra1959note} and A$^*$~\cite{hart1968formal} compute shortest paths over weighted graphs, while anytime and incremental/replanning variants such as D$^*$ Lite~\cite{koenig2002d}, Anytime Repairing A$^*$ (ARA$^*$)~\cite{likhachev2003ara}, Lifelong Planning A$^*$ (LPA$^*$)~\cite{koenig2004lifelong}, and Anytime Dynamic A$^*$ (AD$^*$)~\cite{likhachev2005anytime} improve responsiveness under time constraints or changing environments. 
For kinodynamic systems, this formulation is commonly extended through state lattices~\cite{pivtoraiko2009differentially, likhachev2009planning} or precomputed motion primitives~\cite{liu2017search}, which encode feasible local maneuvers into the graph. 
These methods can provide guarantees with respect to the chosen graph, lattice, or primitive set, but their performance depends strongly on the resolution of the underlying graph or state lattice and on the quality and coverage of the motion-primitive library.
To reduce this search burden, compact geometric representations of collision-free space have been explored, including convex obstacle-free regions~\cite{deits2015computing}, rectangular corridor decompositions~\cite{gonzalez2026fast}, and graphs of convex sets~\cite{marcucci2023motion}.
Such representations can compactly encode collision-free space and support efficient planning, with corridor-based decompositions enabling substantial search-space reduction and rapid online planning in structured environments.
However, depending on the formulation, these approaches may require precomputed decompositions, environment-specific geometric structure, simplified dynamics, or analytical steering assumptions, thereby introducing additional preprocessing or modeling requirements for cluttered occupancy-grid environments with nonlinear kinodynamic constraints.

Hybrid search--optimization planners attempt to bridge discrete topological reasoning and continuous dynamic feasibility by combining graph search with trajectory optimization. Frameworks such as INterleaved Search And Trajectory optimization (INSAT)~\cite{natarajan2021interleaving} and its parallel extension PINSAT~\cite{natarajan2024pinsat} interleave low-dimensional graph search with full-dimensional trajectory optimization to evaluate and refine partial trajectories during the search, while Iterative Discontinuity Bounded A$^*$ (iDb-A$^*$) searches over short motion primitives connected with bounded discontinuities and subsequently repairs the discontinuities in the resulting primitive sequence through local trajectory optimization~\cite{ortiz2024idb}.
These methods demonstrate the value of coupling global search structure with local continuous optimization, but they face a practical trade-off: the search component depends on graph resolution or primitive-set coverage, whereas the optimization component can be sensitive to the quality of the search-generated initialization.
SI-MPPI differs from these hybrid architectures by constructing its search tree directly from admitted MPPI-optimized motion segments in an occupancy-grid environment, thereby coupling global search with local continuous trajectory optimization throughout tree growth, without relying on fixed lattices, precomputed motion primitives, or environment-specific decompositions.

\subsection{MCTS-Based Kinodynamic Planning}\label{subsec:related_mcts}


As a flexible alternative to fixed graph and lattice search, MCTS builds an asymmetric search tree, commonly using UCB-based selection and backpropagated statistics to adaptively balance exploration and exploitation, making it well suited for long-horizon robotic planning under complex constraints~\cite{kocsis2006bandit, browne2012survey}.
This adaptive allocation mechanism has been applied in several domains, including Pareto MCTS for multi-objective informative planning~\cite{chen2019pareto}, MCTS integrated with velocity obstacles for dynamic obstacle avoidance~\cite{bonanni2025monte}, Tensor Train Tree Search (TTTS) using tensor-factorized tree representations to reduce computation and storage complexity~\cite{xue2025monte}, parallel MCTS for constraint-aware multilegged contact planning~\cite{xu2025contact}, and MCTS coupled with trajectory optimization or Model Predictive Control (MPC) for manipulation and locomotion~\cite{zhu2023efficient, amatucci2022monte, dhedin2025simultaneous}.
Learned policy--value guidance and state-occupancy regularization have also been used to guide tree growth and improve long-horizon exploration~\cite{riviere2021neural, schramm2024volume}.
Collectively, these works illustrate how task-informed representations, learned guidance, feasibility constraints, and optimization modules can improve MCTS-based robotic planning.

In continuous kinodynamic settings, key difficulties include controlling tree growth and obtaining informative finite-sample rollout estimates~\cite{veness2011variance}.
To manage the large effective branching factor induced by continuous state--action spaces, continuous-space MCTS variants often employ progressive or double progressive widening, in which new actions or successor states are gradually added as node visit counts increase~\cite{couetoux2011continuous, sunberg2018online}.
To improve continuous-action selection and exploit local smoothness, Value-Gradient UCT (VG-UCT) couples global exploration based on Upper Confidence bounds applied to Trees (UCT) with gradient-based local optimization~\cite{lee2020monte}.
More recent approaches attempt to manage continuous dynamics through representation and clustering: Spectral Expansion Tree Search (SETS) constructs a low-complexity spectral representation from the locally linearized system~\cite{riviere2024monte}, while Continuous Monte Carlo Graph Search (CMCGS) clusters similar states into stochastic action-bandit nodes that share action policies~\cite{kujanpaa2024continuous}.
Although these extensions broaden the applicability of MCTS beyond purely discrete domains, they manage complexity through widening rules, spectral representations, state clustering, learned guidance, or task-specific feasibility and optimization mechanisms.
Consequently, directly embedding a GPU-batched, sampling-based nonlinear trajectory optimizer as the MCTS edge-generation and rollout mechanism remains insufficiently studied for collision-aware kinodynamic navigation in non-convex, cluttered occupancy-grid environments.

\subsection{MPPI-Based Trajectory Optimization}
\label{subsec:related_mppi}

Recent MPPI research has sought to improve finite-horizon optimization through adaptive or biased sampling distributions and risk-aware trajectory evaluation~\cite{asmar2023model, mohamed2022autonomous, yin2023risk, mohamed2025toward}, with broader developments, including learning-guided priors and open challenges, reviewed in~\cite{mohamed2026mppi_priors}.
Although these methods can improve exploration efficiency, robustness, or safety, they do not by themselves overcome the limited lookahead of finite-horizon MPPI, leaving the optimizer susceptible to obstacle-induced local minima in cluttered, non-convex environments.
A longer optimization horizon can provide greater lookahead, but can also increase the dimension of the control-search space and the associated computational burden~\cite{chatterjee2026improving}.
To extend guidance beyond the immediate horizon, GP-MPPI uses a sparse Gaussian process (GP)-based local perception model to recommend subgoals for navigation in unknown environments~\cite{mohamed2023gpmppi}; however, its guidance remains perception-driven and does not maintain a persistent search tree over competing long-horizon alternatives. 
More explicit global guidance has been introduced through hierarchical architectures.
For instance, RRT-guided MPPI uses an RRT-generated geometric path to initialize the nominal control distribution~\cite{tao2023rrt}, while a subsequent variant incorporates CBFs for safety-aware adaptation of the sampling distribution~\cite{tao2026guided}.
Similarly, planning-tree value functions have been employed as terminal costs to approximate infinite-horizon behavior~\cite{hatch2021value}.
Across these approaches, guidance remains structurally separate from local MPPI optimization and is supplied through a perception-derived subgoal, geometric path, or terminal cost.
SI-MPPI instead embeds MPPI within the MCTS tree-expansion process: GPU-batched MPPI rollouts generate and score candidate motion segments from selected frontier states, admitted segments become tree edges, and rollout-derived costs are backpropagated to guide subsequent tree selection.

\section{Problem Statement}
\label{sec:problem}

Consider a robotic system with state $\mathbf{x}_k \in \mathcal{X} \subseteq \mathbb{R}^{n_x}$ and nominal control input $\mathbf{u}_k \in \mathcal{U} \subseteq \mathbb{R}^{n_u}$ at discrete time step $k$. The system evolves according to the nonlinear, discrete-time stochastic dynamics
\begin{equation}
\label{eq:dynamics_system}
    \mathbf{x}_{k+1}
    =
    f\!\left(\mathbf{x}_k,\mathbf{w}_k\right),
    \qquad
    \mathbf{w}_k
    =
    \mathbf{u}_k+\delta\mathbf{u}_k,
\end{equation}
where $f:\mathcal{X}\times\mathbb{R}^{n_u}\rightarrow\mathcal{X}$ denotes the system dynamics, $\mathbf{w}_k\in\mathbb{R}^{n_u}$ is the perturbed control input, and $\delta\mathbf{u}_k\sim\mathcal{N}(\mathbf{0},\boldsymbol{\Sigma}_{\mathbf{u}})$ is an injected zero-mean Gaussian control perturbation with covariance $\boldsymbol{\Sigma}_{\mathbf{u}}\in\mathbb{R}^{n_u\times n_u}$, used to model actuation uncertainty and to generate stochastic control rollouts during the sampling-based optimization described in Section~\ref{subsec:mppi_overview}. Equivalently, $\mathbf{w}_k\sim\mathcal{N}(\mathbf{u}_k,\boldsymbol{\Sigma}_{\mathbf{u}})$. Over a finite prediction horizon $N$, let $\mathbf{U}=[\mathbf{u}_{0},\ldots,\mathbf{u}_{N-1}]^{\top}\in\mathbb{R}^{N\times n_u}$ denote the nominal control sequence and $\tau=[\mathbf{x}_{0},\ldots,\mathbf{x}_{N}]^{\top}\in\mathbb{R}^{(N+1)\times n_x}$ denote a corresponding state trajectory.

The robot operates in a cluttered workspace $\mathcal{W}\subseteq\mathbb{R}^{d}$, where $\mathcal{O}\subset\mathcal{W}$ denotes the obstacle region and $\mathcal{A}(\mathbf{x})\subset\mathcal{W}$ the robot footprint at state $\mathbf{x}$. The collision-free state space is $\mathcal{X}_{\mathrm{free}} = \left\{ \mathbf{x}\in\mathcal{X}  \,\middle|\, \mathcal{A}(\mathbf{x})\cap\mathcal{O}  = \emptyset \right\}$. Given an initial state $\mathbf{x}_{\mathrm{start}}\in\mathcal{X}_{\mathrm{free}}$ and a target $\mathbf{x}_{\mathrm{goal}}\in\mathcal{X}_{\mathrm{goal}}\subseteq\mathcal{X}_{\mathrm{free}}$, the finite-horizon stochastic optimal-control problem is to determine a feasible control sequence $\mathbf{U}$ and corresponding state trajectory $\tau$ that steer the robot toward the goal region $\mathcal{X}_{\mathrm{goal}}$ while minimizing the expected trajectory cost:
\begin{subequations}
    \label{eq:finite_horizon_problem}
    \begin{align}
        \min_{\mathbf{U},\tau}
        \quad
        &\mathbb{E}
        \left[
            \phi(\mathbf{x}_{N})
            +
            \sum_{k=0}^{N-1}
            \left(
                q(\mathbf{x}_{k})
                +
                \frac{1}{2}
                \mathbf{u}_{k}^{\top}
                \mathbf{R}
                \mathbf{u}_{k}
            \right)
        \right]
        \label{eq:finite_horizon_objective}
        \\
        &\mathrm{s.t.}
        \quad
        \mathbf{x}_{k+1}
        =
        f(\mathbf{x}_{k},\mathbf{w}_{k}),
        \;\;
        \mathbf{w}_{k}
        \sim
        \mathcal{N}
        \left(
            \mathbf{u}_{k},
            \boldsymbol{\Sigma}_{\mathbf{u}}
        \right),
        \label{eq:finite_horizon_dynamics}
        \\
        &\phantom{\mathrm{s.t.}\quad}
        \mathbf{x}_{0}
        =
        \mathbf{x}_{\mathrm{start}},
        \qquad \qquad \! \!
        \mathbf{u}_{k}
        \in
        \mathcal{U},
        \label{eq:finite_horizon_initial}
        \\
        &\phantom{\mathrm{s.t.}\quad}
        \mathbf{x}_{k}
        \in
        \mathcal{X}_{\mathrm{free}}
        \subseteq
        \mathcal{X},
        \qquad
        h(\mathbf{x}_{k},\mathbf{u}_{k})
        \preceq
        \mathbf{0}.
        \label{eq:finite_horizon_constraints}
    \end{align}
    \end{subequations}
    where
$q(\mathbf{x}_{k})\in\mathbb{R}_{\geq 0}$ and $\phi(\mathbf{x}_{N})\in\mathbb{R}_{\geq 0}$ denote the running and terminal state costs, respectively,
$\mathbf{R}\in\mathbb{R}^{n_u\times n_u}$, with $\mathbf{R}=\mathbf{R}^{\top}\succ\mathbf{0}$, weights the control effort, $\mathbf{x}_{k}\in\mathcal{X}_{\mathrm{free}}$
enforces collision avoidance along the feasible nominal trajectory, and $h(\mathbf{x}_{k},\mathbf{u}_{k})$ represents additional state and control constraints.
It is worth noting that the conditions in~\eqref{eq:finite_horizon_problem} hold for $k=0,\ldots,N-1$, while the collision-avoidance constraint $\mathbf{x}_{k}\in\mathcal{X}_{\mathrm{free}}$ in~\eqref{eq:finite_horizon_constraints} is additionally enforced at $k=N$ to ensure that the terminal state of the rollout is collision-free.

The goal region $\mathcal{X}_{\mathrm{goal}}$ may not be reachable from $\mathbf{x}_{\mathrm{start}}$ within a single prediction horizon $N$. Although repeated solutions of~\eqref{eq:finite_horizon_problem} can extend the trajectory beyond this horizon, finite-horizon optimization remains locally informed and may become trapped in obstacle-induced local minima in cluttered, non-convex environments. The complete kinodynamic planning problem therefore requires coordinating a finite sequence of dynamically feasible, collision-free state--control trajectories whose ordered concatenation connects $\mathbf{x}_{\mathrm{start}}$ to $\mathcal{X}_{\mathrm{goal}}$.
Let $S\in\mathbb{N}_{>0}$ denote the number of local motion segments, and let $\{(\tau^{(s)},\mathbf{U}^{(s)})\}_{s=1}^{S}$ denote the corresponding state and control sequences, each spanning $N$ control intervals. The complete mission trajectory is defined by the ordered concatenation
\begin{equation}
    \label{eq:global_trajectory}
        \Gamma
        =
        \tau^{(1)}
        \oplus
        \tau^{(2)}
        \oplus
        \cdots
        \oplus
        \tau^{(S)}
        \in
        \mathbb{R}^{(L+1)\times n_x},
\end{equation}
where $\oplus$ denotes temporal concatenation without duplicating shared boundary states, and $L=SN$ is the total number of control intervals in the completed mission.
The associated mission control sequence is
\begin{equation}
\mathbf{U}_{\Gamma}
=
\mathbf{U}^{(1)}
\oplus
\mathbf{U}^{(2)}
\oplus
\cdots
\oplus
\mathbf{U}^{(S)}
\in
\mathbb{R}^{L\times n_u}.
\end{equation}
The constituent trajectories must satisfy the mission boundary and intersegment continuity conditions
\begin{equation}
\label{eq:global_boundary_conditions}
\begin{aligned}
    \mathbf{x}_{0}^{(1)}
    &=
    \mathbf{x}_{\mathrm{start}},
    \qquad
    \mathbf{x}_{N}^{(S)}
    \in
    \mathcal{X}_{\mathrm{goal}},
    \\
    \mathbf{x}_{N}^{(s)}
    &=
    \mathbf{x}_{0}^{(s+1)},
    \qquad
    \forall s\in\{1,\ldots,S-1\}.
\end{aligned}
\end{equation}
where the mission boundary conditions initialize the first segment at $\mathbf{x}_{\mathrm{start}}$ and require the terminal state of the final segment, $\mathbf{x}_{N}^{(S)}$, to belong to $\mathcal{X}_{\mathrm{goal}}$, while the intersegment equalities identify the terminal state of segment $s$ with the initial state of segment $s+1$, thereby ensuring state continuity, i.e., $C^{0}$ continuity, at the segment boundaries. Moreover, each constituent pair $(\tau^{(s)},\mathbf{U}^{(s)})$, $s\in\{1,\ldots,S\}$, must satisfy the system dynamics, collision-avoidance requirements, admissible-control conditions, and additional state--control constraints in~\eqref{eq:finite_horizon_problem}.

Let $\Pi_{\mathrm{feas}}$ denote the set of all finite state--control plans $\pi=(\Gamma,\mathbf{U}_{\Gamma})$ that satisfy \eqref{eq:global_trajectory}--\eqref{eq:global_boundary_conditions} and all segment-wise feasibility requirements, where the number of segments $S$ may vary across plans. 
Given a nonnegative trajectory-cost function $c:\Pi_{\mathrm{feas}}\rightarrow\mathbb{R}_{\geq 0}$, and assuming that $\Pi_{\mathrm{feas}}\neq\emptyset$ and that $c$ attains its minimum on $\Pi_{\mathrm{feas}}$, the optimal kinodynamic planning problem can be formulated as the search for a minimum-cost feasible plan
\vspace{-0.6em}
\begin{equation}
\label{eq:optimal_kinodynamic_problem}
    \pi^{\star}
    \in
    \arg\min_{\pi\in\Pi_{\mathrm{feas}}}
    c(\pi).
\end{equation}
\vspace{-0.6em}
\begin{remark}
    Since $\Pi_{\mathrm{feas}}$ contains feasible plans with different numbers of motion segments, the minimization in~\eqref{eq:optimal_kinodynamic_problem} implicitly ranges over plan lengths; thus, any minimizer $\pi^{\star}$ induces a number of segments $S^{\star}$ and a mission horizon $L^{\star}=S^{\star}N$, without $S$ or $L$ being treated as independent decision variables. In practice, planners operating under finite computational budgets generally seek feasible, low-cost approximations to the optimal plan $\pi^{\star}$.
    \end{remark}

\section{Preliminaries}
\label{sec:preliminaries}

This section reviews the two building blocks underlying the proposed framework: MPPI for approximating the local finite-horizon optimal-control problem in~\eqref{eq:finite_horizon_problem} and MCTS for guiding long-horizon search over sequences of motion segments.

\subsection{MPPI Algorithm Overview}
\label{subsec:mppi_overview}

MPPI is a sampling-based stochastic optimal-control method for nonlinear dynamical systems based on the path-integral formulation~\cite{williams2017model,williams2018information}. For the discrete-time dynamics in~\eqref{eq:dynamics_system}, MPPI provides a sampling-based approximation to the finite-horizon stochastic optimal-control problem in~\eqref{eq:finite_horizon_problem} by sampling perturbations about a nominal control sequence and forward-propagating $M$ trajectories under the dynamics $f$ over the prediction horizon $N$.
The nominal control sequence $\mathbf{U} = \{\mathbf{u}_{k}\}_{k=0}^{N-1}$ is then iteratively updated using a cost-weighted sum of the sampled control perturbations:
\begin{equation}
\label{eq:mppi_optimal-control}
\mathbf{u}_{k}
\leftarrow
\mathbf{u}_{k}
+
\sum_{m=1}^{M}
w_m\,\delta\mathbf{u}_{k,m},
\end{equation}
where $w_m$ denotes the normalized importance weight assigned to the $m$th sampled trajectory and is computed from its trajectory cost $J_m$ as
\begin{equation}
\label{eq:mppi_weights}
w_m
=
\frac{
\exp\Bigl(-\frac{1}{\lambda}\bigl(J_m-J_{\min}\bigr)\!\Bigr)
}{
\sum_{j=1}^{M}
\exp\Bigl(-\frac{1}{\lambda}\bigl(J_j-J_{\min}\bigr)\!\Bigr)
}, \;\; m\in\{1,\ldots,M\}, 
\end{equation}
with $J_{\min}=\min_{j\in\{1,\ldots,M\}}J_j$ and $\lambda>0$ controlling the sensitivity of the weighting to differences in trajectory cost. The cost-to-go of the $m$th sampled trajectory is calculated as
\begin{equation}
\label{eq:cost-to-go}
    J_m
    =
    \phi\!\left(\mathbf{x}_{N,m}\right)
    +
    \sum_{k=0}^{N-1}
    \widetilde{q}
    \!\left(
        \mathbf{x}_{k,m},
        \mathbf{u}_{k},
        \delta\mathbf{u}_{k,m}
    \right),
\end{equation}
where the instantaneous running cost $\widetilde{q}$ is decomposed as
\begin{equation}
\label{eq:mppi_running_cost}
    \widetilde{q}
    \!\left(
        \mathbf{x}_{k,m},
        \mathbf{u}_{k},
        \delta\mathbf{u}_{k,m}
    \right)
    =
    q\!\left(\mathbf{x}_{k,m}\right)
    +
    q_{\mathbf{u}}
    \!\left(
        \mathbf{u}_{k},
        \delta\mathbf{u}_{k,m}
    \right).
\end{equation}
Here, $q(\mathbf{x}_{k,m})$ denotes the task-dependent state cost, while the quadratic MPPI control cost $q_{\mathbf{u}}(\mathbf{u}_{k},\delta\mathbf{u}_{k,m})$ follows the standard formulation in~\cite{williams2017model} and is given by
\begin{equation}
\label{eq:mppi_control_cost}
    q_{\mathbf{u}}
    =
    \gamma_{\mathbf{u}}
    \delta\mathbf{u}_{k,m}^{\top}
    \mathbf{R}
    \delta\mathbf{u}_{k,m}
    +
    \mathbf{u}_{k}^{\top}
    \mathbf{R}
    \delta\mathbf{u}_{k,m}
    +
    \frac{1}{2}
    \mathbf{u}_{k}^{\top}
    \mathbf{R}
    \mathbf{u}_{k},
\end{equation}
where $\gamma_{\mathbf{u}}=\frac{1}{2}\frac{(\nu-1)}{\nu}$ and $\nu\geq1$ is the exploration parameter. Repeating the update in~\eqref{eq:mppi_optimal-control} refines the nominal control sequence using the costs of the sampled trajectories.

\begin{figure*}[t]
    \centering
    \includegraphics[width=\textwidth]{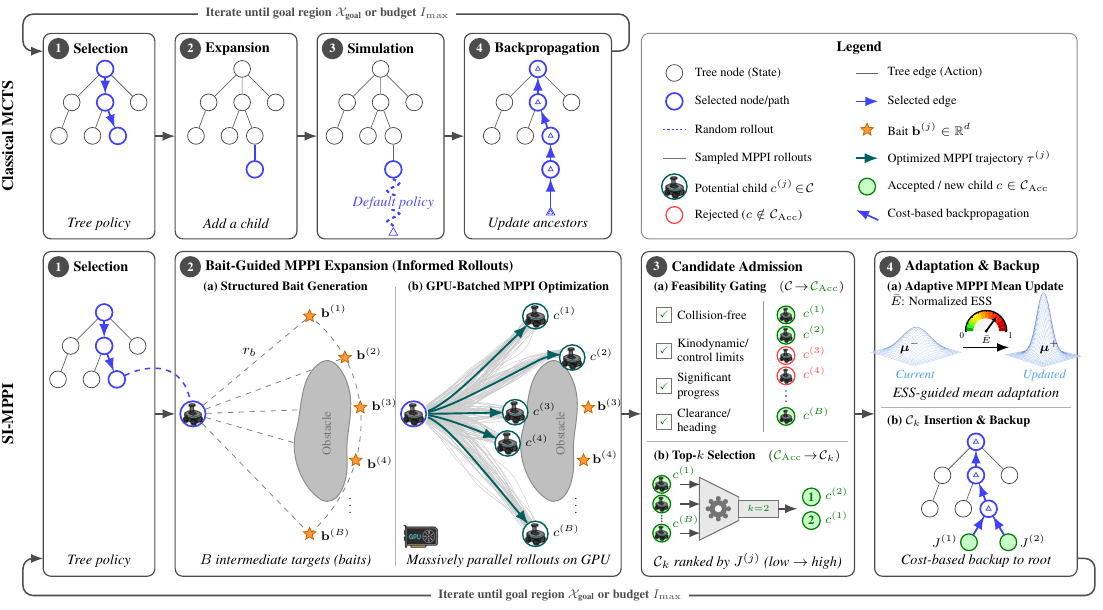}
    \vspace*{-16pt}
    \caption{Classical MCTS~(top) versus the proposed SI-MPPI pipeline~(bottom). Both employ UCB-based selection and iterate until the goal region $\mathcal{X}_{\mathrm{goal}}$ is reached or the iteration budget $I_{\max}$ is exhausted. Classical MCTS follows the standard Selection $\rightarrow$ Expansion $\rightarrow$ Simulation $\rightarrow$ Backpropagation cycle. SI-MPPI instead proceeds through four stages: (1)~Selection returns a selected node $n_{\mathrm{s}}$ under the UCB tree policy; (2)~Bait-Guided MPPI Expansion constructs the structured bait set $\mathcal{B}=\{\mathbf{b}^{(j)}\}_{j=1}^{B}$, after which parallel MPPI optimizations are performed for the retained baits $\mathcal{B}_{\mathrm{ret}}$, each serving as a local target, yielding the potential-candidate set $\mathcal{C}=\{c^{(j)}\}_{j=1}^{B_{\mathrm{ret}}}$, where $B_{\mathrm{ret}}=|\mathcal{B}_{\mathrm{ret}}|\leq B$ and $c^{(j)}=(\mathbf{U}^{(j)},\tau^{(j)},J^{(j)})$; (3)~Candidate Admission maps $\mathcal{C}\to\mathcal{C}_{\mathrm{Acc}}$ through feasibility gating and subsequently $\mathcal{C}_{\mathrm{Acc}}\to\mathcal{C}_{k}$ through top-$k$ ranking based on $J^{(j)}$; and (4)~Adaptation~\&~Backup updates the MPPI nominal control mean $\boldsymbol{\mu}$ using the candidate statistics and backpropagates the backup cost $J_{\mathrm{bp}}$ along the selected tree path to the root.}
    \label{fig:mcts_si-mppi_diagram}
\end{figure*}
\subsection{MCTS Algorithm Overview}
\label{subsec:mcts}
MCTS is an anytime, sampling-based tree-search method that incrementally builds a partial search tree while balancing the exploitation of branches
with high estimated value against the exploration of less-visited alternatives~\cite{kocsis2006bandit,browne2012survey}. 
In the classical formulation, each node represents a state of the underlying decision process, while each directed edge represents an action leading to a successor state. Following the classical pipeline in Fig.~\ref{fig:mcts_si-mppi_diagram}~(top), each MCTS iteration comprises four phases: Selection, Expansion, Simulation, and Backpropagation. Selection and Expansion constitute the \emph{tree policy}, whereas Simulation follows a \emph{default policy}.

\textit{1) Selection:}
Starting from the root node $n_0$, the algorithm descends through the existing tree by recursively selecting child nodes according to the tree policy until a terminal or expandable node is reached.
A widely used MCTS tree-selection policy is UCT~\cite{kocsis2006bandit}, which selects, at each internal node, the child node $n_i$ that maximizes the UCB score
\begin{equation}
\label{eq:ucb_prelim}
    \mathrm{UCB}(n_i)
    =
    \bar{R}_i
    +
    c_{\mathrm{ucb}}
    \sqrt{
        \frac{\ln V_p}{V_i}
    },
\end{equation}
where $\bar{R}_i$ is the empirical average reward of node $n_i$, $V_i$ and $V_p$ are the visit counts of node $n_i$ and its parent, respectively, and $c_{\mathrm{ucb}}\in\mathbb{R}^{+}$ controls the exploration--exploitation trade-off. 

\textit{2) Expansion:}
When the selection phase terminates at an expandable node, the tree is enlarged by adding one or more child nodes corresponding to actions not yet represented among the node's outgoing branches.

\textit{3) Simulation:}
From a newly added child node, a rollout, or play-out, is executed according to the default policy to estimate the reward of the corresponding branch. In classical MCTS, the default policy is commonly random and is applied until a terminal condition or a prescribed simulation horizon is reached.

\textit{4) Backpropagation:}
The resulting reward is propagated from the expanded node to the root along the selected path. 
The visit counts and empirical reward estimates of the traversed nodes are then updated, thereby influencing subsequent tree-policy decisions.

By iterating this cycle, MCTS expands the search tree asymmetrically, using accumulated node statistics to prioritize promising branches while continuing to explore less-visited alternatives.
The classical MCTS formulation is particularly well suited to settings in which the action set is finite or enumerable and default-policy simulations are inexpensive~\cite{browne2012survey}.
Its direct application to continuous kinodynamic planning is more challenging because candidate actions are drawn from a continuous control space, while each rollout requires repeated dynamics propagation, constraint evaluation, and collision checking. Random default-policy simulations may therefore be computationally expensive and produce high-variance or weakly informative reward estimates. 
These limitations motivate a dynamics-aware expansion mechanism capable of generating informative, dynamically feasible local motion in continuous control spaces.

\section{Search-Informed MPPI (SI-MPPI) Framework}
\label{sec:si-mppi}
To exploit the complementary strengths of MCTS and MPPI while addressing the limitations of classical MCTS in continuous kinodynamic planning, we develop a unified search--optimization framework that integrates UCB-guided tree search with GPU-batched MPPI trajectory generation for the long-horizon kinodynamic planning problem in~\eqref{eq:optimal_kinodynamic_problem}.
As illustrated in the bottom panel of Fig.~\ref{fig:mcts_si-mppi_diagram}, SI-MPPI retains the MCTS selection and backpropagation stages while unifying the roles of expansion and simulation within a bait-guided MPPI stage. Specifically, at each selected tree node, strategically generated intermediate targets, termed \emph{baits}, define a diverse set of local optimization objectives. GPU-batched MPPI then generates the corresponding control--trajectory candidates, which are evaluated and filtered before selected motion segments are admitted as tree edges. Candidate costs update the tree statistics, while the associated optimized control sequences provide warm-starts for subsequent MPPI expansions. Through this coupling, MCTS allocates computation among competing long-horizon alternatives, whereas MPPI generates cost-informed local motion segments consistent with the system dynamics.

\textit{Tree Representation and Planning Cycle:}
SI-MPPI maintains a rooted directed tree $\mathcal{T}=(\mathcal{V},\mathcal{E})$, where $\mathcal{V}$ and $\mathcal{E}$ denote the sets of nodes and directed edges, respectively. The root node $n_0\in\mathcal{V}$ represents the mission start state $\mathbf{x}_{\mathrm{start}}$. For a parent node $n_p\in\mathcal{V}$, each admitted candidate creates a child node $n_i\in\mathcal{V}$ representing the terminal state $\mathbf{x}_N^{(i)}$ of the finite-horizon trajectory $\tau^{(i)}=[\mathbf{x}_0^{(i)},\ldots,\mathbf{x}_N^{(i)}]^{\top}$ induced by the optimized control sequence $\mathbf{U}^{(i)}=[\mathbf{u}_0^{(i)},\ldots,\mathbf{u}_{N-1}^{(i)}]^{\top}$, where $\mathbf{x}_0^{(i)}$ is the state represented by the parent node $n_p$. 
The directed edge $e_{p,i}=(n_p,n_i)\in\mathcal{E}$ stores the pair $(\mathbf{U}^{(i)},\tau^{(i)})$ and is associated with the candidate cost $J^{(i)}$, thereby encoding the finite-horizon state--control transition from $n_p$ to $n_i$ and its corresponding cost.
Since branching occurs only at the terminal states of admitted motion segments, every nonroot node is connected to the root through a sequence of dynamics-consistent state--control transitions. A root-to-goal path in $\mathcal{T}$ therefore corresponds to the concatenated plan $\pi=(\Gamma,\mathbf{U}_{\Gamma})$ defined in \eqref{eq:global_trajectory}--\eqref{eq:global_boundary_conditions}. 
This node--edge representation enables the long-horizon planning problem to be posed as a search over sequences of MPPI-generated local motion segments.

With this tree representation in place, our proposed planner incrementally expands $\mathcal{T}$ through the four-stage planning cycle depicted in the bottom panel of Fig.~\ref{fig:mcts_si-mppi_diagram} and described as follows:

\textit{1) Selection:}
The UCB tree policy traverses $\mathcal{T}$ to select a frontier node for expansion, balancing the exploitation of branches with lower estimated cost against the exploration of less-visited alternatives
(Sec.~\ref{subsec:selection}).

\textit{2) Bait-Guided MPPI Expansion:}
From the selected node, $B$ structured baits are generated and screened. Of these, $B_{\mathrm{ret}}\leq B$ baits are retained as local optimization targets for a GPU-batched MPPI stage,  which generates the candidate set $\mathcal{C}=\{c^{(j)}\}_{j=1}^{B_{\mathrm{ret}}}$, where $c^{(j)}=(\mathbf{U}^{(j)},\tau^{(j)},J^{(j)})$
contains an optimized control sequence, its induced trajectory, and the corresponding candidate cost (Sec.~\ref{subsec:expansion}).

\textit{3) Candidate Admission and Tree Integration:}
Feasibility gating applies the prescribed collision-safety and progress criteria to obtain the accepted subset $\mathcal{C}_{\mathrm{Acc}}\subseteq\mathcal{C}$. 
At most $k_b$ of the lowest-cost accepted candidates form $\mathcal{C}_{k}\subseteq\mathcal{C}_{\mathrm{Acc}}$; their terminal states are added as child nodes, while the corresponding control sequences and trajectories are stored on the incoming tree edges and the child reward statistics are initialized from the associated candidate costs (Sec.~\ref{subsec:filtering}).

\needspace{3\baselineskip} 
\textit{4) Proposal Adaptation and Backpropagation:}
Candidate-cost statistics are used to adapt the MPPI nominal control mean, thereby providing a warm start for subsequent expansions, while a cost-based backup statistic derived from the retained candidate set $\mathcal{C}_{k}$ is backpropagated along the selected tree path to update the visit counts and edge-cost-seeded reward statistics used by UCB selection (Sec.~\ref{subsec:backprop}).

The planning cycle continues until either the state represented by an admitted node $n_i$ lies in the goal region $\mathcal{X}_{\mathrm{goal}}$, i.e., $\mathbf{x}_{N}^{(i)}\in\mathcal{X}_{\mathrm{goal}}$, or the iteration budget $I_{\max}$ is exhausted.

\subsection{Selection}
\label{subsec:selection}

The selection stage identifies the node from which the next bait-guided MPPI expansion is initiated. Starting from the root node $n_0$, the tree policy recursively descends $\mathcal{T}$ according to the following rule. If the current node has not yet undergone the bait-guided MPPI expansion described in Section~\ref{subsec:expansion}, the traversal terminates and that node is returned as the selected node $n_{\mathrm{s}}$. Otherwise, if the node has children, the traversal continues through the child node $n_i$ that maximizes the UCB score in~\eqref{eq:ucb_prelim}. If the node has already undergone expansion but has no children, the corresponding branch is treated as a dead end. For an unvisited child, i.e., $V_i=0$, we assign $\mathrm{UCB}(n_i)=+\infty$, ensuring that each unvisited child is selected before previously visited siblings are reconsidered.

To align the UCT policy with the cost-minimization objective of kinodynamic planning, our proposed framework retains the standard exploration term in~\eqref{eq:ucb_prelim} while adapting the exploitation term $\bar{R}_i$. Unlike classical MCTS, which typically estimates node values from default-policy simulation rewards, SI-MPPI forms the reward of each retained child from the cost of its incoming MPPI-generated motion segment and subsequently updates it through cost backpropagation. Let $J_i^{\mathrm{edge}}$ denote the candidate cost associated with the incoming edge of node $n_i$, and let $J_{\mathrm{bp},i}^{(\ell)}$ denote the expansion-level cost backpropagated through $n_i$ during its $\ell$th visit, as defined in Section~\ref{subsubsec:backprop}. For $V_i>0$, the edge-cost-seeded reward statistic $\bar{R}_i$ used as the exploitation term is defined as
\begin{equation}
\label{eq:avg_reward}
    \bar{R}_i
    =
    \frac{R_i}{V_i}
    =
    -
    \frac{1}{V_i}
    \Biggl[
        J_i^{\mathrm{edge}}
        +
        \sum_{\ell=1}^{V_i}
        J_{\mathrm{bp},i}^{(\ell)}
    \Biggr],
\end{equation}
where $R_i$ denotes the cumulative reward stored at node $n_i$, and $J_0^{\mathrm{edge}}=0$ for the root node. Specifically, the incoming-edge cost $J_i^{\mathrm{edge}}$ provides a direct evaluation of the motion segment that created the node, whereas the accumulated backpropagated costs $J_{\mathrm{bp},i}^{(\ell)}$ reflect the outcomes of subsequent expansions along that branch. The incorporation of $J_i^{\mathrm{edge}}$ into the node reward in~\eqref{eq:avg_reward} is termed \emph{edge-cost seeding}: when a retained child is inserted, its cumulative reward is initialized as $R_i=-J_i^{\mathrm{edge}}$ with $V_i=0$, as formalized in~\eqref{eq:edge_reward_initialization}; subsequent backpropagation updates $R_i$ beyond this seed. This construction is conceptually related to search seeding in MCTS~\cite{browne2012survey}, where newly created node statistics are initialized using heuristic knowledge; in SI-MPPI, however, the seed is obtained online from the cost of the retained motion segment rather than from an external heuristic.

It is worth noting that both cost contributions enter $R_i$ with negative signs, thereby expressing the cost-minimization objective in the reward-maximization form used by UCB; hence, larger values of $\bar{R}_i$ correspond to lower-cost branches. Nevertheless, since $\bar{R}_i$ is formed from unnormalized edge and backup costs whose magnitudes may vary considerably across the tree, using it directly in~\eqref{eq:ucb_prelim} would make the selection sensitive to the absolute cost scale and could allow the exploitation term to overwhelm the exploration term. To prevent this, when evaluating the UCB score, $\bar{R}_i$ is min--max normalized to $[0,1]$ among the visited siblings, following the same procedure as in~\eqref{eq:norm_cost}, so that both terms remain on comparable scales and $c_{\mathrm{ucb}}$ retains its role in controlling the exploration--exploitation trade-off. Consequently, maximizing the UCB score directs the tree traversal toward branches with lower estimated cost while preserving exploration of less-visited alternatives.

\subsection{Bait-Guided MPPI Expansion}
\label{subsec:expansion}
Given the selected node $n_{\mathrm{s}}$ returned by the selection stage and its associated state $\mathbf{x}_{s}\in\mathcal{X}$, the expansion stage aims to generate candidate finite-horizon motion segments through a unified bait-guided MPPI procedure, rather than through the discrete-action branching and random default-policy rollouts typical of classical MCTS. This procedure comprises two sequential steps: structured bait generation, which defines local optimization targets, and GPU-batched MPPI optimization, which optimizes a control sequence toward each retained bait.

\subsubsection{Structured Bait Generation}

Let $\mathbf{p}_{s}\in\mathcal{W}\subseteq\mathbb{R}^{d}$ denote the task-relevant workspace position associated with $\mathbf{x}_{s}$, and let $\mathbf{d}_{s}\in\mathbb{S}^{d-1}$ denote the preferred local expansion direction determined by the system state and task geometry. For planar mobile robots, $\mathbf{d}_{s}$ is naturally defined by the unit heading direction. For vehicles operating in three-dimensional workspaces, such as aerial or underwater robots, it may be defined by the body-forward or velocity direction, whereas for other robotic systems, such as manipulators, it may be selected according to the task.
To guide the local MPPI optimizations initiated from $n_{\mathrm{s}}$, we construct $B$ intermediate workspace targets, termed \emph{baits}, forming the set
\begin{equation}
\label{eq:bait_set}
    \mathcal{B}
    =
    \bigl\{\mathbf{b}^{(j)}\bigr\}_{j=1}^{B}
    \subset
    \mathbb{R}^{d}.
\end{equation}
Each bait $\mathbf{b}^{(j)}$ is placed on a local expansion shell of radius $r_b>0$ centered at $\mathbf{p}_{s}$:
\begin{equation}
\label{eq:baits}
    \mathbf{b}^{(j)}
    =
    \mathbf{p}_{s}
    +
    r_b\,\mathbf{d}^{(j)},
    \quad
    \mathbf{d}^{(j)}\in\mathbb{S}^{d-1},
    \quad
    \forall j\in\{1,\ldots,B\},
\end{equation}
where $\mathbb{S}^{d-1}=\{\mathbf{d}\in\mathbb{R}^{d}\mid\|\mathbf{d}\|_{2}=1\}$ denotes the unit sphere in $\mathbb{R}^{d}$. The directional set $\{\mathbf{d}^{(j)}\}_{j=1}^{B}$ is distributed over a forward-oriented subset of $\mathbb{S}^{d-1}$ centered on $\mathbf{d}_{s}$, thereby biasing the expansion toward locally preferred motion while retaining directional diversity to support obstacle avoidance.
For example, in planar workspaces ($d=2$), the directions may be distributed over a semicircular arc centered on $\mathbf{d}_{s}$, as illustrated in the \emph{Structured Bait Generation} panel of Fig.~\ref{fig:mcts_si-mppi_diagram}. In three-dimensional workspaces ($d=3$), they may instead be sampled over a spherical cap whose symmetry axis is $\mathbf{d}_{s}$, thereby preserving lateral and vertical diversity. Alternative constructions may be adopted depending on the system and task geometry. The planar parameterization used in the experiments is specified in Section~\ref{subsubsec:planning_setup}.

Following the bait construction in~\eqref{eq:baits}, we apply an \textit{adaptive bait-handling} rule to screen geometrically invalid workspace targets prior to the GPU-batched MPPI stage. Since each bait $\mathbf{b}^{(j)}$ specifies only a workspace position in $\mathbb{R}^{d}$, pointwise validity alone does not guarantee collision-free placement of the robot footprint. Let $\mathcal{A}_{\mathrm{b}}(\mathbf{b}^{(j)})$ denote the robot footprint placed at $\mathbf{b}^{(j)}$ under the prescribed bait-screening orientation. The bait $\mathbf{b}^{(j)}$ is then considered valid only if its footprint $\mathcal{A}_{\mathrm{b}}(\mathbf{b}^{(j)})$ lies entirely within the workspace $\mathcal{W}$ and does not intersect the obstacle region $\mathcal{O}$, i.e.,
\begin{equation}
\label{eq:bait_validity}
    \mathcal{A}_{\mathrm{b}}\!\left(\mathbf{b}^{(j)}\right)
    \subseteq
    \mathcal{W},
    \qquad
    \mathcal{A}_{\mathrm{b}}\!\left(\mathbf{b}^{(j)}\right)
    \cap
    \mathcal{O}
    =
    \emptyset.
\end{equation}
A bait that violates either condition is radially retracted toward the workspace position $\mathbf{p}_{s}$ of the selected node $n_{\mathrm{s}}$ according to
\begin{equation}
\label{eq:bait_retraction}
    \widetilde{\mathbf{b}}^{(j)}
    =
    \mathbf{p}_{s}
    +
    \eta
    \bigl(
        \mathbf{b}^{(j)}-\mathbf{p}_{s}
    \bigr),
\end{equation}
where $\eta\in(0,1)$ is the retraction factor. This radial contraction preserves the original exploration direction while reducing the bait's distance from $\mathbf{p}_{s}$. The retracted bait $\widetilde{\mathbf{b}}^{(j)}$ is then reevaluated under the validity conditions in~\eqref{eq:bait_validity}. If valid, it replaces the original bait $\mathbf{b}^{(j)}$; otherwise, the bait is discarded before GPU-batched MPPI optimization.
For notational simplicity, all baits retained after this screening, including valid unretracted baits and valid retracted replacements, are reindexed and denoted by $\mathbf{b}^{(j)}$, forming the retained bait set
\begin{equation}
\label{eq:retained_bait_set}
    \mathcal{B}_{\mathrm{ret}}
    =
    \bigl\{\mathbf{b}^{(j)}\bigr\}_{j=1}^{B_{\mathrm{ret}}},
    \qquad
    B_{\mathrm{ret}}
    =
    \left|\mathcal{B}_{\mathrm{ret}}\right|
    \leq
    B,
\end{equation}
where $|\cdot|$ denotes the number of elements in a finite set.
The effective expansion radius associated with each retained bait is defined as
\begin{equation}
\label{eq:effective_bait_radius}
    r_{b,\mathrm{eff}}^{(j)}
    =
    \left\|
        \mathbf{b}^{(j)}-\mathbf{p}_{s}
    \right\|_{2},
    \qquad
    \forall j\in\{1,\ldots,B_{\mathrm{ret}}\}.
\end{equation}
Accordingly, $r_{b,\mathrm{eff}}^{(j)}=r_b$ for an unretracted bait, whereas $r_{b,\mathrm{eff}}^{(j)}<r_b$ for a retained bait that has undergone retraction. Only the $B_{\mathrm{ret}}$ elements of $\mathcal{B}_{\mathrm{ret}}$ are supplied to the GPU-batched MPPI step. The progress criterion introduced in Section~\ref{subsec:filtering} is evaluated relative to $r_{b,\mathrm{eff}}^{(j)}$, rather than the nominal shell radius $r_b$.

\subsubsection{GPU-Batched MPPI Optimization}\label{sec:GPU-Batched-MPPI-Optimization}

For each retained bait $\mathbf{b}^{(j)}\in\mathcal{B}_{\mathrm{ret}}$, we formulate a local finite-horizon optimal-control problem of the form~\eqref{eq:finite_horizon_problem}, initialized at $\mathbf{x}_{0}^{(j)}=\mathbf{x}_{s}$. Accordingly, each retained bait is augmented with the task-relevant state components to form a local bait state $\mathbf{x}_{\mathrm{b}}^{(j)}\in\mathcal{X}$. These components may include, where applicable, a target orientation selected to promote progress toward the goal region $\mathcal{X}_{\mathrm{goal}}$ while accounting for local obstacle clearance. MPPI, as reviewed in Section~\ref{subsec:mppi_overview}, then computes an approximate solution to each bait-conditioned problem and returns an optimized control sequence $\mathbf{U}^{(j)}$. Forward propagation of this sequence through the dynamics in~\eqref{eq:dynamics_system} generates the corresponding state trajectory $\tau^{(j)}$. Together with the associated candidate cost $J^{(j)}$, these quantities define the \emph{potential candidate} $c^{(j)}$ as
\begin{equation}
\label{eq:candidate_tuple}
    c^{(j)}
    =
    \bigl(
        \mathbf{U}^{(j)},
        \,\tau^{(j)},
        \,J^{(j)}
    \bigr).
\end{equation}
Following the instantaneous running-cost decomposition in~\eqref{eq:mppi_running_cost}, each candidate is reevaluated by
deterministically propagating its corresponding control sequence, yielding the candidate cost $J^{(j)}$ as
\begin{equation}
\label{eq:bait_cost}
    J^{(j)}
    =
    \underbrace{
        \sum_{k=0}^{N-1}
        \widetilde{q}\!\left(
            \mathbf{x}_{k}^{(j)},
            \mathbf{u}_{k}^{(j)},
            \mathbf{0};
            \mathbf{x}_{\mathrm{b}}^{(j)}
        \right)
    }_{\mbox{\scriptsize Accumulated running cost}}
    +
    \underbrace{
        \phi_{\mathrm{term}}\!\left(
            \mathbf{x}_{N}^{(j)}
        \right)
    }_{\mbox{\scriptsize Terminal cost}}.
\end{equation}
Since no control perturbation is applied during candidate reevaluation, $\delta\mathbf{u}_{k}=\mathbf{0}$, and the quadratic control cost $q_{\mathbf{u}}$ in~\eqref{eq:mppi_control_cost} reduces to
\[
    q_{\mathbf{u}}
    \!\left(
        \mathbf{u}_{k}^{(j)},
        \mathbf{0}
    \right)
    =
    \frac{1}{2}
    \left(
        \mathbf{u}_{k}^{(j)}
    \right)^{\top}
    \mathbf{R}
    \mathbf{u}_{k}^{(j)}.
\]
Consequently, the accumulated term in~\eqref{eq:bait_cost} retains the task-dependent state cost and nominal control-effort penalty of the MPPI formulation, while the perturbation-dependent terms vanish. The resulting accumulated running cost therefore provides a local, bait-conditioned evaluation of each candidate. However, since the retained baits span distinct local directions, two candidates may incur similar running costs while terminating at substantially different distances from the goal state $\mathbf{x}_{\mathrm{goal}}$. The running cost alone may therefore provide insufficient discrimination with respect to mission-level progress. Accordingly, we introduce the terminal cost $\phi_{\mathrm{term}}(\cdot)$ as
\begin{equation}
\label{eq:term_cost}
    \phi_{\mathrm{term}}\!\left(
        \mathbf{x}_{N}^{(j)}
    \right)
    =
    \alpha
    \left\|
        \mathbf{p}_{N}^{(j)}
        -
        \mathbf{p}_{\mathrm{goal}}
    \right\|_{2},
\end{equation}
where $\alpha>0$ is the terminal-cost weighting coefficient, $\mathbf{p}_{N}^{(j)}\in\mathcal{W}$ is the workspace position associated with the terminal state $\mathbf{x}_{N}^{(j)}$, and $\mathbf{p}_{\mathrm{goal}}\in\mathcal{W}$ is the workspace position associated with the goal state $\mathbf{x}_{\mathrm{goal}}$. 
Separating the local trajectory evaluation provided by the accumulated running cost from the goal-progress evaluation provided by $\phi_{\mathrm{term}}(\cdot)$ allows candidates to be compared and ranked using the complete candidate cost $J^{(j)}$ according to both their local trajectory quality and their contribution to mission-level progress.

Beyond candidate-cost evaluation, two complementary design features contribute to the computational efficiency of this stage. 
First, unlike planners that rely on an exact steering operator to enforce terminal-state equality with a prescribed successor state, often through the solution of a BVP, the proposed expansion treats each bait state $\mathbf{x}_{\mathrm{b}}^{(j)}$ as a soft local reference in the running cost $\widetilde{q}(\cdot;\mathbf{x}_{\mathrm{b}}^{(j)})$ in~\eqref{eq:bait_cost}, without imposing the hard terminal constraint $\mathbf{x}_{N}^{(j)}=\mathbf{x}_{\mathrm{b}}^{(j)}$. In this forward-shooting formulation, the optimized control sequence $\mathbf{U}^{(j)}$ is applied from $\mathbf{x}_{s}$ under the system dynamics in~\eqref{eq:dynamics_system}, generating the candidate trajectory $\tau^{(j)}$. Consequently, the terminal state $\mathbf{x}_{N}^{(j)}$, which serves as the prospective child-node state (Section~\ref{subsec:filtering}), is determined by forward propagation rather than prescribed as a boundary condition. Such an approach avoids solving a separate BVP for each prospective tree edge, which can be computationally demanding for nonlinear systems subject to state and control constraints.
Second, to exploit the inherent parallelism of MPPI sampling and the high-throughput capabilities of GPUs, the $B_{\mathrm{ret}}$ bait-conditioned problems associated with the retained bait set $\mathcal{B}_{\mathrm{ret}}$ in~\eqref{eq:retained_bait_set} are solved concurrently by GPU-batched MPPI rather than through sequential per-bait optimization. At each MPPI refinement iteration, the sampled rollouts associated with all retained baits are evaluated within a common GPU batch, thereby consolidating host--device data transfers and reducing CPU--GPU synchronization overhead.
Together with the adaptive bait-handling rule in~\eqref{eq:bait_retraction}, which removes geometrically invalid targets before candidate generation, this GPU-batched expansion reduces the computational overhead of each tree expansion and improves GPU utilization. 
Collectively, these mechanisms support fast planning in large-scale, cluttered environments, as demonstrated in Section~\ref{sec:results}.

At this stage, the resulting $B_{\mathrm{ret}}$ potential candidates form the set
\begin{equation}
\label{eq:candidate_set}
    \mathcal{C}
    =
    \bigl\{
        c^{(j)}
    \bigr\}_{j=1}^{B_{\mathrm{ret}}},
\end{equation}
where each $c^{(j)}$ remains a transient candidate until the following candidate-admission stage, in which explicit feasibility screening and cost-based retention determine whether its associated motion segment is incorporated into $\mathcal{T}$. In this manner, the fused
expansion--simulation stage produces cost-informed, dynamically feasible motion candidates rather than random default-policy rollouts.

\subsection{Candidate Admission and Tree Integration}
\label{subsec:filtering}

Given the transient candidate set $\mathcal{C}$ generated by the preceding expansion stage and defined in~\eqref{eq:candidate_set}, the candidate-admission stage aims to determine which motion segments are eligible for incorporation into $\mathcal{T}$, thereby maintaining a sparse, physically valid search tree and directing subsequent growth toward promising branches. To this end, we process $\mathcal{C}$ through two sequential operations: feasibility gating, which maps $\mathcal{C}$ to an accepted subset $\mathcal{C}_{\mathrm{Acc}}$, followed by cost-based top-$k$ selection, which produces the subset $\mathcal{C}_{k}$ for tree integration.

\subsubsection{Feasibility Gating}

Each potential candidate $c^{(j)}\in\mathcal{C}$, as defined in~\eqref{eq:candidate_tuple}, is evaluated based on its induced trajectory $\tau^{(j)}$. As established in Section~\ref{subsec:expansion}, $\tau^{(j)}$ is obtained by propagating the system dynamics from $\mathbf{x}_{s}$ under the optimized control sequence $\mathbf{U}^{(j)}$, with $\mathbf{u}_{k}^{(j)}\in\mathcal{U}$ for all $k\in\{0,\ldots,N-1\}$. Candidate generation therefore establishes consistency with the system dynamics and admissibility of the applied controls. The feasibility gate subsequently verifies collision safety and significant progress before the associated motion segment becomes eligible for integration into $\mathcal{T}$.

\textit{Collision Safety:}
Although MPPI approximates the local finite-horizon problem in~\eqref{eq:finite_horizon_problem} using an objective function that penalizes collisions with obstacles, the collision-avoidance constraint $\mathbf{x}_{k}^{(j)}\in\mathcal{X}_{\mathrm{free}}$ specified in~\eqref{eq:finite_horizon_constraints} is not imposed as a hard constraint during optimization. Consequently, the resulting candidate trajectory $\tau^{(j)}$ is not guaranteed to be collision-free and is subjected to a deterministic post-optimization collision check.
Specifically, the collision checker applies the conditions in~\eqref{eq:bait_validity} to the state-dependent robot footprint $\mathcal{A}(\mathbf{x}_{k}^{(j)})$ at selected states of $\tau^{(j)}$, i.e.,
\begin{equation}
\label{eq:candidate_collision_safety}
    \mathcal{A}\!\left(\mathbf{x}_{k}^{(j)}\right)
    \subseteq
    \mathcal{W},
    \qquad
    \mathcal{A}\!\left(\mathbf{x}_{k}^{(j)}\right)
    \cap
    \mathcal{O}
    =
    \emptyset,
    \qquad
    \forall k\in\mathcal{K}_{\mathrm{chk}},
    \end{equation}
where $\mathcal{K}_{\mathrm{chk}}\subseteq\{0,\ldots,N\}$ denotes the set of trajectory indices selected for collision checking. For a prescribed collision-checking stride $s_{\mathrm{chk}}\in\mathbb{N}_{>0}$, this set is defined as 
\begin{equation}
\label{eq:collision_check_indices}
    \mathcal{K}_{\mathrm{chk}}
    =
    \left\{
        m s_{\mathrm{chk}}
        \;\middle|\;
        m\in\mathbb{N}_{0},
        \;
        m s_{\mathrm{chk}}\leq N
    \right\}
    \cup
    \{N\},
\end{equation}
so that the terminal state $\mathbf{x}_{N}^{(j)}$ is always verified independently of the selected stride.

\textit{Significant Progress:}
In addition to satisfying~\eqref{eq:candidate_collision_safety}, a candidate must demonstrate significant progress from the selected node $n_{\mathrm{s}}$ to prevent the insertion of stationary or negligible motion segments. We first evaluate this requirement using the terminal displacement normalized by the effective expansion radius $r_{b,\mathrm{eff}}^{(j)}$ of the assigned bait according to
\begin{equation}
\label{eq:candidate_progress}
    \rho^{(j)}
    =
    \frac{
        \left\|
            \mathbf{p}_{N}^{(j)}
            -
            \mathbf{p}_{s}
        \right\|_{2}
    }{
        r_{b,\mathrm{eff}}^{(j)}
    }
    \geq
    \eta_{\mathrm{dist}},
\end{equation}
where $\eta_{\mathrm{dist}}\in(0,1]$ denotes the minimum accepted fraction of $r_{b,\mathrm{eff}}^{(j)}$. This displacement-based condition serves as the primary progress criterion, while the clearance-gain and heading-alignment tests introduced below provide optional supplementary criteria. Using $r_{b,\mathrm{eff}}^{(j)}$ rather than the nominal shell radius $r_b$ ensures that the normalized progress criterion in~\eqref{eq:candidate_progress} is applied consistently to both unretracted and retracted baits.

\textit{Supplementary Progress Criteria:}
In cluttered regions, obstacle geometry may limit the translational displacement attainable by a candidate, even when its terminal state provides improved clearance or a more favorable orientation for subsequent expansion. To retain such locally useful motion segments when the displacement condition in~\eqref{eq:candidate_progress} is not satisfied, the feasibility gate can additionally evaluate clearance-gain and heading-alignment tests.
When enabled, the clearance-gain test compares the obstacle clearance of the robot placement at the candidate endpoint with that at the selected node $n_{\mathrm{s}}$. 
A candidate satisfies the clearance-gain criterion if the obstacle clearance at its terminal position $\mathbf{p}_{N}^{(j)}$ exceeds that at the selected-node position $\mathbf{p}_{s}$ by at least a prescribed margin $\Delta_{\mathrm{clear}}>0$.
The clearance values can be obtained, for example, from a Euclidean distance field (EDF) computed from the occupancy map via an exact Euclidean distance transform~\cite{maurer2003linear}.
Alternative geometric clearance measures may be used when an EDF is unavailable.
Similarly, the heading-alignment test evaluates whether the terminal orientation reduces the angular error relative to the target orientation encoded in $\mathbf{x}_{\mathrm{b}}^{(j)}$, as defined in the expansion stage.
This criterion discourages retrograde or poorly aligned terminal orientations while retaining candidates that establish a more favorable orientation for subsequent bait-guided expansions, even when their displacement under~\eqref{eq:candidate_progress} is limited.

Let $\operatorname{Prog}(c^{(j)})$ denote the resulting progress decision. This condition is satisfied when the normalized-displacement criterion in~\eqref{eq:candidate_progress} is met or when at least one enabled supplementary criterion is satisfied. Regardless of the progress criterion used, collision safety in~\eqref{eq:candidate_collision_safety} remains mandatory.
Accordingly, the accepted candidate set is defined as
\begin{equation}
\label{eq:accepted_set}
    \mathcal{C}_{\mathrm{Acc}}
    =
    \left\{
        c^{(j)}\in\mathcal{C}
        \;\middle|\;
        \operatorname{Safe}\!\left(\tau^{(j)}\right)
        \land
        \operatorname{Prog}\!\left(c^{(j)}\right)
    \right\},
\end{equation}
where $\operatorname{Safe}(\tau^{(j)})$ denotes satisfaction of~\eqref{eq:candidate_collision_safety}. Candidates that fail either requirement are discarded. If $\mathcal{C}_{\mathrm{Acc}}=\emptyset$, no child node is added from the current expansion, and the selected node $n_{\mathrm{s}}$ is treated as a dead end.

\subsubsection{Top-$k$ Selection}

Following feasibility gating, top-$k$ selection restricts tree insertion to a limited number of low-cost candidates. Without this restriction, admitting every candidate in $\mathcal{C}_{\mathrm{Acc}}$ would create a wide set of sibling branches, including candidates with comparatively high costs. 
Moreover, since UCB-based selection assigns maximal priority to unvisited children, the search would initially allocate effort to each newly inserted sibling before revisiting previously explored branches. 
To control this breadth-oriented growth, we retain at most $k_b$ accepted candidates. Accordingly, for a nonempty accepted candidate set $\mathcal{C}_{\mathrm{Acc}}$ (i.e., $\mathcal{C}_{\mathrm{Acc}}\neq\emptyset$), we define the effective branching factor $k_{\mathrm{eff}}\in\mathbb{N}_{>0}$ as
\begin{equation}
\label{eq:effective_topk}
    k_{\mathrm{eff}}
    =
    \min
    \left\{
        k_b,
        \left|
            \mathcal{C}_{\mathrm{Acc}}
        \right|
    \right\},
\end{equation}
where $k_b\in\mathbb{N}_{>0}$ is the branching limit. The retained candidate set $\mathcal{C}_{k}$ is selected according to
\begin{equation}
\label{eq:topk_set}
    \mathcal{C}_{k}
    \in
    \operatorname*{arg\,min}_{
        \substack{
            \mathcal{S}
            \subseteq
            \mathcal{C}_{\mathrm{Acc}}
            \\
            |\mathcal{S}|
            =
            k_{\mathrm{eff}}
        }
    }
    \;
    \sum_{c^{(j)}\in\mathcal{S}}
    J^{(j)},
\end{equation}
where the minimization is taken over all subsets $\mathcal{S}\subseteq\mathcal{C}_{\mathrm{Acc}}$ satisfying $|\mathcal{S}|=k_{\mathrm{eff}}$.
Accordingly, $\mathcal{C}_{k}$ contains $k_{\mathrm{eff}}\leq k_b$ accepted candidates with the lowest costs. For each $c^{(j)}\in\mathcal{C}_{k}$, the terminal state $\mathbf{x}_{N}^{(j)}$ is added as a child node $n_j$ of $n_{\mathrm{s}}$, while the associated pair $(\mathbf{U}^{(j)},\tau^{(j)})$ is stored on the corresponding incoming edge, consistently with the tree representation introduced above. At insertion, the incoming-edge cost and initial node statistics are assigned as
\begin{equation}
\label{eq:edge_reward_initialization}
    J_j^{\mathrm{edge}}
    =
    J^{(j)},
    \qquad
    R_j
    =
    -J_j^{\mathrm{edge}},
    \qquad
    V_j
    =
    0.
\end{equation}

Since all $B_{\mathrm{ret}}$ potential candidates have already been generated and cost-evaluated within the same GPU batch, top-$k$ selection is performed directly on the costs of the accepted candidates without additional trajectory optimization.
This rule decouples the number of candidates evaluated in parallel, $B_{\mathrm{ret}}$, from the number of branches permanently introduced into the tree, $k_{\mathrm{eff}}$. Consequently, it controls the branching factor, limits memory growth and subsequent UCB-selection overhead, and promotes asymmetric tree growth toward lower-cost branches. 
At the same time, choosing $k_b>1$ allows the search to retain multiple low-cost candidates associated with different bait directions, thereby preserving alternative local expansion directions for subsequent tree exploration.
The branching limit $k_b$ may either be held fixed at a prescribed value throughout the search (e.g., $k_b=4$) or, in variants that permit repeated expansion of a node, adapted online as a function of that node's visit count using progressive widening~\cite{lee2020monte}. In either case, \eqref{eq:effective_topk} ensures that the selection remains well-defined when fewer than $k_b$ candidates pass feasibility gating.
\subsection{Proposal Adaptation and Backpropagation}
\label{subsec:backprop}

Following feasibility gating and top-$k$ selection, our proposed framework uses the candidate-cost statistics to perform two complementary updates. First, the normalized candidate weights and optimized control sequences are used to adapt the MPPI nominal control mean, thereby providing a warm start for subsequent expansions. Second, an expansion-level scalar backup cost formed from the candidates retained by top-$k$ selection, $\mathcal{C}_{k}$, is backpropagated along the selected tree path to update the visit counts and edge-cost-seeded reward statistics used by UCB selection. The following subsections define the statistical weighting, control-mean adaptation, and cost-based backpropagation operations.

\subsubsection{Cost Normalization and Statistical Weighting}

Let $B_{\mathrm{Acc}}=|\mathcal{C}_{\mathrm{Acc}}|$ denote the number of accepted candidates, and assume $\mathcal{C}_{\mathrm{Acc}}\neq\emptyset$. We first define the minimum and maximum costs within the accepted candidate set as
\begin{equation}
\label{eq:accepted_cost_extrema}
    J_{\min}
    =
    \min_{c^{(j)}\in\mathcal{C}_{\mathrm{Acc}}}
    J^{(j)},
    \qquad
    J_{\max}
    =
    \max_{c^{(j)}\in\mathcal{C}_{\mathrm{Acc}}}
    J^{(j)}.
\end{equation}
To improve numerical stability across expansion steps, we apply \textit{min--max normalization} to the accepted candidate costs before computing their corresponding weights, as direct exponential weighting of raw costs with substantially different magnitudes and spreads may lead to numerical underflow and excessive concentration of the weight distribution. Using the cost extrema $J_{\min}$ and $J_{\max}$ defined in~\eqref{eq:accepted_cost_extrema}, the normalized cost associated with each candidate $c^{(j)}\in\mathcal{C}_{\mathrm{Acc}}$ is then computed as
\begin{equation}
\label{eq:norm_cost}
    \widetilde{J}^{(j)}
    =
    \frac{
        J^{(j)}-J_{\min}
    }{
        J_{\max}-J_{\min}+\varepsilon
    },
\end{equation}
where $\varepsilon>0$ is a small constant that prevents division by zero when the accepted candidate costs are identical and improves numerical stability when their range is very small. The corresponding normalized candidate weights are then computed as
\begin{equation}
\label{eq:softmax_weights}
    w^{(j)}
    =
    \frac{
        \exp\!\left(
            -\widetilde{J}^{(j)}/\beta_{\mathrm{w}}
        \right)
    }{
        \displaystyle
        \sum_{c^{(\ell)}\in\mathcal{C}_{\mathrm{Acc}}}
        \exp\!\left(
            -\widetilde{J}^{(\ell)}/\beta_{\mathrm{w}}
        \right)
    },
    \qquad
    c^{(j)}\in\mathcal{C}_{\mathrm{Acc}},
\end{equation}
where $\beta_{\mathrm{w}}>0$, distinct from the MPPI temperature $\lambda$ in~\eqref{eq:mppi_weights}, is the candidate-level weighting temperature that controls selectivity. Specifically, smaller values of $\beta_{\mathrm{w}}$ concentrate the weights on lower-cost candidates, whereas larger values produce a more uniform distribution. By construction, the weights satisfy
$
    \sum_{c^{(j)}\in\mathcal{C}_{\mathrm{Acc}}}
    w^{(j)}
    =
    1,
$
with lower-cost candidates receiving larger weights. Moreover, the min--max normalization in~\eqref{eq:norm_cost} makes the weighting depend on relative cost differences within the current expansion rather than on the absolute cost scale.

\subsubsection{ESS-Guided Control-Mean Adaptation}
\label{subsubsec:control_update}

As reviewed in Section~\ref{subsec:mppi_overview}, MPPI samples control perturbations $\delta\mathbf{u}_{k}$ about a nominal control sequence $\mathbf{U}=\{\mathbf{u}_{k}\}_{k=0}^{N-1}$ under the exploration covariance $\boldsymbol{\Sigma}_{\mathbf{u}}$ and iteratively refines $\mathbf{U}$ according to~\eqref{eq:mppi_optimal-control}. In a conventional receding-horizon implementation, $\mathbf{U}$ is typically initialized using a default nominal sequence, commonly a zero-control sequence when no prior solution is available, and then warm-started across successive control cycles. Specifically, after the first control input $\mathbf{u}_{0}$ is applied to the robot, the remaining optimized controls are shifted forward and reused to initialize the next optimization, thereby exploiting the temporal continuity between successive closed-loop states.
In the proposed framework, by contrast, MPPI is employed as a local trajectory optimizer at each selected node $n_{\mathrm{s}}$, rather than as the closed-loop controller that directly commands the robot. Starting from the associated state $\mathbf{x}_{s}$, the batched optimizer executes $N_{\mathrm{opt}}$ refinement iterations over the local optimization problems induced by the retained bait set $\mathcal{B}_{\mathrm{ret}}$, where $N_{\mathrm{opt}}\in\mathbb{N}_{>0}$ specifies the number of MPPI refinement iterations allocated to each tree expansion.

In this context, the MPPI-based tree-expansion setting differs from conventional receding-horizon execution in two important respects. First, the optimized control sequences generated at $n_{\mathrm{s}}$ are not immediately applied to the robot. Consequently, neither an executed first control input nor a subsequent closed-loop state is available for constructing the usual shifted-horizon initialization. Second, all retained-bait optimizations are solved concurrently within a common GPU batch. Although sequential per-bait optimization could initialize each local problem using the solution obtained for the preceding bait, such an approach would introduce an arbitrary dependence on bait ordering and serialize the computation, thereby eliminating the computational benefits of batched execution discussed in Section~\ref{sec:GPU-Batched-MPPI-Optimization}.
Accordingly, the proposed framework constructs a common warm-start sequence $\mathbf{U}_{\mathrm{ws}}$ using the optimized control sequences and cost-derived statistics of the accepted candidates $\mathcal{C}_{\mathrm{Acc}}$, and supplies it simultaneously to all retained-bait optimizations in the subsequent expansion. When no transferred control information is available, as in the initial tree expansion, $\mathbf{U}_{\mathrm{ws}}$ is initialized using the default nominal sequence. Thereafter, the adapted warm start shifts the nominal control sequence toward patterns associated with low-cost motion segments that advance toward the goal, while the fixed exploration covariance $\boldsymbol{\Sigma}_{\mathbf{u}}$ preserves stochastic exploration. 
Compared with repeated initialization from the default nominal sequence, the transferred control information provides a more informative warm start, enabling MPPI to generate higher-quality trajectories with a smaller refinement budget $N_{\mathrm{opt}}$. 
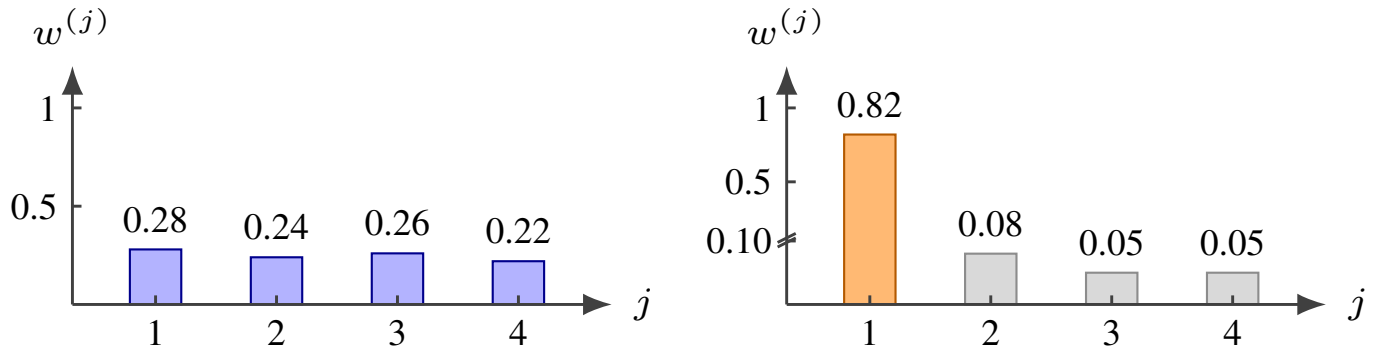
\begin{figure}[!t]
    \centering
    \begin{subfigure}[t]{0.48\columnwidth}
        \centering
        \input{figures/ess_update_regimes_a.tikz}
        \vspace*{-9pt}
        \caption{Nearly uniform weights (high $\Ebar$; moment-matching regime).}
        \label{fig:ess_update_regimes_a}
    \end{subfigure}
    \hfill
    \begin{subfigure}[t]{0.48\columnwidth}
        \centering
        \input{figures/ess_update_regimes_b.tikz}
        \vspace*{-9pt}
        \caption{Single dominant weight (low $\Ebar$; winner-step regime).}
        \label{fig:ess_update_regimes_b}
    \end{subfigure}
    \vspace*{-5pt}
    \caption{Illustration of the ESS-guided mean-update regimes for an accepted candidate set with $B_{\mathrm{Acc}}=4$. In~(a), the nearly uniform weights yield $\operatorname{ESS}\approx3.97$ and $\Ebar\approx0.99$, illustrating the moment-matching regime, in which information from multiple accepted candidates is aggregated. In~(b), a single dominant weight yields $\operatorname{ESS}\approx1.46$ and $\Ebar\approx0.37$, illustrating the winner-step regime, in which the nominal control mean is updated toward the corresponding control sequence. The broken vertical axis in~(b) magnifies the interval below $0.10$ to improve the visibility of the smaller weights.}
    \label{fig:ess_update_regimes}
\end{figure}

To formalize this adaptation, let $\boldsymbol{\mu}_{k}^{-}\in\mathbb{R}^{n_u}$ and $\boldsymbol{\mu}_{k}^{+}\in\mathbb{R}^{n_u}$ denote the nominal control means before and after the current update, respectively, at discrete time index $k\in\{0,\ldots,N-1\}$. To determine whether the mean update should aggregate information from several accepted candidates or rely on a dominant low-cost candidate, we characterize the distribution of the normalized candidate weights in~\eqref{eq:softmax_weights} using the ESS~\cite{kong1994sequential}. Specifically, the ESS and its normalized form $\Ebar$ are defined as
\begin{equation}
\label{eq:ess_measure}
    \operatorname{ESS}
    =
    \frac{1}{
        \displaystyle
        \sum_{c^{(j)}\in\mathcal{C}_{\mathrm{Acc}}}
        \left(w^{(j)}\right)^2
    },
    \quad
    \Ebar
    =
    \frac{\operatorname{ESS}}{B_{\mathrm{Acc}}}
    \in
    \left[
        \frac{1}{B_{\mathrm{Acc}}},
        1
    \right].
\end{equation}
A value of $\Ebar$ near one indicates that the weights are nearly uniform  across the accepted candidates, as illustrated in Fig.~\ref{fig:ess_update_regimes_a}, whereas smaller values indicate increasing dominance by one or a few candidates, as illustrated in Fig.~\ref{fig:ess_update_regimes_b}. To supplement this distribution-level measure, we define the largest candidate weight as
\begin{equation}
\label{eq:max_candidate_weight}
    w_{\max}
    =
    \max_{c^{(j)}\in\mathcal{C}_{\mathrm{Acc}}}
    w^{(j)}.
\end{equation}
Together, $\Ebar$ and $w_{\max}$ indicate whether the candidate weights are distributed across several accepted candidates or dominated by a single low-cost candidate. Accordingly, we employ a hybrid mean-update mechanism comprising two regimes: a \textit{moment-matching} regime that aggregates information from multiple accepted candidates and a \textit{winner-step} regime that updates the nominal mean toward the control sequence of the dominant candidate. To select between these regimes, let $\eta_{\mathrm{ess}}\in(0,1]$ and $\eta_{w}\in(0,1]$ denote the normalized-ESS and maximum-weight thresholds, respectively. The update mode is then determined according to
\begin{equation}
\label{eq:proposal_update_mode}
    \mathsf{mode}
    =
    \begin{cases}
        \mathrm{winner},
        &
        \Ebar
        \leq
        \eta_{\mathrm{ess}}
        \ \text{or}\
        w_{\max}
        \geq
        \eta_{w},
        \\[2mm]
        \mathrm{moment},
        &
        \text{otherwise}.
    \end{cases}
\end{equation}

\paragraph{Moment-Matching Regime}
When $\mathsf{mode}=\mathrm{moment}$, no single candidate dominates the weight distribution, as illustrated in Fig.~\ref{fig:ess_update_regimes_a}. Accordingly, all accepted candidates contribute to the cost-weighted control mean 
\begin{equation}
\label{eq:weighted_control_mean}
    \overline{\mathbf{u}}_{k}
    =
    \sum_{c^{(j)}\in\mathcal{C}_{\mathrm{Acc}}}
    w^{(j)}
    \mathbf{u}_{k}^{(j)},
    \qquad
    k\in\{0,\ldots,N-1\}.
\end{equation}
The nominal control mean is then updated by exponential smoothing according to
\begin{equation}
\label{eq:moment_mean_update}
    \boldsymbol{\mu}_{k}^{+}
    =
    \left(1-\alpha_{\mu}\right)
    \boldsymbol{\mu}_{k}^{-}
    +
    \alpha_{\mu}
    \overline{\mathbf{u}}_{k},
    \qquad
    \alpha_{\mu}\in(0,1].
\end{equation}
Here, $\alpha_{\mu}$ controls the influence of the cost-weighted control mean $\overline{\mathbf{u}}_{k}$ on the updated nominal mean $\boldsymbol{\mu}_{k}^{+}$. This moment-matching regime combines control information from multiple accepted candidates, assigning greater influence to lower-cost candidates while moderating changes relative to the prior mean $\boldsymbol{\mu}_{k}^{-}$. The update in~\eqref{eq:moment_mean_update} can be interpreted as a cost-weighted moment-matching step related to those used in Cross-Entropy Method (CEM) formulations~\cite{de2005tutorial}.

\paragraph{Winner-Step Regime}
When $\mathsf{mode}=\mathrm{winner}$, the accepted-candidate weight distribution favors a dominant low-cost candidate, as illustrated in Fig.~\ref{fig:ess_update_regimes_b}. In this case, averaging over all accepted candidates may attenuate the corresponding low-cost control pattern. We therefore identify the minimum-cost candidate among those retained by top-$k$ selection as
\begin{equation}
\label{eq:winner_candidate}
    j^{\star}
    \in
    \operatorname*{arg\,min}_{
        c^{(j)}\in\mathcal{C}_{k}
    }
    J^{(j)}.
\end{equation}
The nominal control mean is then updated toward the optimized control sequence of the winning candidate according to
\begin{equation}
\label{eq:winner_mean_update}
    \boldsymbol{\mu}_{k}^{+}
    =
    \boldsymbol{\mu}_{k}^{-}
    +
    \eta_{\mu}
    \left(
        \mathbf{u}_{k}^{(j^{\star})}
        -
        \boldsymbol{\mu}_{k}^{-}
    \right),
    \qquad
    \eta_{\mu}\in(0,1].
\end{equation}
Here, $\eta_{\mu}$ controls the influence of the winning control $\mathbf{u}_{k}^{(j^{\star})}$ on the updated nominal mean $\boldsymbol{\mu}_{k}^{+}$. This winner-step regime preserves the dominant low-cost control pattern rather than averaging it with less competitive candidates, while limiting the deviation from the prior mean $\boldsymbol{\mu}_{k}^{-}$.

\paragraph{Warm-Start Construction}
After either update regime, the updated nominal means are assembled into the full-horizon sequence
\[
    \boldsymbol{\mu}_{0:N-1}^{+}
    =
    \begin{bmatrix}
        \boldsymbol{\mu}_{0}^{+}
        &
        \cdots
        &
        \boldsymbol{\mu}_{N-1}^{+}
    \end{bmatrix}^{\top}
    \in
    \mathbb{R}^{N\times n_u}.
\]
The warm-start sequence $\mathbf{U}_{\mathrm{ws}}$ supplied to the next MPPI expansion can be constructed in one of two forms. Specifically, the \emph{full-sequence mode} retains the complete time-indexed mean sequence, whereas the \emph{horizon-averaged mode} first computes its temporal average according to
\begin{equation}
\label{eq:horizon_averaged_mean}
    \overline{\boldsymbol{\mu}}^{+}
    =
    \frac{1}{N}
    \sum_{k=0}^{N-1}
    \boldsymbol{\mu}_{k}^{+},
    \qquad
    \overline{\boldsymbol{\mu}}^{+}
    \in
    \mathbb{R}^{n_u}.
\end{equation}
Accordingly, the warm-start sequence is defined as
\begin{equation}
\label{eq:warmstart_modes}
    \mathbf{U}_{\mathrm{ws}}
    =
    \begin{cases}
        \boldsymbol{\mu}_{0:N-1}^{+},
        &
        \text{full-sequence mode},
        \\[1mm]
        \mathbf{1}_{N}
        \left(
            \overline{\boldsymbol{\mu}}^{+}
        \right)^{\top},
        &
        \text{horizon-averaged mode},
    \end{cases}
\end{equation}
where $\mathbf{1}_{N}\in\mathbb{R}^{N}$ denotes a vector of ones.
The two modes preserve different forms of information. The full-sequence mode transfers the complete temporal control structure learned during the preceding expansion and may therefore be advantageous when consecutive expansions have similar initial states and local objectives. However, its time-indexed structure may also encode timing information specific to the preceding selected state and bait geometry, which may not transfer effectively to a new expansion. By contrast, the horizon-averaged mode retains the aggregate low-cost control tendency while removing this expansion-specific timing information. 
Replicating $\overline{\boldsymbol{\mu}}^{+}$ across the prediction horizon therefore provides a common, stage-independent initialization from which each subsequent local MPPI optimization can construct a target-specific, time-varying control sequence through iterative refinement.
For this reason, the horizon-averaged mode is used by default, while the full-sequence mode remains available as an alternative. In either mode, $\mathbf{U}_{\mathrm{ws}}$ initializes the iterative MPPI update in~\eqref{eq:mppi_optimal-control}, while the exploration covariance $\boldsymbol{\Sigma}_{\mathbf{u}}$ remains fixed.


\subsubsection{Cost-Based Backpropagation}
\label{subsubsec:backprop}
Following control-mean adaptation, the costs of the candidates retained by top-$k$ selection are used to form an expansion-level scalar backup cost that is backpropagated along the selected tree path toward the root, thereby updating the visit counts and edge-cost-seeded reward statistics used by the UCB selection policy. To construct this backup cost, we consider two backpropagation modes: an \emph{expected-cost mode}, which aggregates the costs of all retained candidates, and a \emph{winner-cost mode}, which uses only the cost of the minimum-cost retained candidate. 

\paragraph{Expected-Cost Mode}
The candidate weights $w^{(j)}$ in~\eqref{eq:softmax_weights} are normalized over the accepted set $\mathcal{C}_{\mathrm{Acc}}$, whereas the expected-cost statistic is formed only from the retained set $\mathcal{C}_{k}$. Accordingly, for $\mathcal{C}_{k}\neq\emptyset$, the corresponding weights are renormalized over $\mathcal{C}_{k}$ as
\begin{equation}
\label{eq:retained_candidate_weights}
    w_{\mathrm{ret}}^{(j)}
    =
    \frac{w^{(j)}}{\displaystyle\sum_{c^{(\ell)}\in\mathcal{C}_{k}} w^{(\ell)}},
    \qquad
    c^{(j)}\in\mathcal{C}_{k},
\end{equation}
so that $\sum_{c^{(j)}\in\mathcal{C}_{k}}w_{\mathrm{ret}}^{(j)}=1$. Using these weights, the expansion-level expected cost is computed at the selected parent $n_{\mathrm{s}}$ as
\begin{equation}
\label{eq:expected_cost}
    \widehat{J}
    =
    \sum_{c^{(j)}\in\mathcal{C}_{k}}
    w_{\mathrm{ret}}^{(j)}
    J^{(j)}.
\end{equation}
Thus, $\widehat{J}$ provides a parent-level summary of the relative costs of all candidates retained from the current expansion.

\paragraph{Winner-Cost Mode}
The winner-cost mode instead uses the cost of the minimum-cost candidate in $\mathcal{C}_{k}$,
\begin{equation}
\label{eq:winner_backup_cost}
    J_{\mathrm{win}}
    =
    \min_{c^{(j)}\in\mathcal{C}_{k}}
    J^{(j)}.
\end{equation}
Let $n_{+}$ denote the child node instantiated from the candidate attaining $J_{\mathrm{win}}$. Whereas $\widehat{J}$ summarizes the expansion rooted at $n_{\mathrm{s}}$, $J_{\mathrm{win}}$ is the individual cost assigned to the motion segment from $n_{\mathrm{s}}$ to $n_{+}$ and is therefore associated with that child rather than with the parent expansion as a whole.

\paragraph{Mode Selection and Tree Update}
Let $\mathsf{mode}_{\mathrm{bp}}\in\{\mathrm{expected},\mathrm{winner}\}$ denote the selected backpropagation mode. The scalar backup cost used in the current search iteration is then defined as
\begin{equation}
\label{eq:backpropagation_cost}
    J_{\mathrm{bp}}
    =
    \begin{cases} 
        \widehat{J},
        &
        \mathsf{mode}_{\mathrm{bp}}=\mathrm{expected},
        \\[1mm]
        J_{\mathrm{win}},
        &
        \mathsf{mode}_{\mathrm{bp}}=\mathrm{winner}.
    \end{cases}
\end{equation}
The two modes provide different summaries of the same retained set $\mathcal{C}_{k}$. Expected-cost backpropagation incorporates the relative costs of all retained candidates through the parent-level statistic $\widehat{J}$, whereas winner-cost backpropagation emphasizes the minimum-cost retained motion segment through the child-level cost $J_{\mathrm{win}}$. No general preference is imposed between these alternatives; $\mathsf{mode}_{\mathrm{bp}}$ is treated as a configurable algorithmic choice whose effect is assessed empirically in Section~\ref{subsubsec:ablation_analysis}.
Importantly, the backpropagation mode $\mathsf{mode}_{\mathrm{bp}}$ is selected independently of the ESS-guided control-mean regime in~\eqref{eq:proposal_update_mode}. The latter determines how the MPPI nominal mean is adapted for subsequent expansions, whereas~\eqref{eq:backpropagation_cost} selects the scalar cost used to update the tree statistics employed by UCB selection.

Let $\mathcal{P}_{+}$ denote the path from the root node $n_{0}$ to $n_{+}$, comprising the tree path selected during the current iteration and the newly inserted edge from $n_{\mathrm{s}}$ to $n_{+}$. For both backpropagation modes, the update is initiated from the minimum-cost retained child $n_{+}$ and proceeds recursively toward the root along the selected path $\mathcal{P}_{+}$; the modes differ only in the scalar backup cost $J_{\mathrm{bp}}$ propagated along this path. For every node $n_i\in\mathcal{P}_{+}$, the cumulative reward and visit count are updated according to
\begin{equation}
\label{eq:backpropagation_update}
    R_i^{+}
    =
    R_i
    -
    J_{\mathrm{bp}},
    \qquad
    V_i^{+}
    =
    V_i+1,
    \qquad
    \bar{R}_i^{+}
    =
    \frac{R_i^{+}}{V_i^{+}},
\end{equation}
where the superscript $+$ denotes the values after the update. Together with the edge-cost initialization in~\eqref{eq:edge_reward_initialization}, repeated application of~\eqref{eq:backpropagation_update} yields the edge-cost-seeded statistic in~\eqref{eq:avg_reward}. More specifically, the edge initialization in~\eqref{eq:edge_reward_initialization} and the backup update in~\eqref{eq:backpropagation_update} serve complementary roles: the former evaluates the incoming motion segment when a child is inserted, whereas the latter propagates the outcome of the current expansion along $\mathcal{P}_{+}$.  Since SI-MPPI does not execute a separate default-policy simulation after child insertion, $J_{\mathrm{bp}}$ serves as a cost-informed expansion return for backpropagation. For $n_i=n_{+}$, the incoming-edge cost satisfies $J_i^{\mathrm{edge}}=J_{\mathrm{win}}$. Consequently, in winner-cost mode, $J_{\mathrm{bp}}=J_{\mathrm{win}}$, so the same numerical cost contributes once through edge initialization and once through backpropagation, while the two contributions retain distinct algorithmic roles. In expected-cost mode, by contrast, the edge contribution remains $J_{\mathrm{win}}$, whereas the backup cost is $J_{\mathrm{bp}}=\widehat{J}$.

If $\mathcal{C}_{\mathrm{Acc}}=\emptyset$, no child is inserted and $n_{+}$ is undefined. In this case, a large finite failure cost $J_{\mathrm{fail}}>0$ is used in place of $J_{\mathrm{bp}}$, and~\eqref{eq:backpropagation_update} is applied along the selected path from $n_{\mathrm{s}}$ to the root node $n_{0}$.

\subsection{Goal-Directed Expansion and Termination}
\label{subsec:goal_connection}

The four-stage planning cycle is repeated until an expansion yields a minimum-cost retained child $n_{+}$ whose state lies in the goal region $\mathcal{X}_{\mathrm{goal}}$, or until the iteration budget $I_{\max}$ is exhausted. To promote goal-directed expansion as the tree approaches the mission target, let $r_{\mathrm{goal}}>0$ denote the goal-directed activation radius. When the selected node $n_{\mathrm{s}}$ satisfies
\begin{equation}
\label{eq:goal_directed_activation}
    \left\|
        \mathbf{p}_{s}
        -
        \mathbf{p}_{\mathrm{goal}}
    \right\|_{2}
    \leq
    r_{\mathrm{goal}},
\end{equation}
the goal workspace position $\mathbf{p}_{\mathrm{goal}}$ is incorporated into the structured bait set $\mathcal{B}$ according to the branching limit $k_b$. For $k_b>1$, $\mathbf{p}_{\mathrm{goal}}$ replaces one element of $\mathcal{B}$ while the remaining structured baits are retained. This policy introduces a direct goal-oriented candidate while preserving alternative local expansion directions; thus, if the goal-directed candidate is rejected during the candidate-admission stage or not retained by top-$k$ selection due to its comparatively higher cost, the remaining baits can still support continued tree expansion, thereby improving robustness near the goal.
For $k_b=1$, $\mathcal{B}$ is instead reduced to the singleton $\{\mathbf{p}_{\mathrm{goal}}\}$, ensuring that the local optimization is directed exclusively toward $\mathbf{p}_{\mathrm{goal}}$ while preventing a lower-cost exploratory candidate from occupying the single retained branch.
Since the goal-directed target $\mathbf{p}_{\mathrm{goal}}$ corresponds to the collision-free goal state $\mathbf{x}_{\mathrm{goal}}\in\mathcal{X}_{\mathrm{free}}$, it is retained directly without applying the radial-retraction rule in~\eqref{eq:bait_retraction}, while its effective expansion radius is computed according to~\eqref{eq:effective_bait_radius}. The resulting retained bait set is then processed through the same GPU-batched MPPI procedure. Any resulting motion segment becomes eligible for tree integration only if its candidate belongs to the accepted set $\mathcal{C}_{\mathrm{Acc}}$ in~\eqref{eq:accepted_set} and is subsequently retained in $\mathcal{C}_{k}$ according to~\eqref{eq:topk_set}. Thus, the local expansion is biased toward $\mathbf{p}_{\mathrm{goal}}$ without requiring a separate terminal steering or optimization procedure.

Following tree integration, let $\mathbf{x}_{+}$ and $\mathbf{p}_{+}$ denote, respectively, the state and associated workspace position represented by $n_{+}$. A goal-reaching solution is identified when
\begin{equation}
\label{eq:goal_termination}
    \mathbf{x}_{+}
    \in
    \mathcal{X}_{\mathrm{goal}}.
\end{equation}
For the 2D navigation problem considered here, the goal-reaching condition in~\eqref{eq:goal_termination} is evaluated according to
\begin{equation}
\label{eq:goal_position_condition}
    \left\|
        \mathbf{p}_{+}
        -
        \mathbf{p}_{\mathrm{goal}}
    \right\|_{2}
    \leq
    \delta_{\mathrm{goal}},
\end{equation}
where $\delta_{\mathrm{goal}}>0$ denotes the terminal-position tolerance. When~\eqref{eq:goal_termination} is satisfied, the complete plan $\pi=(\Gamma,\mathbf{U}_{\Gamma})$ is recovered by concatenating the state and control trajectories stored along $\mathcal{P}_{+}$, the path from the root node $n_{0}\in\mathcal{V}$ to $n_{+}$. The resulting plan satisfies the mission-trajectory and boundary conditions in~\eqref{eq:global_trajectory}--\eqref{eq:global_boundary_conditions}. Otherwise, the planning cycle continues, and the planner returns failure if no goal-reaching node is obtained before the iteration budget $I_{\max}$ is exhausted.

\subsection{Algorithmic Summary}
\label{subsec:algorithm_summary}

\begin{algorithm}[!h]
\caption{SI-MPPI Planning Cycle}
\label{alg:simppi}
\begin{algorithmic}[1]
\Require
Initial state $\mathbf{x}_{\mathrm{start}}$,
goal state $\mathbf{x}_{\mathrm{goal}}$,
iteration budget $I_{\max}$;
dynamics and MPPI parameters
$f$, $N$, $M$, $N_{\mathrm{opt}}$, $\lambda$, $\boldsymbol{\Sigma}_{\mathbf{u}}$;
SI-MPPI parameters
$B$, $r_b$, $\eta$, $k_b$, $r_{\mathrm{goal}}$, $\delta_{\mathrm{goal}}$, $J_{\mathrm{fail}}$;
proposal-adaptation and backpropagation parameters
$\beta_{\mathrm{w}}$, $\eta_{\mathrm{ess}}$, $\eta_{w}$, $\alpha_{\mu}$, $\eta_{\mu}$, $\mathsf{mode}_{\mathrm{bp}}$
\Ensure
Plan $\pi=(\Gamma,\mathbf{U}_{\Gamma})$ terminating in $\mathcal{X}_{\mathrm{goal}}$, or failure

\State Initialize $\mathcal{T}=(\mathcal{V},\mathcal{E})$ with root node $n_0$ representing $\mathbf{x}_{\mathrm{start}}$
\State Initialize  $\mathbf{U}_{\mathrm{ws}}$ using the default nominal sequence

\For{$i\leftarrow 1$ \textbf{to} $I_{\max}$}

    \Statex \textcolor{gray}{\textit{Stage 1: Selection}}
    \State $(n_{\mathrm{s}},\chi_{\mathrm{s}})\leftarrow\operatorname{SelectFrontier}(\mathcal{T})$
    \If{$\neg\chi_{\mathrm{s}}$}
        \State \textbf{continue}
    \EndIf

    \Statex \textcolor{gray}{\textit{Stage 2: Bait-Guided MPPI Expansion}}
    \State $\mathcal{B}\leftarrow \operatorname{GenerateBaits}(\mathbf{p}_{s},\mathbf{d}_{s},r_b,B)$
    \If{$\|\mathbf{p}_{s}-\mathbf{p}_{\mathrm{goal}}\|_{2} \leq r_{\mathrm{goal}}$}
        \State $\mathcal{B}\leftarrow\operatorname{ReplaceOneBait}(\mathcal{B},\mathbf{p}_{\mathrm{goal}})$
    \EndIf
    \State $\bigl(\mathcal{B}_{\mathrm{ret}},\{r_{b,\mathrm{eff}}^{(j)}\}\bigr)\leftarrow\operatorname{AdaptiveBaitHandling}(\mathcal{B},\mathbf{p}_{s})$
    \If{$\mathcal{B}_{\mathrm{ret}}=\emptyset$}
        \State $\operatorname{MarkDeadEnd}(n_{\mathrm{s}})$
        \State $\operatorname{Backpropagate}(n_{\mathrm{s}},J_{\mathrm{fail}})$
        \State \textbf{continue}
    \EndIf
    \State $\mathcal{C}\leftarrow\operatorname{GPUBatchedMPPI}(\mathbf{x}_{s},\mathcal{B}_{\mathrm{ret}},\mathbf{U}_{\mathrm{ws}},\boldsymbol{\Sigma}_{\mathbf{u}})$

    \Statex \textcolor{gray}{\textit{Stage 3: Candidate Admission and Tree Integration}}
    \State
    $\mathcal{C}_{\mathrm{Acc}}
    \leftarrow
    \left\{
        c^{(j)}\in\mathcal{C}
        \;\middle|\;
        \operatorname{Safe}\!\left(\tau^{(j)}\right)
        \land
        \operatorname{Prog}\!\left(c^{(j)}\right)
    \right\}$
    \If{$\mathcal{C}_{\mathrm{Acc}}=\emptyset$}
        \State $\operatorname{MarkDeadEnd}(n_{\mathrm{s}})$
        \State $\operatorname{Backpropagate}(n_{\mathrm{s}},J_{\mathrm{fail}})$
        \State \textbf{continue}
    \EndIf
    \State $k_{\mathrm{eff}}\leftarrow\min\{k_b,|\mathcal{C}_{\mathrm{Acc}}|\}$
    \State $\mathcal{C}_{k}\leftarrow\operatorname{SelectTopK}(\mathcal{C}_{\mathrm{Acc}},k_{\mathrm{eff}})$
    \State $(\mathcal{T},n_{+})\leftarrow\operatorname{IntegrateCandidates}(\mathcal{T},n_{\mathrm{s}},\mathcal{C}_{k})$

    \Statex \textcolor{gray}{\textit{Stage 4: Proposal Adaptation and Backpropagation}}
    \State $\{w^{(j)}\}_{c^{(j)}\in\mathcal{C}_{\mathrm{Acc}}}\leftarrow\operatorname{ComputeCandidateWeights}(\mathcal{C}_{\mathrm{Acc}})$
    \State $\boldsymbol{\mu}_{0:N-1}^{+}\leftarrow\operatorname{AdaptControlMean}\bigl(\mathbf{U}_{\mathrm{ws}},\mathcal{C}_{\mathrm{Acc}},\mathcal{C}_{k},\{w^{(j)}\!\}\!\bigr)$
    \State $\mathbf{U}_{\mathrm{ws}}\leftarrow\operatorname{ConstructWarmStart}(\boldsymbol{\mu}_{0:N-1}^{+})$
    \State $J_{\mathrm{bp}}\leftarrow\operatorname{ConstructBackupCost}(\mathcal{C}_{k},\{w^{(j)}\},\mathsf{mode}_{\mathrm{bp}})$
    \State $\operatorname{Backpropagate}(n_{+},J_{\mathrm{bp}})$

    \If{$\|\mathbf{p}_{+}-\mathbf{p}_{\mathrm{goal}}\|_{2}\leq\delta_{\mathrm{goal}}$}
        \State $\pi\leftarrow\operatorname{RecoverPlan}(\mathcal{T},n_{+})$
        \State \Return $\pi$
    \EndIf

\EndFor
\State \Return failure
\end{algorithmic}
\end{algorithm}

Algorithm~\ref{alg:simppi} summarizes the complete SI-MPPI planning procedure. The algorithm initializes the search tree $\mathcal{T}=(\mathcal{V},\mathcal{E})$ at $\mathbf{x}_{\mathrm{start}}$ and the MPPI warm-start sequence $\mathbf{U}_{\mathrm{ws}}$ using the default nominal sequence (lines~1--2). It then executes the four-stage planning cycle for at most $I_{\max}$ iterations (line~3).
During the selection stage, the tree $\mathcal{T}$ is recursively traversed using the UCB policy in~\eqref{eq:ucb_prelim}, with the edge-cost-seeded reward statistic defined in~\eqref{eq:avg_reward}, yielding the selected node $n_{\mathrm{s}}$ and a Boolean flag $\chi_{\mathrm{s}}$ indicating whether it is eligible for expansion (line~4). If $\chi_{\mathrm{s}}=\mathrm{false}$, no valid expandable frontier has been identified and the current iteration is skipped (lines~5--7).
For a valid selected frontier, the planner then proceeds to the bait-guided expansion stage. The structured bait set $\mathcal{B}$ is first generated according to~\eqref{eq:bait_set}--\eqref{eq:baits} (line~8). When the goal-directed activation condition in~\eqref{eq:goal_directed_activation} is satisfied, one structured bait is replaced by $\mathbf{p}_{\mathrm{goal}}$ (lines~9--11). Subsequently, adaptive bait handling produces the retained set $\mathcal{B}_{\mathrm{ret}}$ and its effective expansion radii according to~\eqref{eq:bait_validity}--\eqref{eq:effective_bait_radius} (line~12). If no bait is retained, $n_{\mathrm{s}}$ is marked as a dead end and the failure cost $J_{\mathrm{fail}}$ is backpropagated along the selected tree path (lines~13--17). Otherwise, the local optimization problems associated with the retained baits are solved concurrently by GPU-batched MPPI, yielding the transient candidate set $\mathcal{C}$ defined in~\eqref{eq:candidate_set} (line~18).
With the transient candidate set available, the candidate-admission stage applies the collision-safety and progress conditions to obtain $\mathcal{C}_{\mathrm{Acc}}$ according to~\eqref{eq:accepted_set} (line~19). If the accepted set is empty, the same dead-end and failure-cost update is applied (lines~20--24). Otherwise, the effective branching factor is computed using~\eqref{eq:effective_topk}, the lowest-cost accepted candidates are retained according to~\eqref{eq:topk_set}, and their motion segments are integrated into $\mathcal{T}$ (lines~25--27).
During this integration, each newly inserted child is edge-cost seeded according to~\eqref{eq:edge_reward_initialization}. The minimum-cost retained child resulting from this integration is denoted by $n_{+}$.

Following tree integration, the final stage first normalizes and weights the accepted candidate costs according to~\eqref{eq:norm_cost}--\eqref{eq:softmax_weights} (line~28). The resulting weight distribution then determines the ESS-guided control-mean regime in~\eqref{eq:proposal_update_mode}; the nominal mean is updated using either~\eqref{eq:moment_mean_update} or~\eqref{eq:winner_mean_update}, and the next warm-start sequence is constructed according to~\eqref{eq:warmstart_modes} (lines~29--30). Next, the selected backpropagation mode determines the expansion-level backup cost $J_{\mathrm{bp}}$ according to~\eqref{eq:backpropagation_cost}, which is propagated from $n_{+}$ toward the root node $n_{0}$ using~\eqref{eq:backpropagation_update} (lines~31--32). After completing these updates, $n_{+}$ is evaluated using the goal-reaching condition in~\eqref{eq:goal_position_condition}. If the condition is satisfied, the complete plan $\pi=(\Gamma,\mathbf{U}_{\Gamma})$ is recovered from the $n_{0}$-to-$n_{+}$ path and returned (lines~33--36). Otherwise, the four-stage planning cycle proceeds to the next iteration; if the iteration budget is exhausted without obtaining a goal-reaching node, the algorithm returns failure (lines~37--38).\enlargethispage{3pt}

\vspace*{-3pt}
\section{Results and Analysis}
\label{sec:results}
This section evaluates our proposed framework for kinodynamic motion planning in large-scale, cluttered environments through simulation-based comparisons against three baseline planners and an ablation study of its key components.

\begin{figure*}[!t]
    \centering
    \captionsetup[sub]{justification=centering} 
    \begin{subfigure}[t]{0.24\textwidth}
        \centering
        \includegraphics[width=\linewidth]
        {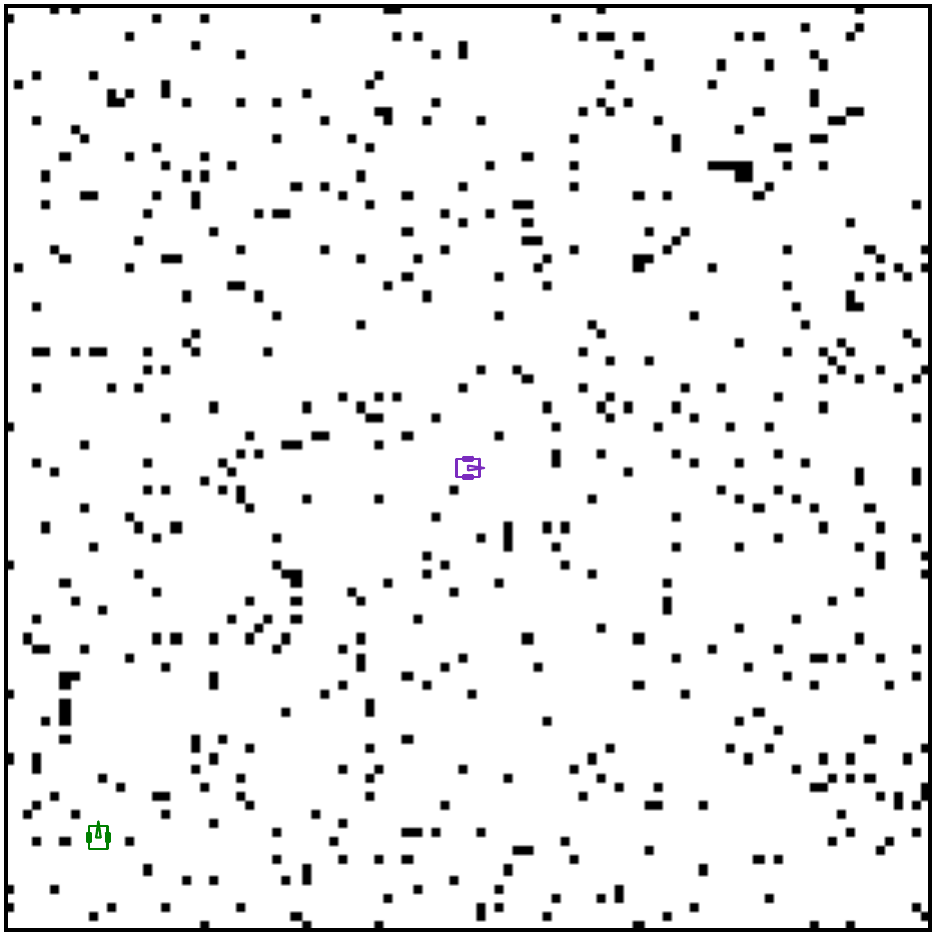}
        \vspace*{-8pt}
        \caption{\emph{Random Obstacle Field} (\mapname{Rand})}
        \label{fig:map_random}
    \end{subfigure}
    \hfill
    \begin{subfigure}[t]{0.24\textwidth}
        \centering
        \includegraphics[width=\linewidth]
        {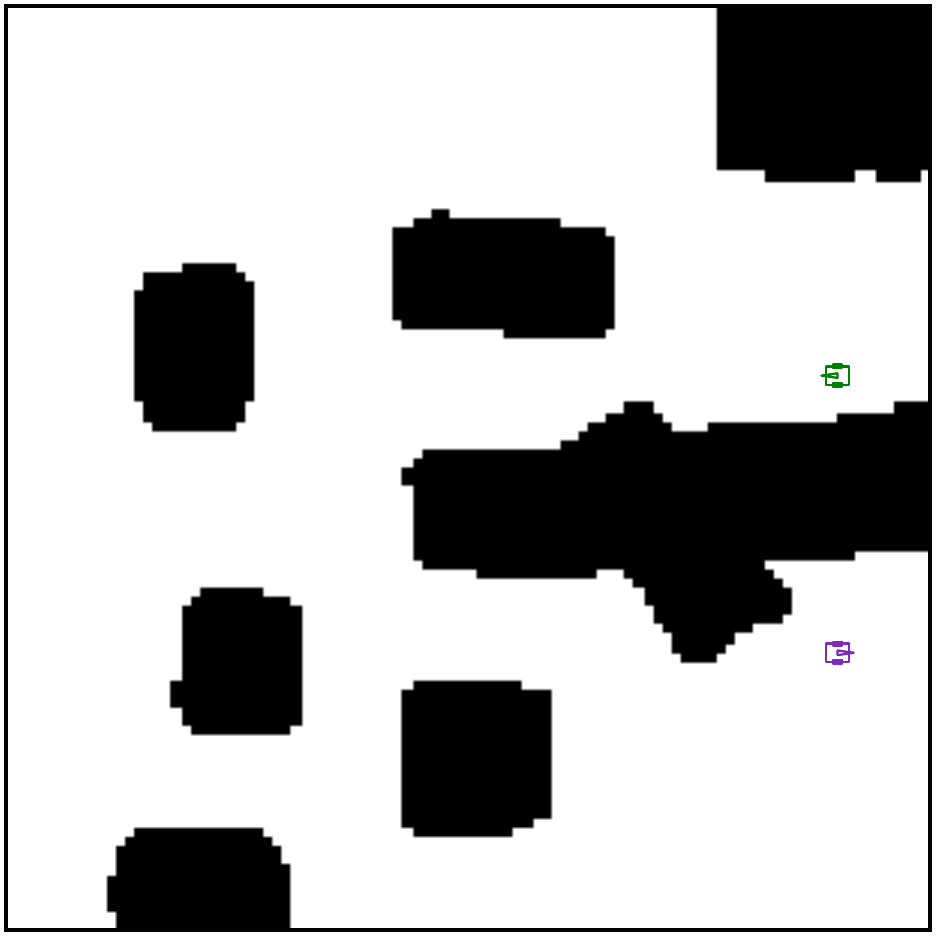}
        \vspace*{-8pt}
        \caption{\emph{Cave Map} (\mapname{Cave})}
        \label{fig:map_cave}
    \end{subfigure}
    \hfill
    \begin{subfigure}[t]{0.24\textwidth}
        \centering
        \includegraphics[width=\linewidth]
        {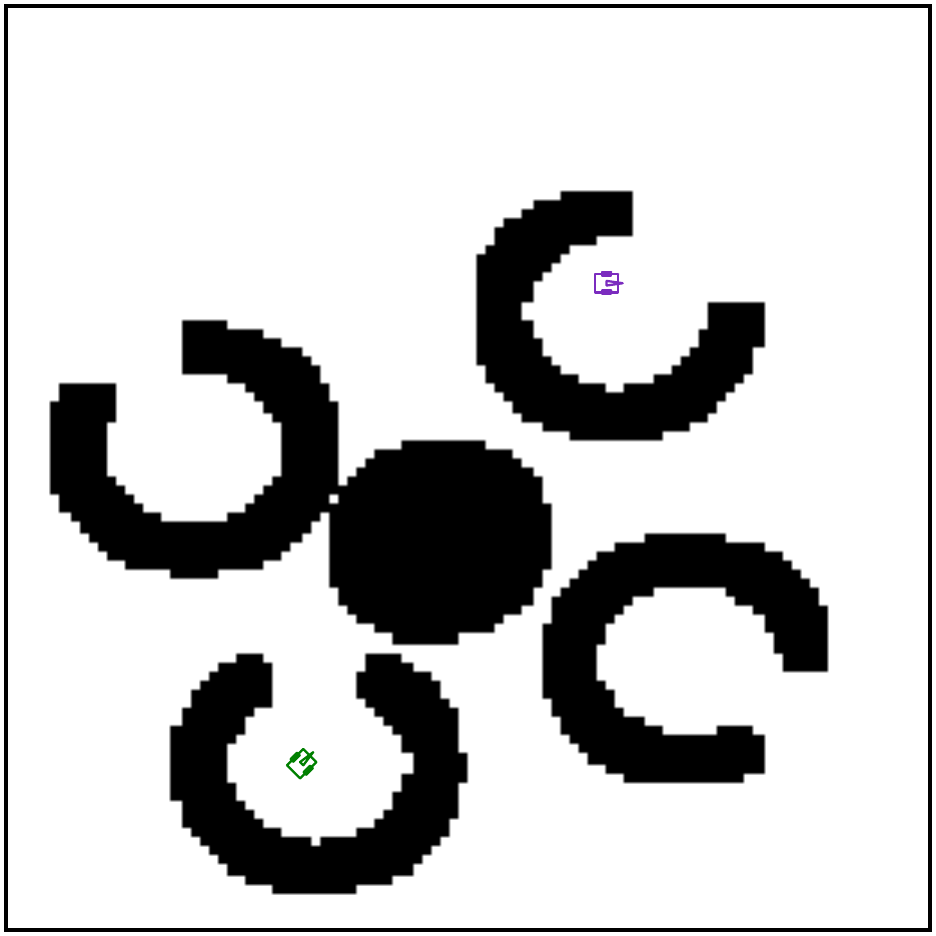}
        \vspace*{-8pt}
        \caption{\emph{Four-Moons Maze} (\mapname{4Moons})}
        \label{fig:map_four_moons}
    \end{subfigure}
    \hfill
    \begin{subfigure}[t]{0.24\textwidth}
        \centering
        \includegraphics[width=\linewidth]
        {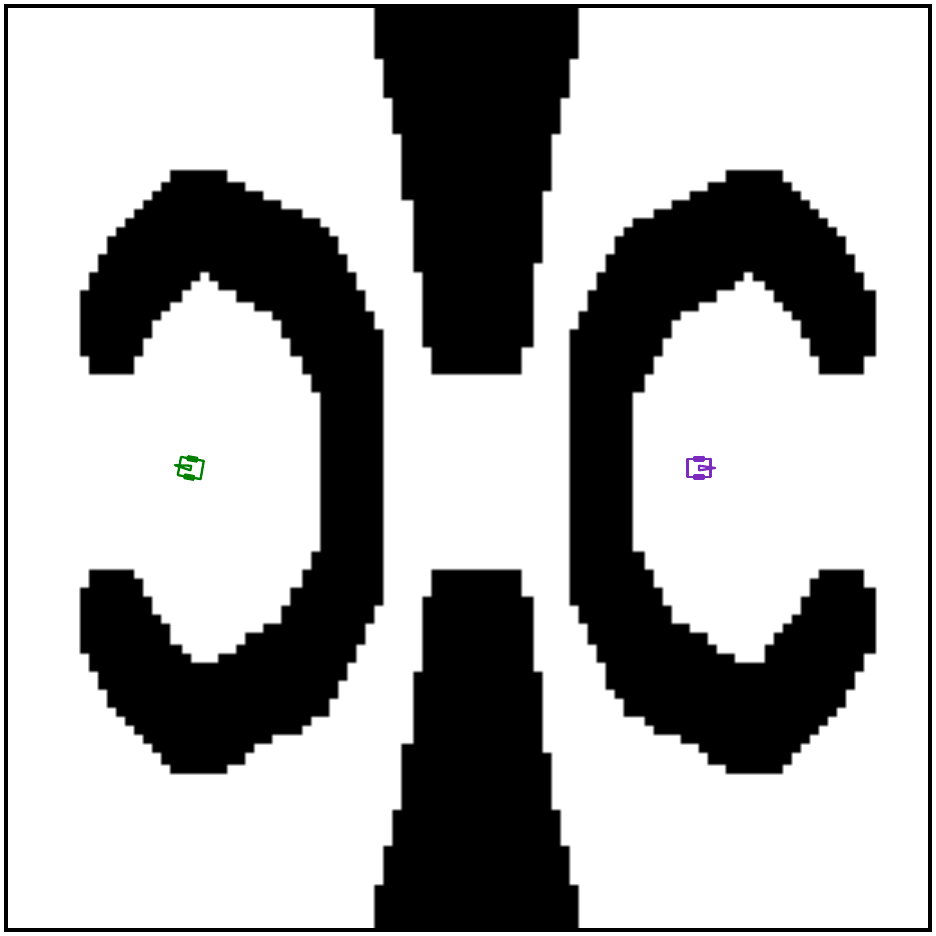}
        \vspace*{-8pt}
        \caption{\emph{Moon-Gate Maze} (\mapname{MGate})}
        \label{fig:map_moon_gate}
    \end{subfigure}
    \vspace*{-8pt}
    \caption{Benchmark $\SI{100}{\metre}\times\SI{100}{\metre}$ occupancy-grid maps used in the offline planning evaluation, where white and black cells depict free and occupied space, respectively, while green and purple markers indicate the robot's start and goal poses shared by all planners.
    }
    \label{fig:benchmark_maps}
\end{figure*}
\subsection{Offline Planning Evaluation}
\label{subsec:offline_planning}

\subsubsection{Planning Setup}
\label{subsubsec:planning_setup}
In this study, we consider kinodynamic planning for a differential-drive robot navigating in static, cluttered 2D environments represented by binary occupancy grids. The forward-Euler discretization of the differential-drive kinematic model used for trajectory propagation is given by
\begin{equation}
\label{eq:unicycle_dynamics}
    \mathbf{x}_{k+1}
    =
    \mathbf{x}_{k}
    +
    T_s
    \begin{bmatrix}
        \cos\theta_{k} & 0
        \\
        \sin\theta_{k} & 0
        \\
        0 & 1
    \end{bmatrix}
    \mathbf{w}_{k},
    \quad
    \mathbf{w}_{k}
    =
    \mathbf{u}_{k}
    +
    \delta\mathbf{u}_{k},
\end{equation}
where $\mathbf{x}_{k}=[\mathbf{p}_{k}^{\top},\theta_{k}]^{\top}$ and $\mathbf{u}_{k}=[v_{k},\omega_{k}]^{\top}$ denote the robot state and nominal control input, respectively, $\mathbf{p}_{k}=[x_{k},y_{k}]^{\top}$ is the Cartesian position, $\theta_{k}$ is the heading angle, and $v_{k}$ and $\omega_{k}$ are the commanded linear and angular velocities. Here, $T_s=1/f_c$ denotes the discrete-time sampling interval associated with the control frequency $f_c$, while $\mathbf{w}_{k}$ is the perturbed control input defined in~\eqref{eq:dynamics_system}. For deterministic kinodynamic propagation, the injected control perturbation is set to $\delta\mathbf{u}_{k}=\mathbf{0}$, and hence $\mathbf{w}_{k}=\mathbf{u}_{k}$.

Throughout the reported SI-MPPI simulations, the robot and algorithm parameters are fixed to the values specified below unless a parameter is explicitly varied in an ablation study. 
The bait-screening and candidate-collision conditions  in~\eqref{eq:bait_validity} and~\eqref{eq:candidate_collision_safety}, respectively, are evaluated using the same rectangular robot-footprint model, with length $\SI{0.5}{\metre}$, width $\SI{0.4}{\metre}$, and a $\SI{0.05}{\metre}$ safety buffer; for candidate-trajectory validation, the index set $\mathcal{K}_{\mathrm{chk}}$ in \eqref{eq:collision_check_indices} is constructed using $s_{\mathrm{chk}}=5$.
The goal-reaching condition in~\eqref{eq:goal_position_condition} is evaluated using $\delta_{\mathrm{goal}}=\SI{1.0}{\metre}$.
Each tree expansion generates $B=8$ structured baits according to~\eqref{eq:bait_set}--\eqref{eq:baits}, with nominal bait radius $r_b=\SI{5.0}{\metre}$. Geometrically invalid baits are retracted using $\eta=0.8$ in~\eqref{eq:bait_retraction}. Consistent with the default evaluation setting, $\operatorname{Prog}(c^{(j)})$ is instantiated solely by the normalized-displacement criterion in~\eqref{eq:candidate_progress} with $\eta_{\mathrm{dist}}=0.8$; the optional clearance-gain and heading-alignment tests are disabled. Goal-directed expansion is activated within $r_{\mathrm{goal}}=\SI{5.0}{\metre}$ according to~\eqref{eq:goal_directed_activation}. The fixed-retention form of the top-$k$ rule in~\eqref{eq:effective_topk}--\eqref{eq:topk_set} retains at most $k_b=3$ accepted candidates from each expansion. The search budget is $I_{\max}=800$ iterations, and an unsuccessful expansion is assigned the large finite failure cost $J_{\mathrm{fail}}=10^{9}$.

For each retained bait, the GPU-batched MPPI optimization evaluates $M=1120$ sampled trajectories per refinement iteration and performs $N_{\mathrm{opt}}=4$ refinement iterations. The prediction horizon contains $N=120$ control intervals at $f_c=\SI{10}{\hertz}$; consequently, $T_s=1/f_c=\SI{0.1}{\second}$ and the horizon duration is $NT_s=\SI{12.0}{\second}$. The covariance of the Gaussian control perturbations in~\eqref{eq:dynamics_system} is held fixed throughout the search at $\boldsymbol{\Sigma}_{\mathbf{u}}=0.25\mathbf{I}_{2}$. The MPPI temperature $\lambda$ in~\eqref{eq:mppi_weights} and the exploration parameter $\nu$ in~\eqref{eq:mppi_control_cost} are set to $10.0$ and $100$, respectively. The nominal control sequence $\mathbf{U}$ in~\eqref{eq:mppi_optimal-control} is initialized for the first expansion by repeating $\boldsymbol{\mu}_{0}=[0.3,0.05]^{\top}$ across the $N$-step prediction horizon and is subsequently warm-started using the horizon-averaged mode in~\eqref{eq:warmstart_modes}. Following MPPI optimization, the resulting control sequences are smoothed using a Savitzky--Golay filter with window length $51$ and polynomial order $3$. Additionally, to enhance planning efficiency, the warm-start sequence $\mathbf{U}_{\mathrm{ws}}$ in~\eqref{eq:warmstart_modes} is replicated for each retained bait $\mathbf{b}^{(j)}$ at the selected node $n_{\mathrm{s}}$, with its first $K$ angular-velocity inputs initialized as
\begin{equation}
\label{eq:bait_yaw_initialization}
    \omega_{k,0}^{(j)}
    =
    (1-\alpha_h)\omega_{\mathrm{ws},k}
    +
    \alpha_h\omega_{\mathrm{bait}}^{(j)},
    \quad k=0,\ldots,K-1,
\end{equation}
where $\omega_{\mathrm{ws},k}$ is the warm-start angular velocity, $\omega_{\mathrm{bait}}^{(j)}=(\varphi^{(j)}-\theta_s)/(KT_s)$ is the desired yaw rate toward $\mathbf{b}^{(j)}$, and $\varphi^{(j)}$ denotes the bearing angle from the selected-node position $\mathbf{p}_s$ to $\mathbf{b}^{(j)}$. Here, we set $K=10$, $\alpha_h=0.5$, and bound $\omega_{\mathrm{bait}}^{(j)}$ by $\pm\SI{0.6}{\radian\per\second}$.
For candidate-level statistical adaptation, the normalization constant in~\eqref{eq:norm_cost} and the candidate-weighting temperature in~\eqref{eq:softmax_weights} are set to $\varepsilon=10^{-6}$ and $\beta_{\mathrm{w}}=0.3$, respectively. The regime-selection rule in~\eqref{eq:proposal_update_mode} uses $\eta_{\mathrm{ess}}=0.3$ and $\eta_{w}=0.55$, while the moment-matching and winner-step gains in \eqref{eq:moment_mean_update} and~\eqref{eq:winner_mean_update} are set to $\alpha_{\mu}=0.3$ and $\eta_{\mu}=0.8$, respectively. For tree selection and backpropagation, the UCB coefficient in~\eqref{eq:ucb_prelim} is set to $c_{\mathrm{ucb}}=0.2$, and the expected-cost mode is selected by setting $\mathsf{mode}_{\mathrm{bp}}=\mathrm{expected}$ in~\eqref{eq:backpropagation_cost}.

For the 2D navigation task, the task-dependent state cost $q(\cdot)$ used in the instantaneous running cost $\widetilde{q}$ in~\eqref{eq:mppi_running_cost}, and hence in the candidate evaluation in~\eqref{eq:bait_cost}, is decomposed into bait-tracking and obstacle-avoidance costs as
\begin{equation}
\label{eq:task_state_cost}
    q(\mathbf{x};\mathbf{x}_{\mathrm{b}})
    =
    q_{\mathrm{bait}}(\mathbf{x};\mathbf{x}_{\mathrm{b}})
    +
    q_{\mathrm{obs}}(\mathbf{x}).
\end{equation}
The bait-tracking cost $q_{\mathrm{bait}}(\mathbf{x};\mathbf{x}_{\mathrm{b}}) = (\mathbf{x}-\mathbf{x}_{\mathrm{b}})^{\top} \mathbf{Q} (\mathbf{x}-\mathbf{x}_{\mathrm{b}})$ penalizes deviation from the desired bait state and thereby guides the local optimization toward it, where $\mathbf{Q}$ is the state-error weighting matrix. The obstacle-avoidance cost is defined as $q_{\mathrm{obs}}(\mathbf{x}) = w_{\mathrm{obs}}\mathbb{I}_{\mathrm{obs}}(\mathbf{x})$, where $\mathbb{I}_{\mathrm{obs}}(\mathbf{x})$ is a Boolean indicator that equals one when the state is in collision with an obstacle or lies outside the workspace, and zero otherwise, while $w_{\mathrm{obs}}$ is the corresponding penalty weight. In addition to the state cost $q(\cdot)$ and the quadratic control cost $q_{\mathbf{u}}$, the instantaneous running cost $\widetilde{q}$ includes the optional cruise-speed cost
\begin{equation}
\label{eq:cruise_cost}
    q_{\mathrm{cruise}}(v_{\mathrm{a}})
    =
    w_c
    \left(
        v_{\mathrm{a}}-v_{\max}
    \right)^2,
    \qquad
    w_c\geq0,
\end{equation}
which encourages the robot to operate near its maximum admissible linear velocity. Here, $v_{\mathrm{a}}=v+\delta v$ during sampled MPPI rollout evaluation, whereas $v_{\mathrm{a}}=v$ during deterministic candidate reevaluation. Within the scope of this work, the state-error and control-effort weighting matrices are set to $\mathbf{Q}=\operatorname{diag}(1,1,1)$ and $\mathbf{R}=\operatorname{diag}(20,20)$, respectively, while the obstacle-avoidance weight is set to $w_{\mathrm{obs}}=10^{4}$. The cruise-speed cost is disabled by setting $w_c=0$, and the terminal-progress coefficient in~\eqref{eq:term_cost} is set to $\alpha=1.2$. 
The bait-generation rule in~\eqref{eq:baits} is specialized to the planar case ($d=2$) using the heading $\theta_s$ associated with the selected state $\mathbf{x}_s$. Specifically, the bait directions are
\begin{equation}
\label{eq:baits_planar}
    \mathbf{d}^{(j)}
    =
    \begin{bmatrix}
        \cos(\theta_s+\alpha_j)
        \\
        \sin(\theta_s+\alpha_j)
    \end{bmatrix},
    \quad
    \alpha_j
    =
    -\frac{\pi}{2}
    +(j-1)\Delta\alpha,
\end{equation}
where $j\in\{1,\ldots,B\}$ and $\Delta\alpha=\pi/B$. Thus, the angular offsets are uniformly distributed over $[-\pi/2,\pi/2)$, placing the baits along a forward semicircular arc centered at the selected-node position $\mathbf{p}_s$. For each retained bait $\mathbf{b}^{(j)}$, the corresponding target heading $\theta_{\mathrm{b}}^{(j)}$ is set by default to the bearing from $\mathbf{b}^{(j)}$ toward the goal position $\mathbf{p}_{\mathrm{goal}}$. 
\subsubsection{Baselines and Benchmark Environments}
\label{subsubsec:baselines_and_benchmark_environments}
To assess the performance of SI-MPPI, we compare it against kinodynamic MCTS~\cite{chen2019pareto}, LQR-RRT$^{*}$~\cite{perez2012lqr}, and Informed Kinodynamic RRT$^{*}$ (IK-RRT$^{*}$).
The kinodynamic MCTS baseline shares the UCB-based tree-selection principle with SI-MPPI but expands the tree using a finite set of discrete steering macro-actions and subsequent simulation rollouts rather than continuous-control MPPI optimization. It therefore helps isolate the contribution of the proposed bait-guided, optimization-based expansion mechanism. LQR-RRT$^{*}$ provides a sampling-based kinodynamic baseline that combines LQR-based local steering with RRT$^{*}$ parent selection and rewiring. IK-RRT$^{*}$ is adapted from Informed RRT$^{*}$~\cite{gammell2014informed} by replacing geometric edge extension with forward-simulated differential-drive steering while retaining informed sampling and RRT$^{*}$ optimization. 

All methods are implemented in Python and evaluated on the same computing platform equipped with an NVIDIA GeForce GTX~1660~Ti laptop GPU. 
SI-MPPI executes its sampled trajectory rollouts on the GPU using PyCUDA, whereas tree management and the three kinodynamic baselines execute on the CPU. To ensure a fair and computationally tractable comparison on the large-scale benchmark maps, we adapted existing baseline implementations to shared evaluation settings---a common occupancy-grid representation, rectangular footprint model, and unified logging pipeline---and introduced targeted implementation refinements while preserving each planner's defining algorithmic mechanisms and method-specific termination criteria. Among these refinements, the LQR-RRT$^{*}$ implementation employs a Joseph-stabilized Riccati update to improve finite-precision numerical stability and accelerate convergence of the LQR steering process toward its target state~\cite{lewis2012optimal}.
\ifanonymous
Source code, parameter files, maps, and benchmarking scripts will be released upon acceptance.
\else
Source code, parameter files, maps, and benchmarking scripts are provided in the accompanying open-source repository.%
\fi

In order to assess the effectiveness of the proposed framework in cluttered environments, we consider four static binary occupancy-grid maps, each spanning $\SI{100}{\metre}\times\SI{100}{\metre}$ and posing a distinct type of planning difficulty, as shown in Fig.~\ref{fig:benchmark_maps}. More precisely, the \emph{Random Obstacle Field} (hereafter referred to as \mapname{Rand}) contains spatially distributed clutter, while the \emph{Cave Map} (\mapname{Cave}) combines irregular, extended obstacles with multiple open regions; the \emph{Four-Moons Maze} (\mapname{4Moons}) contains repeated concave obstacles that create local traps and alternative passages, whereas the \emph{Moon-Gate Maze} (\mapname{MGate}) contains narrow gated passages and symmetric detours. 
For each benchmark configuration, the start and goal poses are fixed across all methods, with positions given in meters and headings in degrees.
In \mapname{Rand}, the poses are $\mathbf{x}_{\mathrm{start}}=[10,10,90]^{\top}$ and $\mathbf{x}_{\mathrm{goal}}=[50,50,0]^{\top}$; in \mapname{Cave}, $\mathbf{x}_{\mathrm{start}}=[90,60,-180]^{\top}$ and $\mathbf{x}_{\mathrm{goal}}=[90,30,0]^{\top}$; in \mapname{4Moons}, $\mathbf{x}_{\mathrm{start}}=[32,18,45]^{\top}$ and $\mathbf{x}_{\mathrm{goal}}=[65,70,0]^{\top}$; and in \mapname{MGate}, $\mathbf{x}_{\mathrm{start}}=[20,50,170]^{\top}$ and $\mathbf{x}_{\mathrm{goal}}=[75,50,0]^{\top}$. Across all environments and methods, the maximum linear velocity is set to $v_{\max}=\SI{2.0}{\metre\per\second}$, while the continuous-control planners use a maximum angular velocity of $\omega_{\max}=\SI{3.0}{\radian\per\second}$. Kinodynamic MCTS instead uses discrete steering macro-actions with a maximum steering angle of $\SI{0.12}{\radian}$, selected empirically since larger steering angles consistently degraded its planning performance.

\subsubsection{Performance Metrics}
\label{subsubsec:performance_metrics}
To assess kinodynamic planning performance, we employ metrics of mission success, search complexity, and trajectory quality, with each successful run yielding a feasible plan $\pi\in\Pi_{\mathrm{feas}}$ that serves as a practical approximation to the optimal plan $\pi^{\star}$ in~\eqref{eq:optimal_kinodynamic_problem}. Mission success is quantified by the success rate $P_{\mathrm{succ}}$, defined as the fraction of trials in which the planner returns a collision-free plan satisfying the terminal-position condition in~\eqref{eq:goal_position_condition} within the allocated computational budget.
Search complexity is characterized by the planning time $T_{\mathrm{plan}}$, the convergence time $T_{\mathrm{conv}}$, and the number of tree nodes $N_{\mathrm{tree}}$, which together capture computational effort and search-tree growth. Specifically, $T_{\mathrm{plan}}$ is the wall-clock time from planning initiation until the planner terminates and returns its reported plan, whereas $T_{\mathrm{conv}}$ is the wall-clock time from planning initiation until the last improvement of the best solution cost (cf.~\cite{gammell2014informed}). For the RRT$^{*}$-based baselines, which continue to refine their solutions until their iteration budgets are exhausted, $T_{\mathrm{conv}}$ is generally smaller than $T_{\mathrm{plan}}$, and the difference $T_{\mathrm{plan}}-T_{\mathrm{conv}}$ corresponds to computation that yields no further improvement. Since SI-MPPI and MCTS terminate upon returning the first feasible goal-reaching plan, their final cost is attained at termination, i.e., $T_{\mathrm{conv}}=T_{\mathrm{plan}}$; accordingly, only $T_{\mathrm{plan}}$ is reported for these planners. Finally, $N_{\mathrm{tree}}$ denotes the number of nodes in the search tree at termination.

\needspace{2\baselineskip} 
In addition to search complexity, robot trajectory quality is characterized by the path length $\ell_{\Gamma}$, angular jitter $\sigma_{\omega}$, and root-mean-square angular acceleration $a_{\omega,\mathrm{rms}}$, which collectively capture path efficiency and steering smoothness. For each successful plan, these metrics are evaluated from the position sequence $\{\mathbf{p}_{k}\}_{k=0}^{L}$ and angular-rate sequence $\{\omega_{k}\}_{k=0}^{L-1}$. Here, $L$ denotes the mission horizon of the returned plan, i.e., the number of control intervals in $\mathbf{U}_{\Gamma}$; $\mathbf{p}_{k}$ is the planar-position component of the $k$th state in $\Gamma$; and $\omega_{k}$ is the angular-velocity component of the $k$th control $\mathbf{u}_{k}$ in $\mathbf{U}_{\Gamma}$.
The path length is computed as
\begin{equation}
\label{eq:path_length_metric}
    \ell_{\Gamma}
    =
    \sum_{k=0}^{L-1}
    \left\|
        \mathbf{p}_{k+1}
        -
        \mathbf{p}_{k}
    \right\|_{2},
\end{equation}
while the angular jitter $\sigma_{\omega}$ is defined as the standard deviation of $\{\omega_{k}\}_{k=0}^{L-1}$. The root-mean-square (RMS) angular acceleration is given by
\begin{equation}
\label{eq:angular_acceleration_metric}
    a_{\omega,\mathrm{rms}}
    =
    \sqrt{
        \frac{1}{L-1}
        \sum_{k=0}^{L-2}
        \left(
            \frac{
                \omega_{k+1}-\omega_{k}
            }{
                T_s
            }
        \right)^{2}
    },
\end{equation}
where $T_s$ is the control sampling interval introduced in \eqref{eq:unicycle_dynamics}. A smaller $\ell_{\Gamma}$ indicates a shorter path, whereas lower values of $\sigma_{\omega}$ and $a_{\omega,\mathrm{rms}}$ indicate reduced angular-rate variation and smoother steering changes, respectively. Notably, the success rate $P_{\mathrm{succ}}$ is computed over all trials, whereas the search-complexity and trajectory-quality metrics are evaluated only over successful runs.

\begin{table*}[!t]
    \caption{Offline planning performance (mean~$\pm$~std over five trials) of SI-MPPI and the three kinodynamic baselines across the four benchmark maps, with gray cells highlighting the best values within each map and $T_{\mathrm{conv}}$ reported only for the RRT$^{*}$-based baselines.}
    \label{tab:offline_benchmark_means}
    \centering
    \footnotesize
    \addtolength{\tabcolsep}{-2.5pt}
    \setlength{\extrarowheight}{1pt}
    \begin{tabular}{|l||c|c|c|c|c|c|}
    \hline
    Method
      & $P_{\mathrm{succ}}$~[\%]\;$\uparrow$
      & $T_{\mathrm{plan}}\,(T_{\mathrm{conv}})$~[\si{\second}]\;$\downarrow$
      & $\ell_{\Gamma}$~[\si{\metre}]\;$\downarrow$
      & $\sigma_{\omega}$~[\si{\radian\per\second}]\;$\downarrow$
      & $a_{\omega,\mathrm{rms}}$~[\si{\radian\per\second\squared}]\;$\downarrow$
      & $N_{\mathrm{tree}}$\;$\downarrow$ \\
    \hline\hline
    \multicolumn{7}{|c|}{\textit{Random Obstacle Field} (\mapname{Rand}): $\SI{100}{\metre}\times\SI{100}{\metre}$ workspace, $v_{\max}=\SI{2.0}{\metre\per\second}$, $\omega_{\max}=\SI{3.0}{\radian\per\second}$} \\
    \hline
    MCTS
      & \cellcolor{gray!30}100
      & \cellcolor{gray!30}$0.32 \pm 0.24$  & $63.30 \pm 4.85$       & $0.45 \pm 0.24$       & $1.97 \pm 0.42$       & $503.40 \pm 271.24$ \\
    LQR-RRT$^{*}$
      & \cellcolor{gray!30}100
      & $354.60 \pm 77.62\,(9.99 \pm 19.31)$
      & $72.11 \pm 6.95$
      & $0.89 \pm 0.06$
      & $6.98 \pm 0.33$
      & $3791.40 \pm 343.74$ \\
    IK-RRT$^{*}$
      & \cellcolor{gray!30}100
      & $231.97 \pm 3.80\,(79.56 \pm 73.32)$
      & \cellcolor{gray!30}$58.18 \pm 0.54$
      & $0.98 \pm 0.14$
      & $9.35 \pm 0.94$
      & $9370.20 \pm 138.47$ \\
    SI-MPPI (Ours)
      & \cellcolor{gray!30}100
      & $4.12 \pm 1.77$      & $60.62 \pm 1.23$
      & \cellcolor{gray!30}$0.10 \pm 0.02$
      & \cellcolor{gray!30}$0.25 \pm 0.06$
      & \cellcolor{gray!30}$216.20 \pm 133.01$ \\
    \hline\hline
    \multicolumn{7}{|c|}{\textit{Cave Map} (\mapname{Cave}): $\SI{100}{\metre}\times\SI{100}{\metre}$ workspace, $v_{\max}=\SI{2.0}{\metre\per\second}$, $\omega_{\max}=\SI{3.0}{\radian\per\second}$} \\
    \hline
    MCTS
      & \cellcolor{gray!30}100
      & $11.73 \pm 4.77$      & $187.76 \pm 28.07$
      & $0.32 \pm 0.12$
      & $1.35 \pm 0.27$
      & $13191.20 \pm 5516.10$ \\
    LQR-RRT$^{*}$
      & \cellcolor{gray!30}100
      & $250.23 \pm 10.99\,(1.33 \pm 0.20)$
      & $146.17 \pm 7.87$
      & $0.92 \pm 0.06$
      & $7.09 \pm 0.18$
      & $3762.00 \pm 30.21$ \\
    IK-RRT$^{*}$
      & \cellcolor{gray!30}100
      & $37.45 \pm 9.43\,(31.15 \pm 9.65)$
      & $119.41 \pm 4.34$
      & $0.66 \pm 0.11$
      & $6.03 \pm 0.74$
      & $3465.60 \pm 354.50$ \\
    SI-MPPI (Ours)
      & \cellcolor{gray!30}100
      & \cellcolor{gray!30}$5.48 \pm 3.27$      & \cellcolor{gray!30}$118.60 \pm 1.80$
      & \cellcolor{gray!30}$0.11 \pm 0.01$
      & \cellcolor{gray!30}$0.27 \pm 0.03$
      & \cellcolor{gray!30}$338.40 \pm 255.27$ \\
    \hline\hline
    \multicolumn{7}{|c|}{\textit{Four-Moons Maze} (\mapname{4Moons}): $\SI{100}{\metre}\times\SI{100}{\metre}$ workspace, $v_{\max}=\SI{2.0}{\metre\per\second}$, $\omega_{\max}=\SI{3.0}{\radian\per\second}$} \\
    \hline
    MCTS
      & \cellcolor{gray!30}100
      & \cellcolor{gray!30}$3.37 \pm 0.82$      & $209.79 \pm 17.92$
      & $0.30 \pm 0.08$
      & $1.46 \pm 0.08$
      & $6039.80 \pm 716.24$ \\
    LQR-RRT$^{*}$
      & \cellcolor{gray!30}100
      & $142.37 \pm 51.42\,(12.60 \pm 20.40)$
      & $177.97 \pm 24.61$
      & $0.96 \pm 0.02$
      & $7.40 \pm 0.15$
      & $2965.60 \pm 380.25$ \\
    IK-RRT$^{*}$
      & \cellcolor{gray!30}100
      & $37.38 \pm 14.07\,(22.13 \pm 18.95)$
      & $138.28 \pm 18.49$
      & $0.70 \pm 0.07$
      & $6.41 \pm 0.85$
      & $4117.80 \pm 803.89$ \\
    SI-MPPI (Ours)
      & \cellcolor{gray!30}100
      & $3.95 \pm 1.02$      & \cellcolor{gray!30}$111.69 \pm 7.57$
      & \cellcolor{gray!30}$0.12 \pm 0.01$
      & \cellcolor{gray!30}$0.29 \pm 0.01$
      & \cellcolor{gray!30}$240.40 \pm 75.12$ \\
    \hline\hline
    \multicolumn{7}{|c|}{\textit{Moon-Gate Maze} (\mapname{MGate}): $\SI{100}{\metre}\times\SI{100}{\metre}$ workspace, $v_{\max}=\SI{2.0}{\metre\per\second}$, $\omega_{\max}=\SI{3.0}{\radian\per\second}$} \\
    \hline
    MCTS
      & \cellcolor{gray!30}100
      & \cellcolor{gray!30}$8.45 \pm 3.23$      & $249.43 \pm 3.37$
      & $0.26 \pm 0.10$
      & $1.33 \pm 0.25$
      & $12397.20 \pm 3786.59$ \\
    LQR-RRT$^{*}$
      & \cellcolor{gray!30}100
      & $151.93 \pm 39.88\,(48.12 \pm 67.90)$
      & $255.34 \pm 14.73$
      & $0.95 \pm 0.07$
      & $7.05 \pm 0.38$
      & $3359.20 \pm 247.28$ \\
    IK-RRT$^{*}$
      & \cellcolor{gray!30}100
      & $16.98 \pm 6.33\,(5.69 \pm 7.18)$
      & \cellcolor{gray!30}$224.89 \pm 8.83$
      & $0.80 \pm 0.04$
      & $7.21 \pm 0.50$
      & $1829.20 \pm 104.73$ \\
    SI-MPPI (Ours)
      & \cellcolor{gray!30}100
      & $19.98 \pm 6.44$      & $264.68 \pm 5.71$
      & \cellcolor{gray!30}$0.11 \pm 0.01$
      & \cellcolor{gray!30}$0.28 \pm 0.03$
      & \cellcolor{gray!30}$1290.40 \pm 401.26$ \\
    \hline
    \end{tabular}
    \end{table*}

\subsubsection{Comparative Performance Analysis}
\label{subsubsec:comparative_performance}
\begin{figure*}[!t]
    \centering
    \begin{subfigure}[t]{0.245\textwidth}
        \centering
        \includegraphics[width=\linewidth,height=0.975\linewidth]{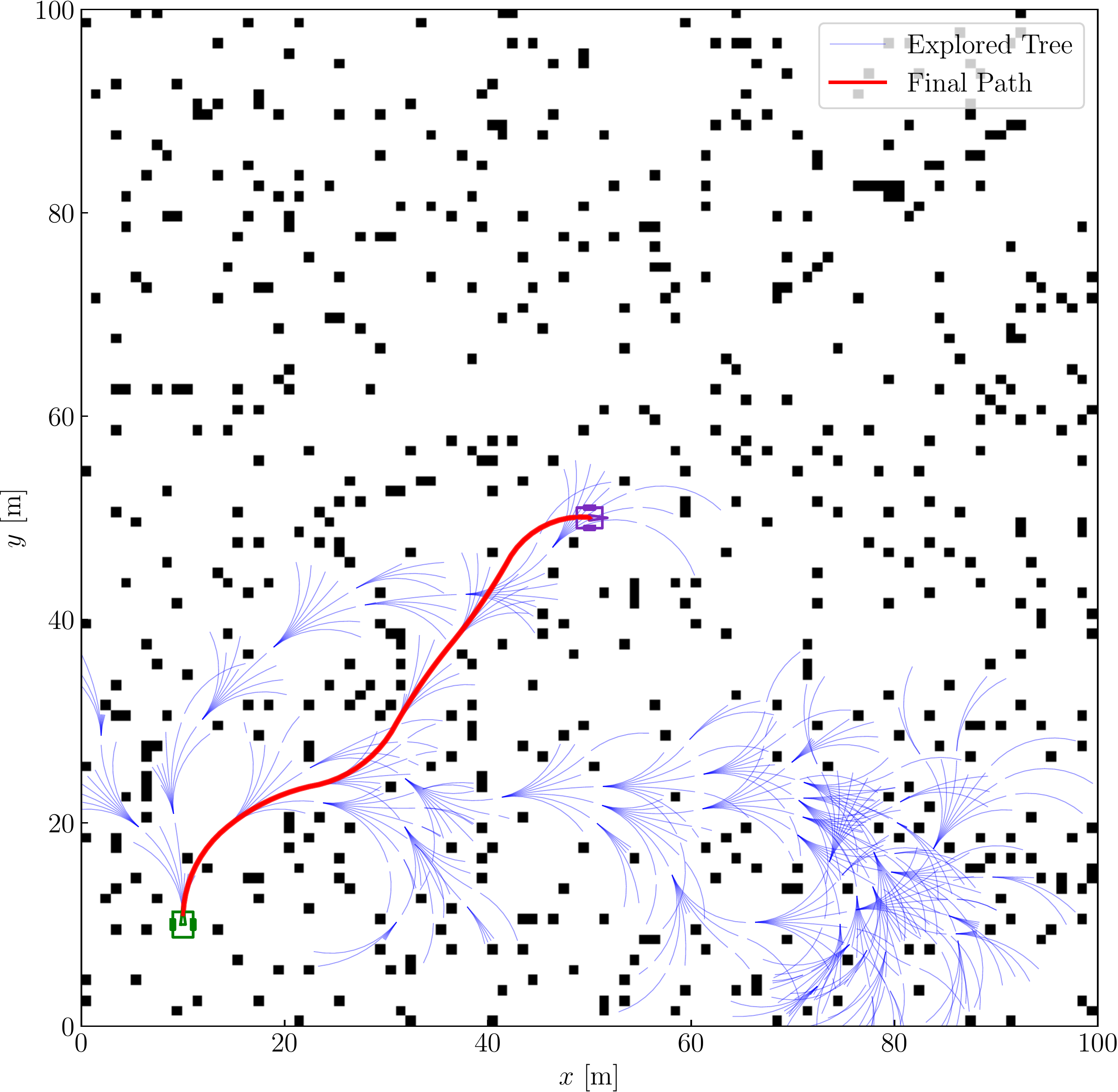}
        \vspace*{-6pt}
        \caption{MCTS, $\ell_{\Gamma}=\SI{60.13}{\metre}$}
        \label{fig:offline_shortest_random_mcts}
    \end{subfigure}%
    \begin{subfigure}[t]{0.245\textwidth}
        \centering
        \includegraphics[width=\linewidth,height=0.975\linewidth]{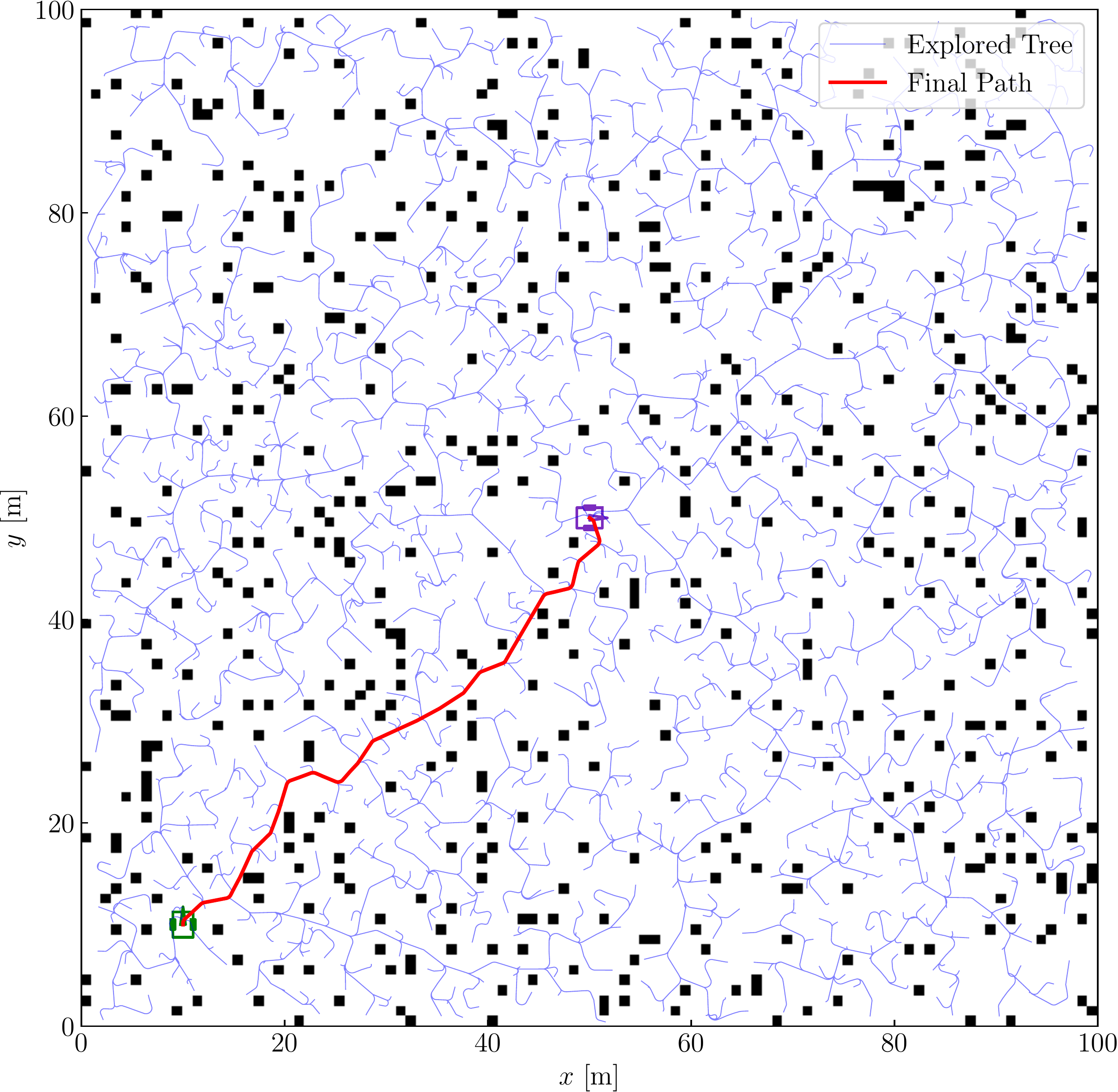}
        \vspace*{-6pt}
        \caption{LQR-RRT$^{*}$, $\ell_{\Gamma}=\SI{63.13}{\metre}$}
        \label{fig:offline_shortest_random_lqr}
    \end{subfigure}
    \begin{subfigure}[t]{0.245\textwidth}
        \centering
        \includegraphics[width=\linewidth,height=0.975\linewidth]{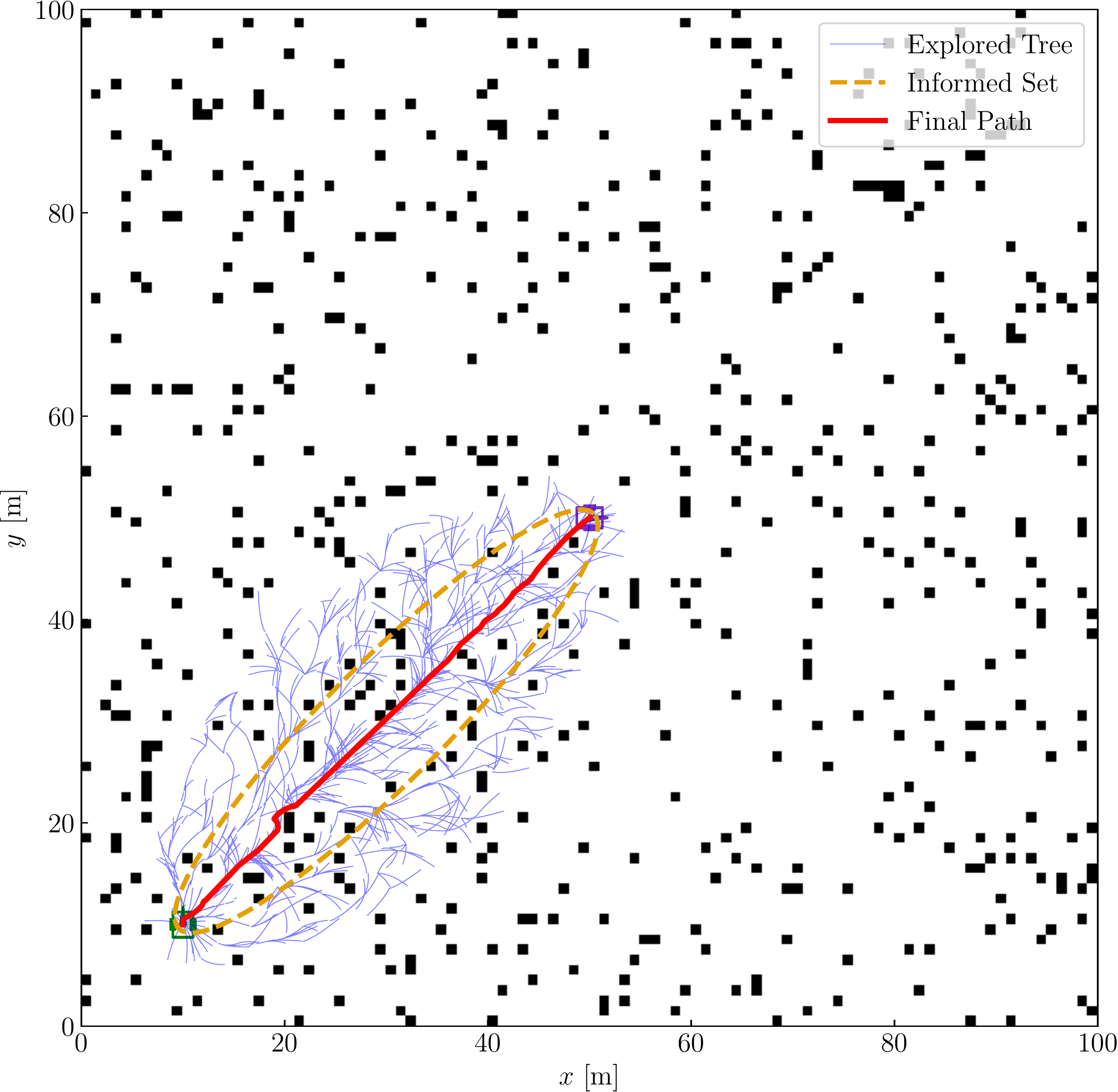}
        \vspace*{-6pt}
        \caption{IK-RRT$^{*}$, $\ell_{\Gamma}=\SI{57.75}{\metre}$}
        \label{fig:offline_shortest_random_kino}
    \end{subfigure}
    \begin{subfigure}[t]{0.245\textwidth}
        \centering
        \includegraphics[width=\linewidth,height=0.975\linewidth]{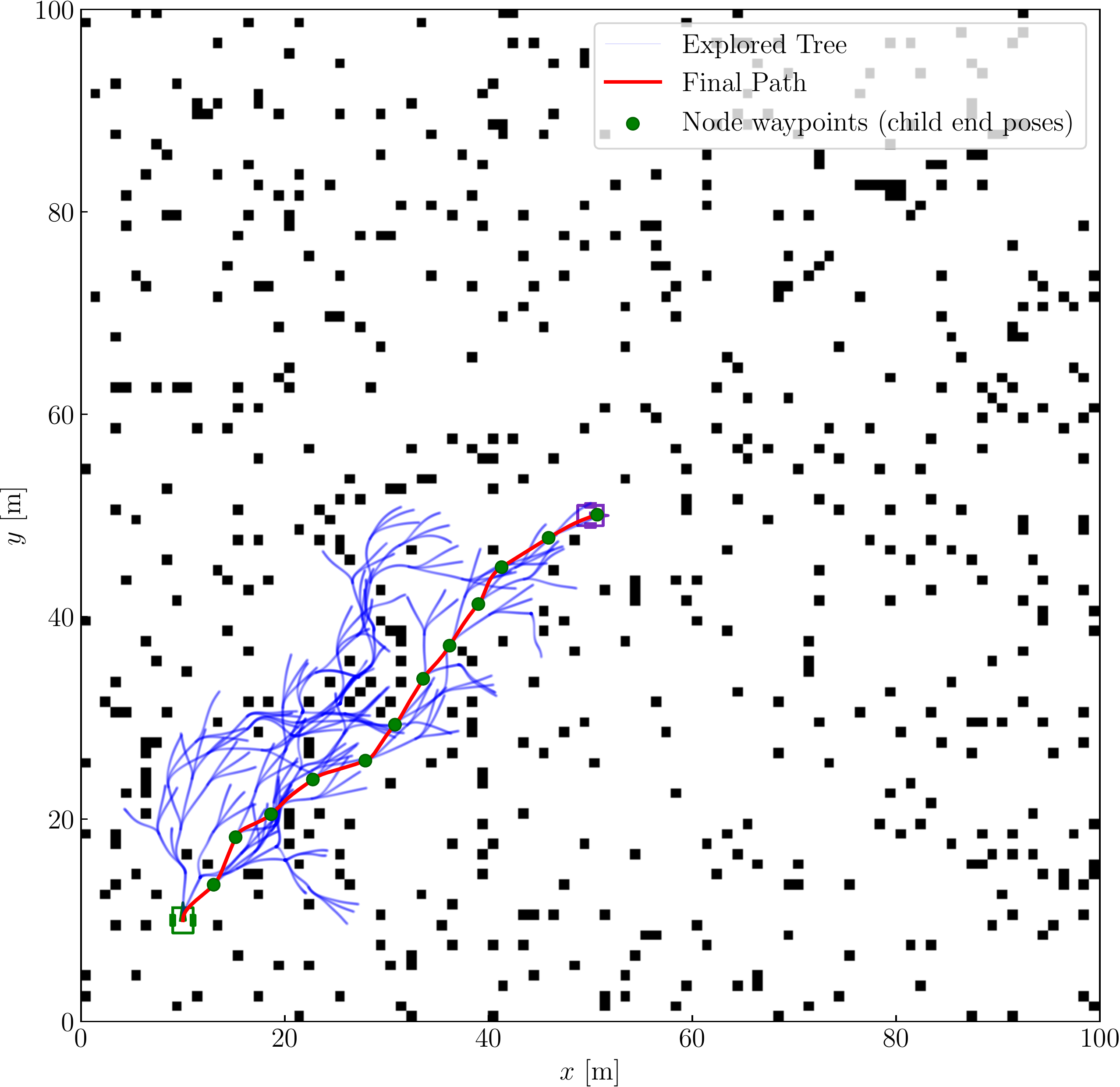}
        \vspace*{-6pt}
        \caption{SI-MPPI (Ours), $\ell_{\Gamma}=\SI{59.30}{\metre}$}
        \label{fig:offline_shortest_random_simppi}
    \end{subfigure}
    \vspace*{-6pt}
    \caption{Shortest planned trajectory (red) among five trials and its corresponding explored search tree (blue) for each planner on \mapname{Rand}.}
    \label{fig:offline_shortest_random}
    \vspace*{-7pt}
\end{figure*}

\begin{figure*}[!t]
    \centering
    \begin{subfigure}[t]{0.245\textwidth}
        \centering
        \includegraphics[width=\linewidth,height=0.975\linewidth]{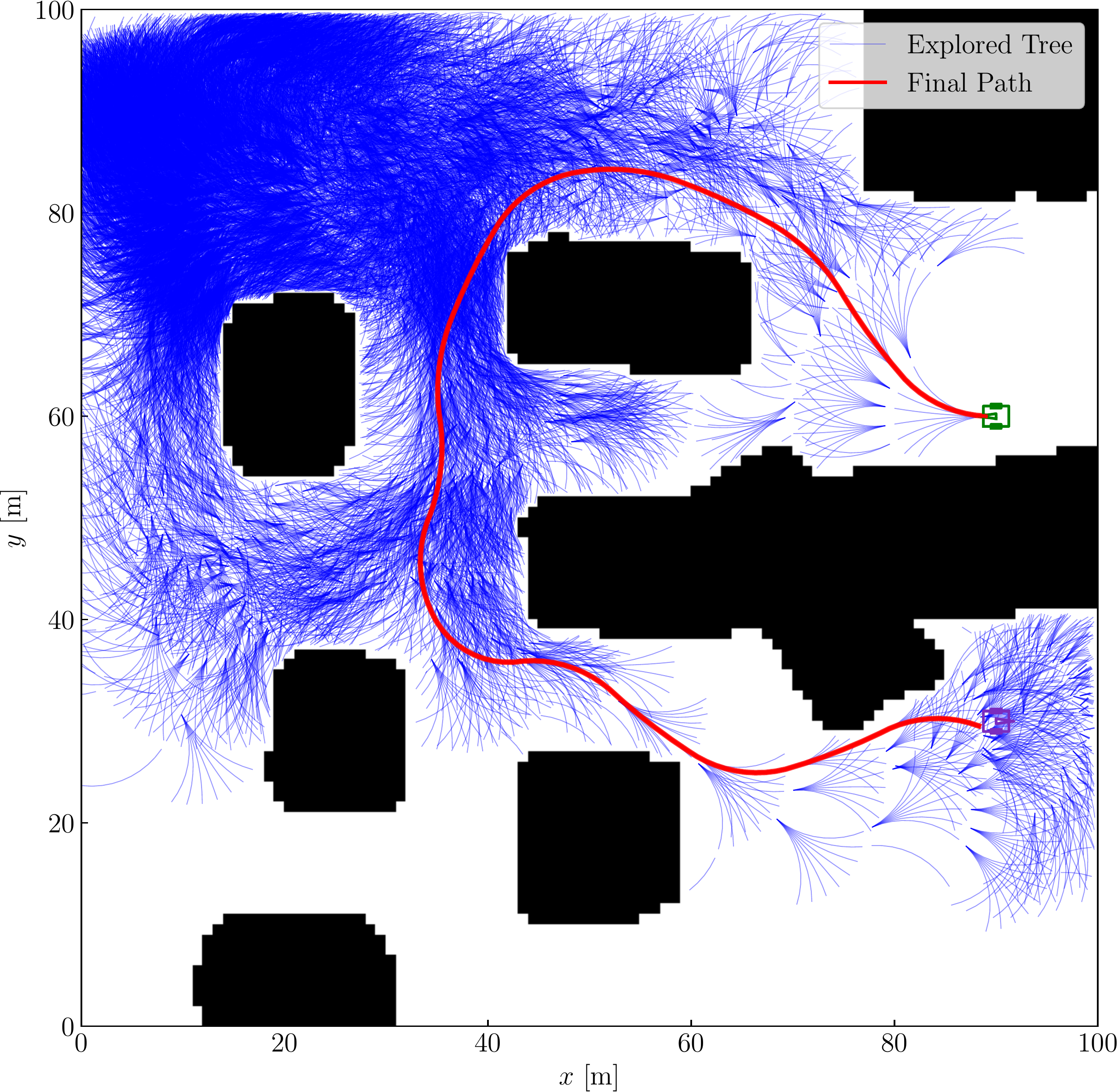}
        \vspace*{-6pt}
        \caption{MCTS, $\ell_{\Gamma}=\SI{160.14}{\metre}$}
        \label{fig:offline_shortest_cave_mcts}
    \end{subfigure}%
    \begin{subfigure}[t]{0.245\textwidth}
        \centering
        \includegraphics[width=\linewidth,height=0.975\linewidth]{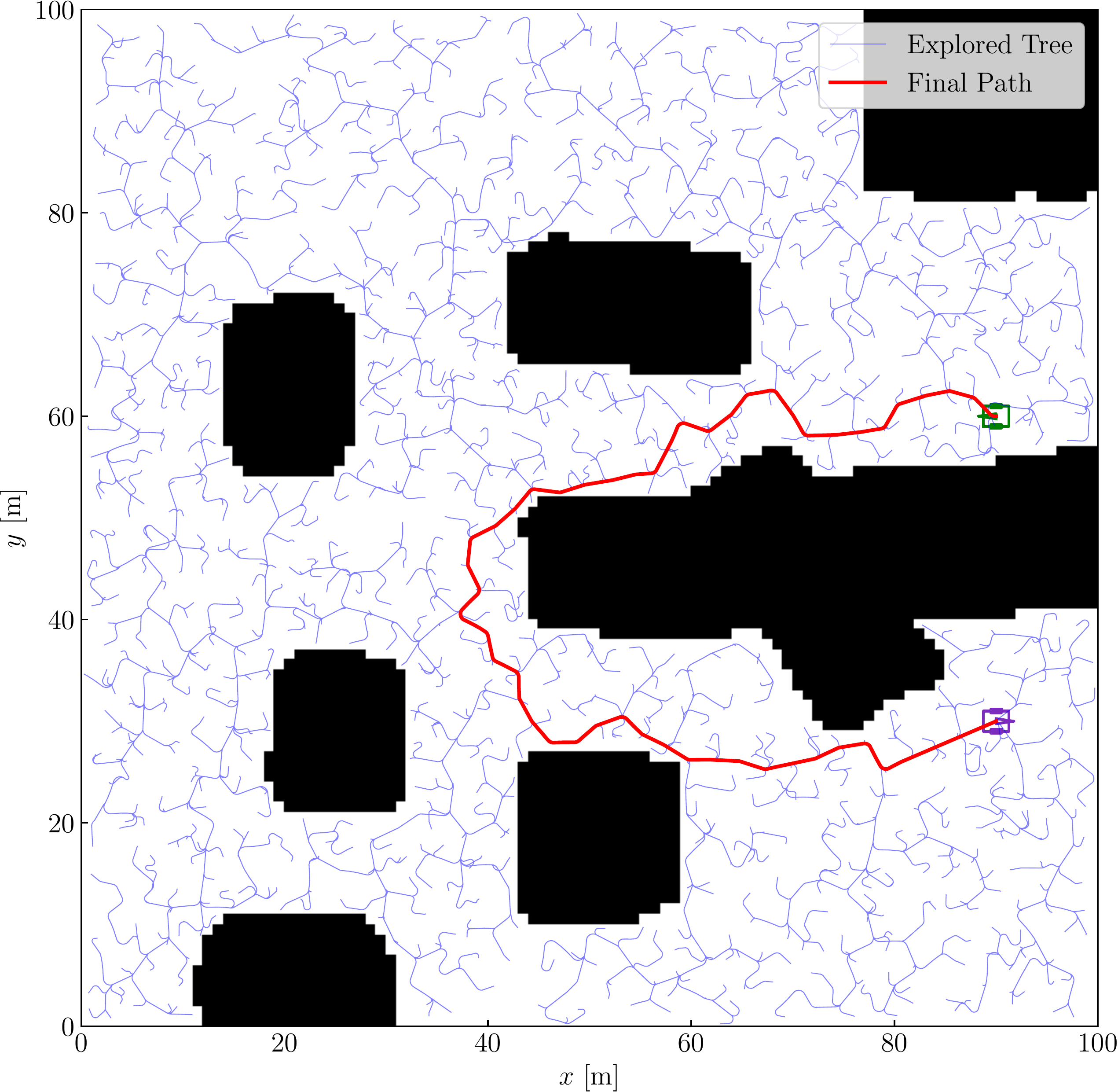}
        \vspace*{-6pt}
        \caption{LQR-RRT$^{*}$, $\ell_{\Gamma}=\SI{136.69}{\metre}$}
        \label{fig:offline_shortest_cave_lqr}
    \end{subfigure}
    \begin{subfigure}[t]{0.245\textwidth}
        \centering
        \includegraphics[width=\linewidth,height=0.975\linewidth]{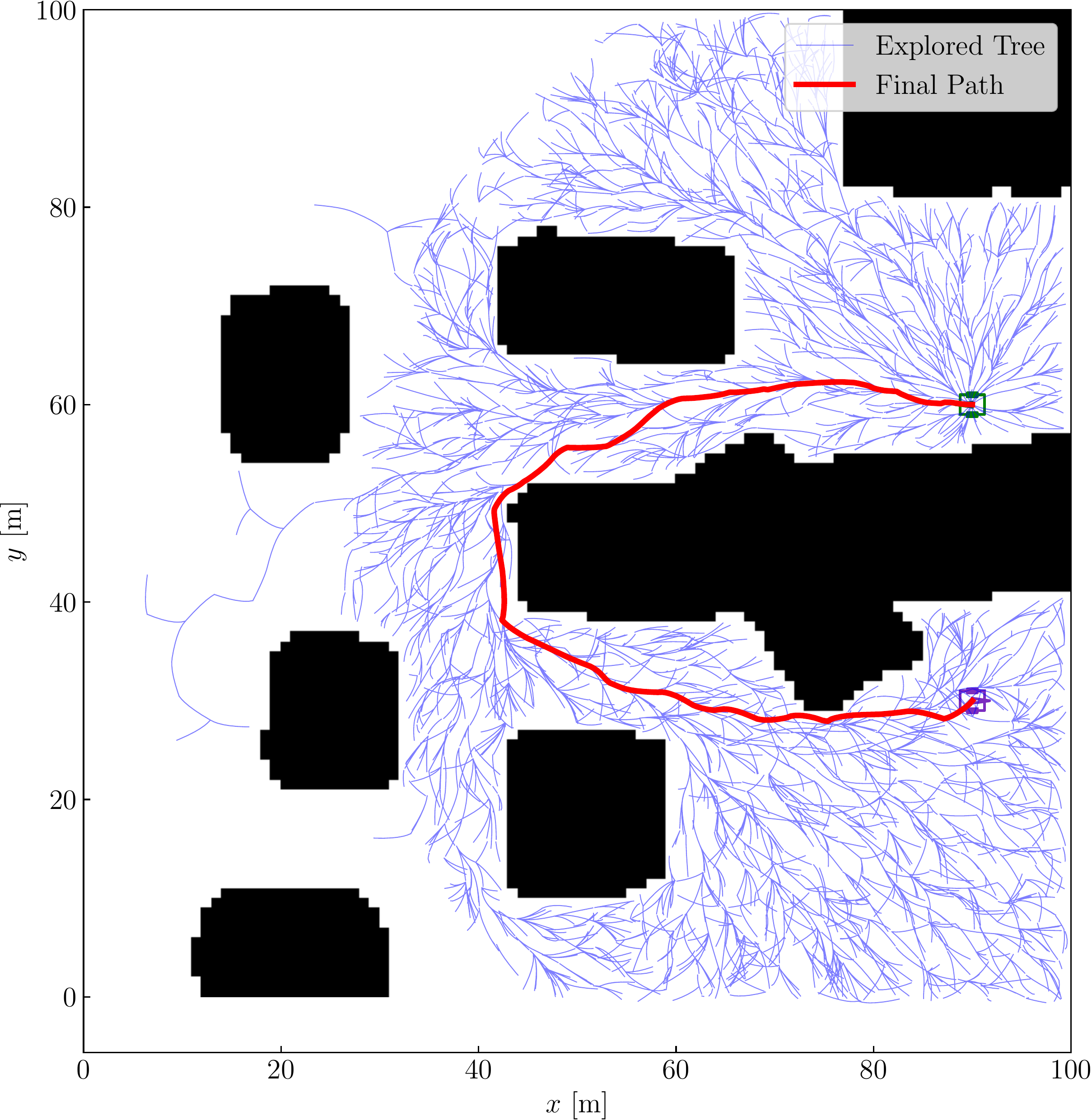}
        \vspace*{-6pt}
        \caption{IK-RRT$^{*}$, $\ell_{\Gamma}=\SI{115.30}{\metre}$}
        \label{fig:offline_shortest_cave_kino}
    \end{subfigure}
    \begin{subfigure}[t]{0.245\textwidth}
        \centering
        \includegraphics[width=\linewidth,height=0.975\linewidth]{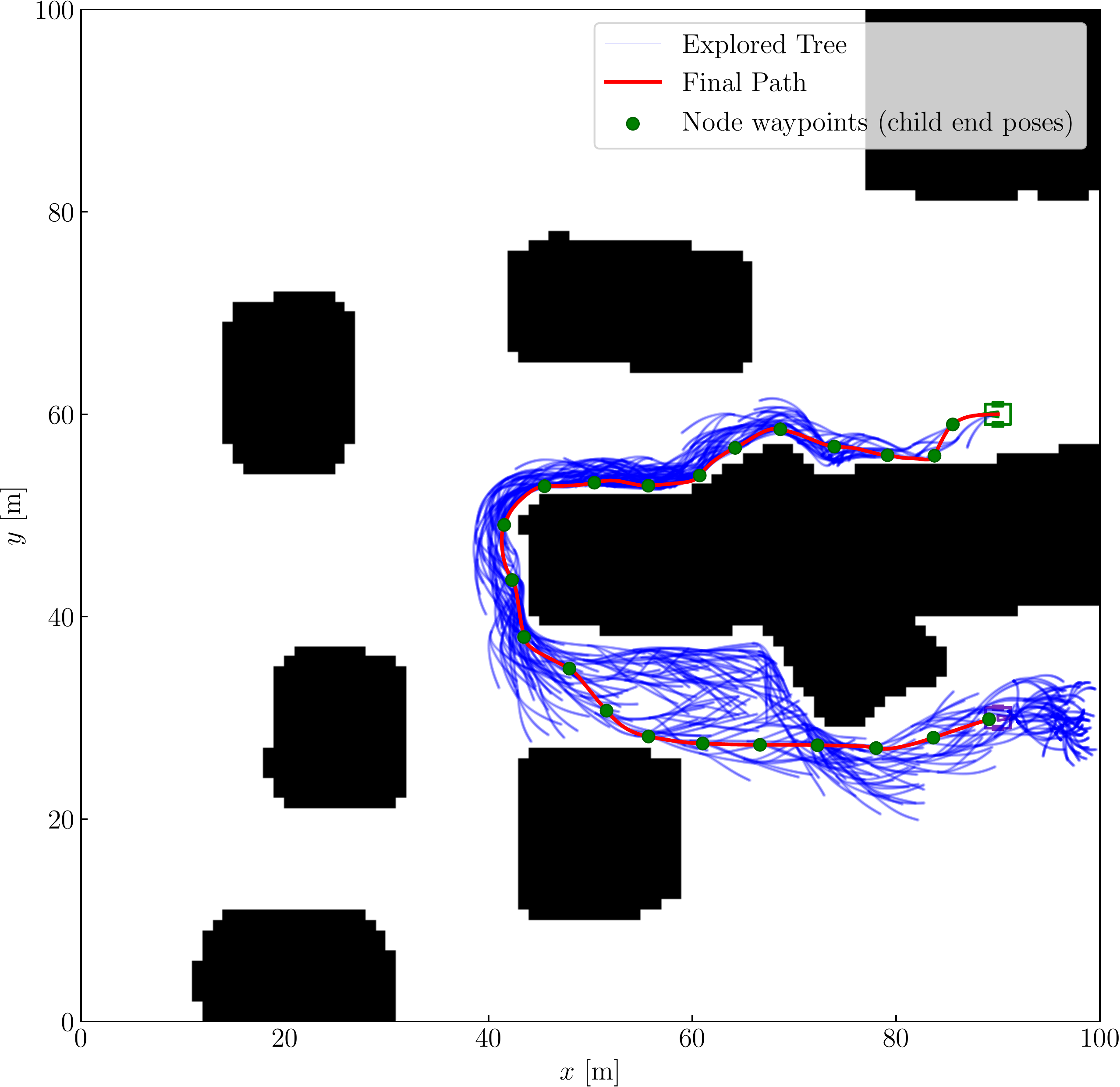}
        \vspace*{-6pt}
        \caption{SI-MPPI (Ours), $\ell_{\Gamma}=\SI{116.06}{\metre}$}
        \label{fig:offline_shortest_cave_simppi}
    \end{subfigure}
    \vspace*{-6pt}
    \caption{Shortest planned trajectory (red) among five trials and its corresponding explored search tree (blue) for each planner on \mapname{Cave}.}
    \label{fig:offline_shortest_cave}
    \vspace*{-7pt}
\end{figure*}

\begin{figure*}[!t]
    \centering
    \begin{subfigure}[t]{0.245\textwidth}
        \centering
        \includegraphics[width=\linewidth,height=0.975\linewidth]{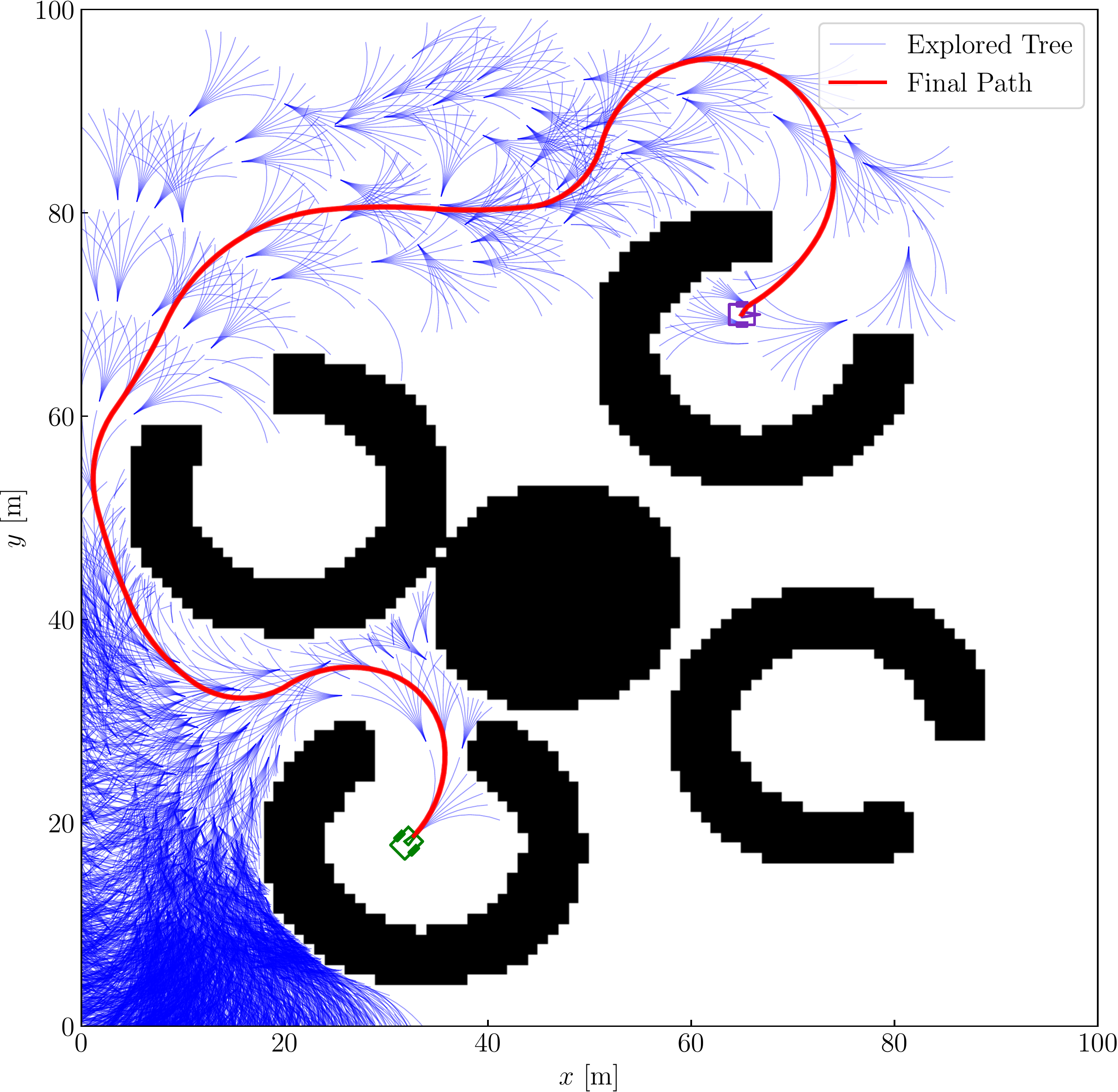}
        \vspace*{-6pt}
        \caption{MCTS, $\ell_{\Gamma}=\SI{181.68}{\metre}$}
        \label{fig:offline_shortest_four_moons_mcts}
    \end{subfigure}%
    \begin{subfigure}[t]{0.245\textwidth}
        \centering
        \includegraphics[width=\linewidth,height=0.975\linewidth]{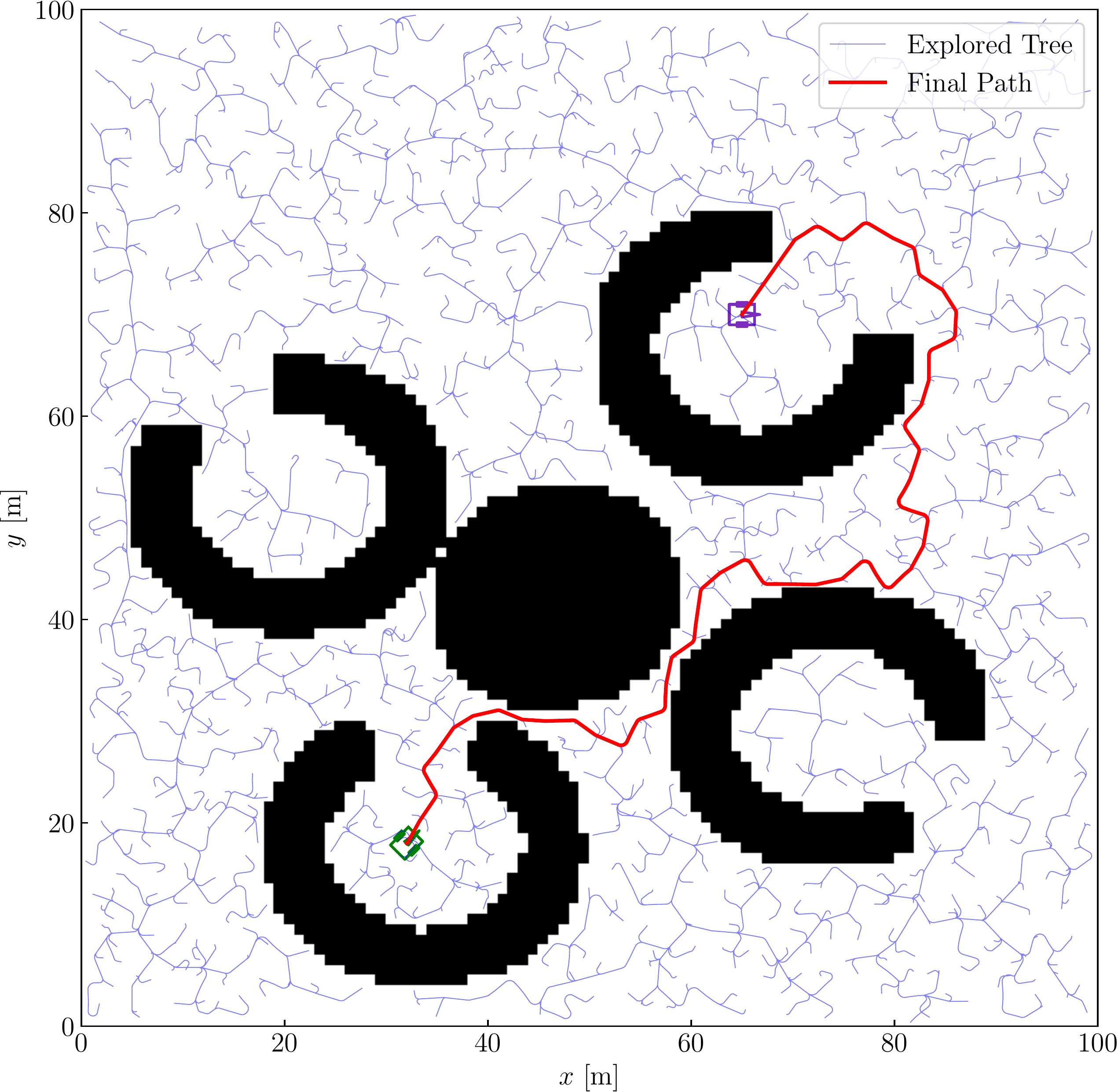}
        \vspace*{-6pt}
        \caption{LQR-RRT$^{*}$, $\ell_{\Gamma}=\SI{136.50}{\metre}$}
        \label{fig:offline_shortest_four_moons_lqr}
    \end{subfigure}
    \begin{subfigure}[t]{0.245\textwidth}
        \centering
        \includegraphics[width=\linewidth,height=0.975\linewidth]{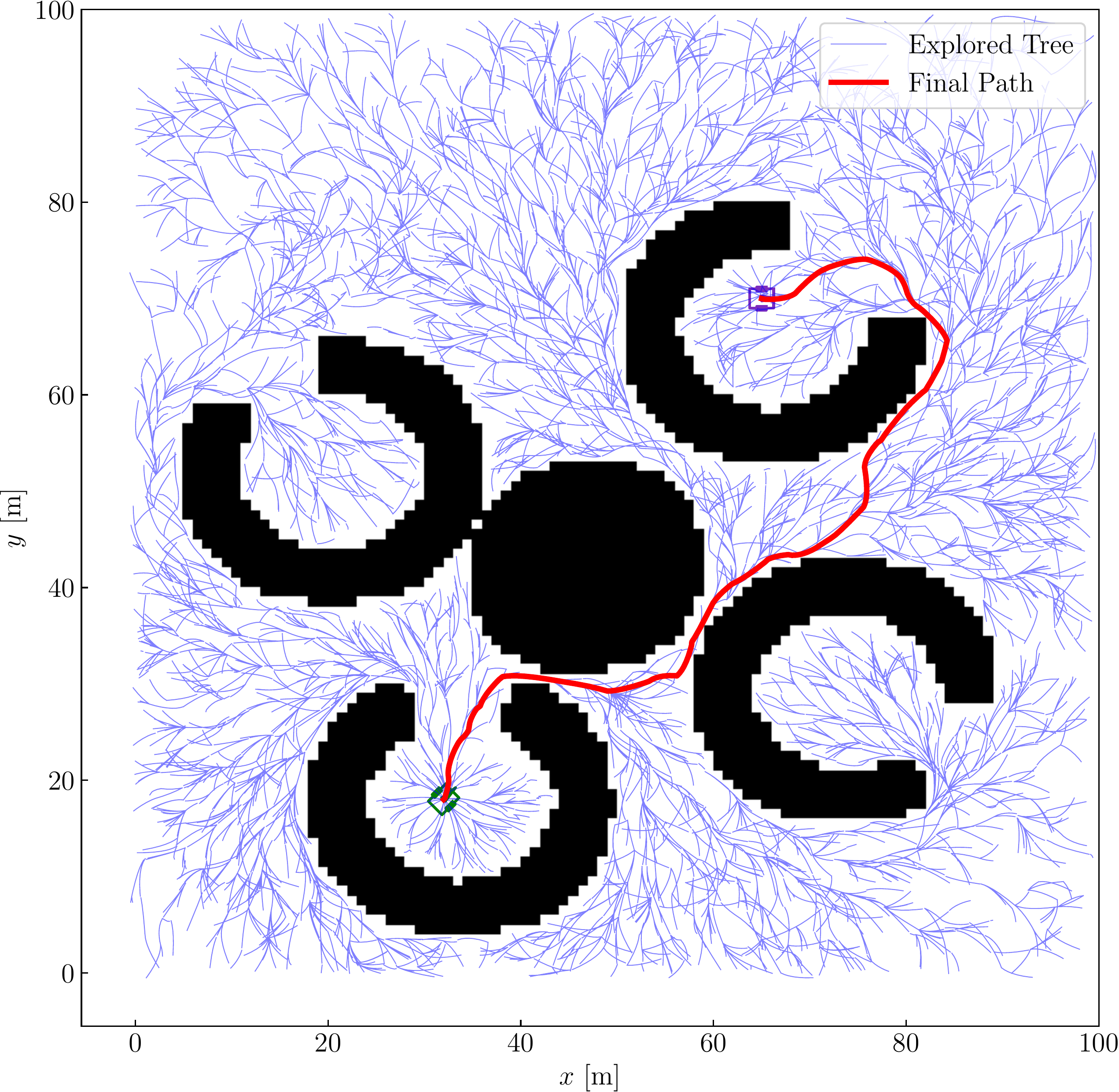}
        \vspace*{-6pt}
        \caption{IK-RRT$^{*}$, $\ell_{\Gamma}=\SI{105.40}{\metre}$}
        \label{fig:offline_shortest_four_moons_kino}
    \end{subfigure}
    \begin{subfigure}[t]{0.245\textwidth}
        \centering
        \includegraphics[width=\linewidth,height=0.975\linewidth]{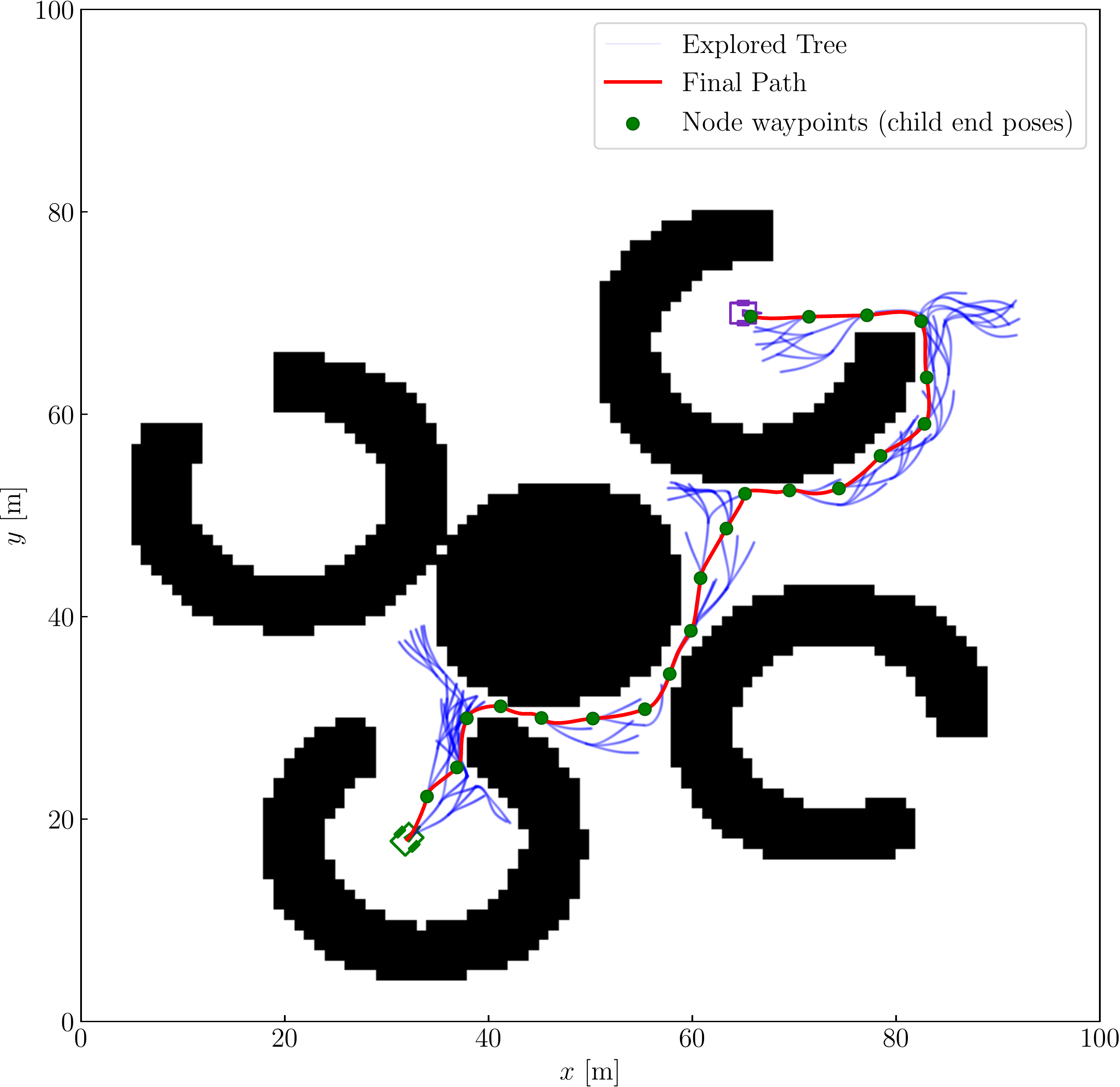}
        \vspace*{-6pt}
        \caption{SI-MPPI (Ours), $\ell_{\Gamma}=\SI{103.83}{\metre}$}
        \label{fig:offline_shortest_four_moons_simppi}
    \end{subfigure}
    \vspace*{-6pt}
    \caption{Shortest planned trajectory (red) among five trials and its corresponding explored search tree (blue) for each planner on \mapname{4Moons}.}
    \label{fig:offline_shortest_four_moons}
    \vspace*{-7pt}
\end{figure*}

\begin{figure*}[!t]
    \centering
    \begin{subfigure}[t]{0.245\textwidth}
        \centering
        \includegraphics[width=\linewidth,height=0.975\linewidth]{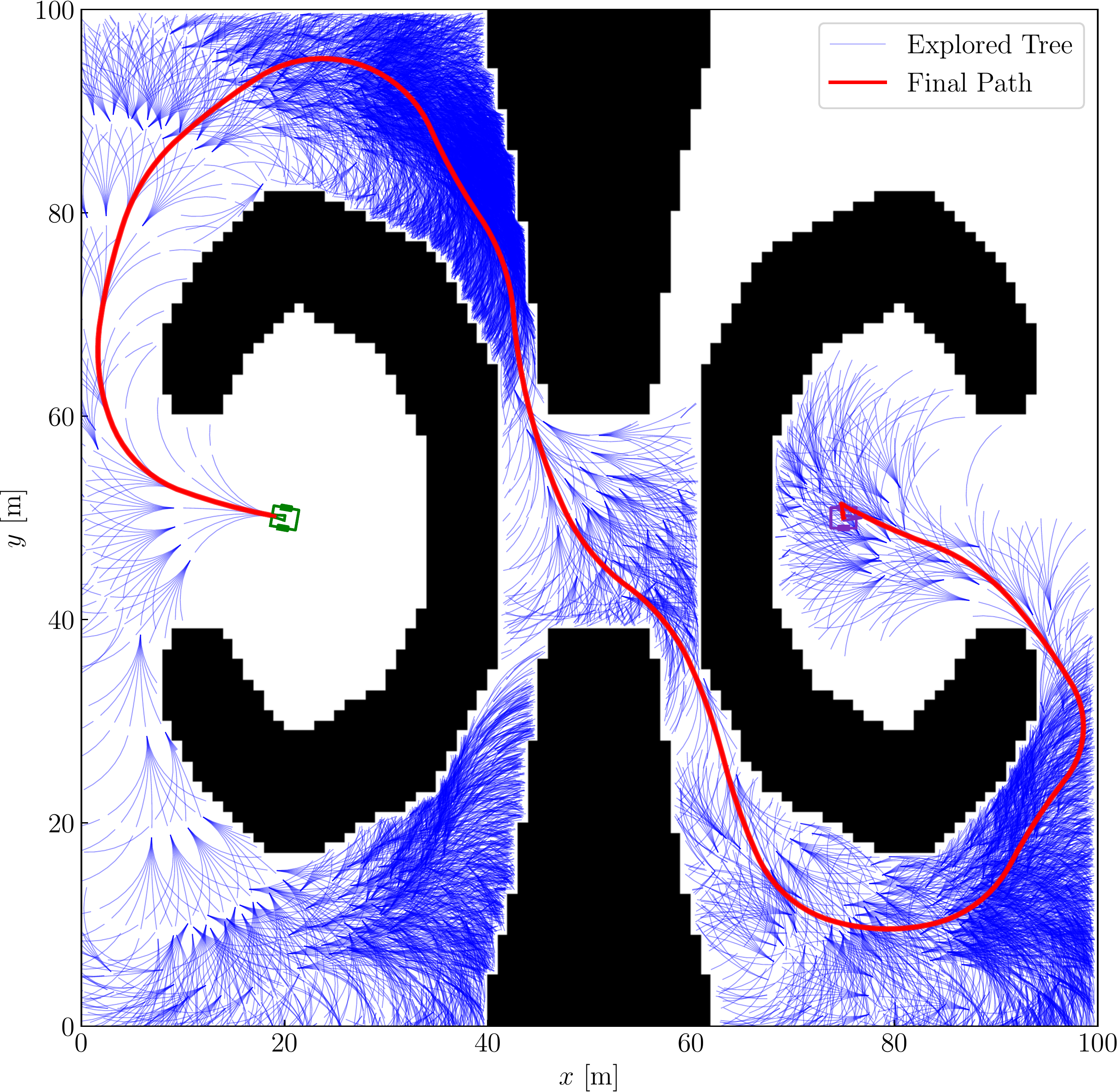}
        \vspace*{-6pt}
        \caption{MCTS, $\ell_{\Gamma}=\SI{243.44}{\metre}$}
        \label{fig:offline_shortest_moon_gate_mcts}
    \end{subfigure}%
    \begin{subfigure}[t]{0.245\textwidth}
        \centering
        \includegraphics[width=\linewidth,height=0.975\linewidth]{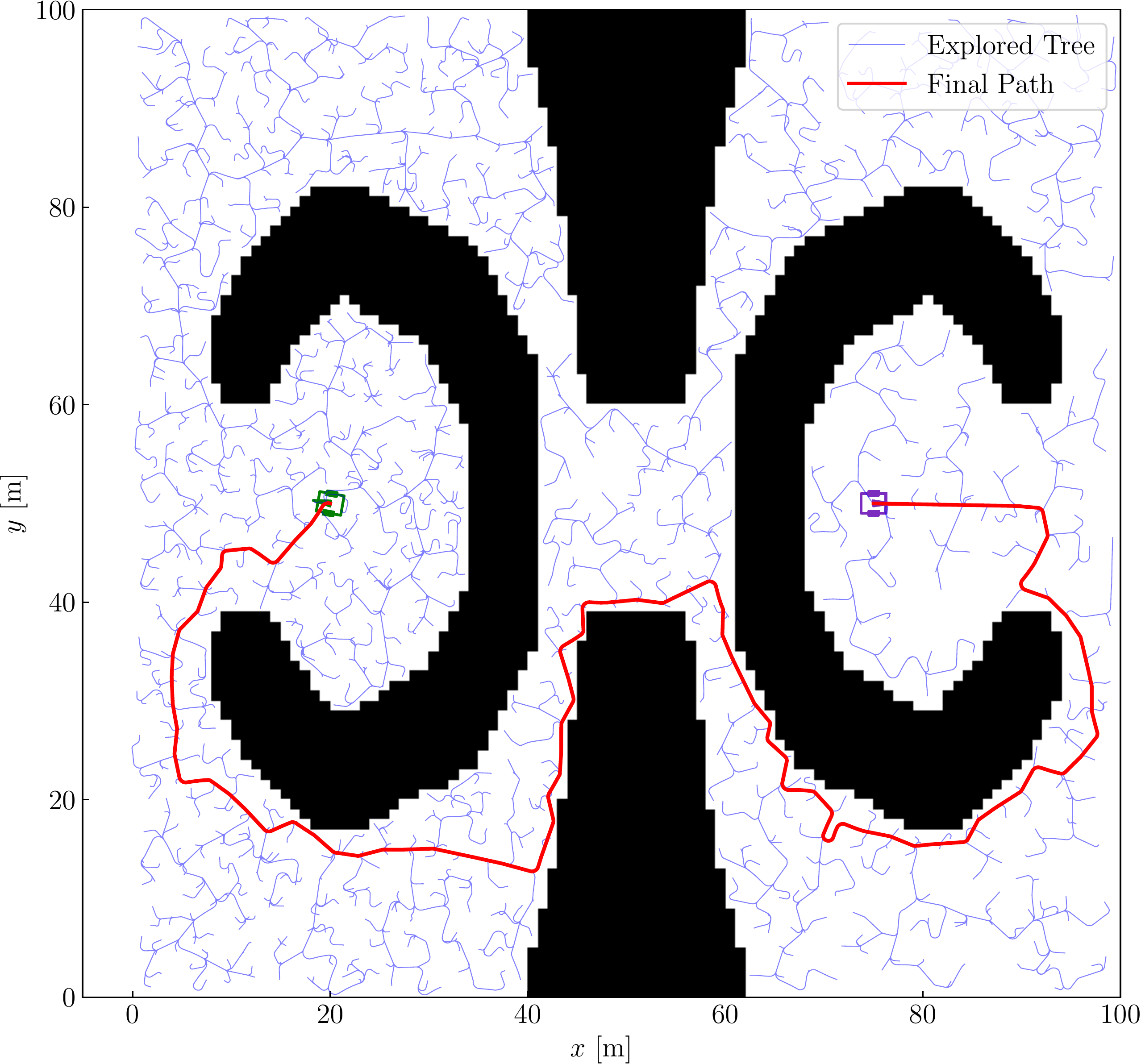}
        \vspace*{-6pt}
        \caption{LQR-RRT$^{*}$, $\ell_{\Gamma}=\SI{232.13}{\metre}$}
        \label{fig:offline_shortest_moon_gate_lqr}
    \end{subfigure}
    \begin{subfigure}[t]{0.245\textwidth}
        \centering
        \includegraphics[width=\linewidth,height=0.975\linewidth]{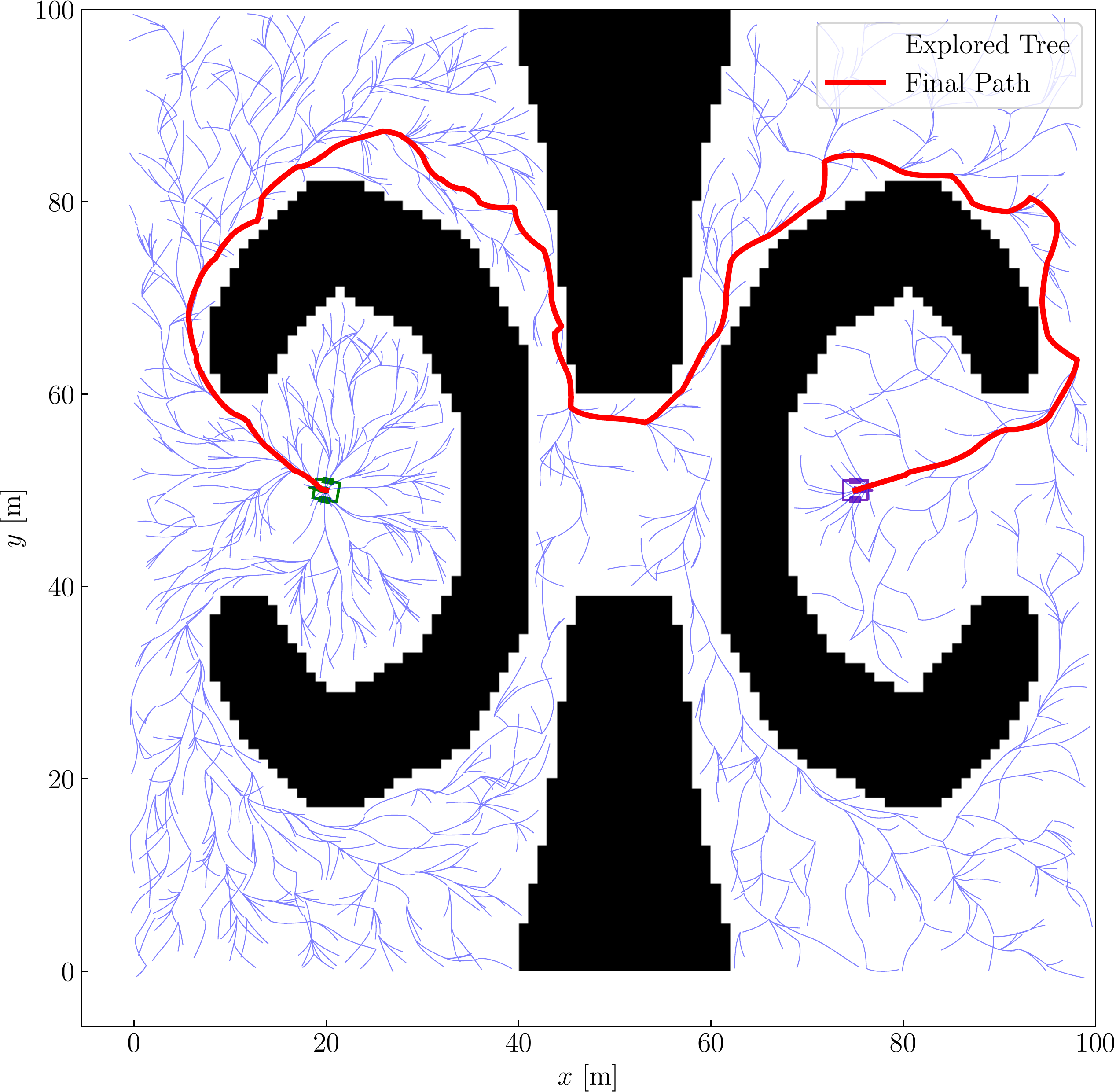}
        \vspace*{-6pt}
        \caption{IK-RRT$^{*}$, $\ell_{\Gamma}=\SI{209.43}{\metre}$}
        \label{fig:offline_shortest_moon_gate_kino}
    \end{subfigure}
    \begin{subfigure}[t]{0.245\textwidth}
        \centering
        \includegraphics[width=\linewidth,height=0.975\linewidth]{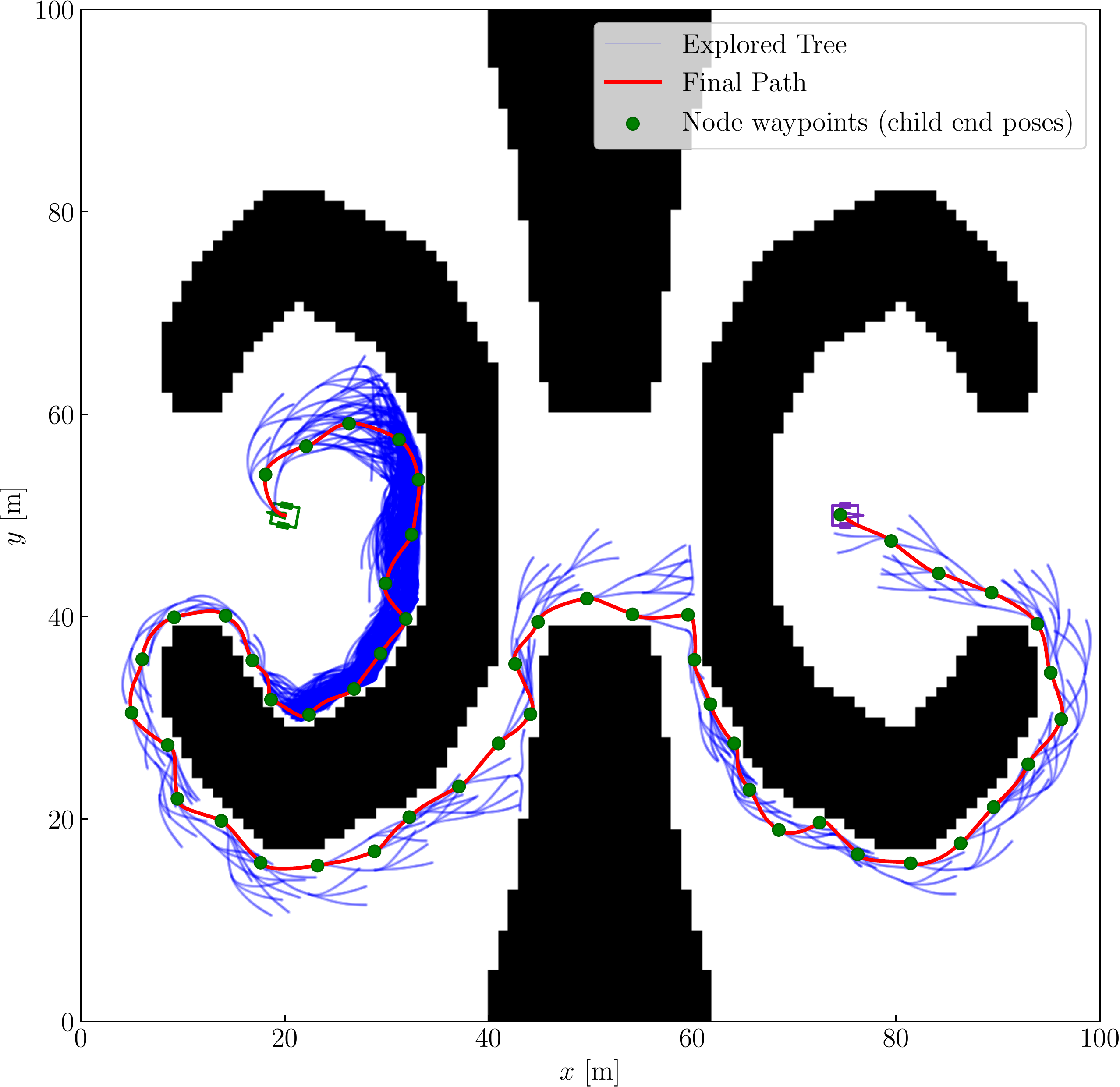}
        \vspace*{-6pt}
        \caption{SI-MPPI (Ours), $\ell_{\Gamma}=\SI{255.37}{\metre}$}
        \label{fig:offline_shortest_moon_gate_simppi}
    \end{subfigure}
    \vspace*{-6pt}
    \caption{Shortest planned trajectory (red) among five trials and its corresponding explored search tree (blue) for each planner on \mapname{MGate}.}
    \label{fig:offline_shortest_moon_gate}
\end{figure*}

Table~\ref{tab:offline_benchmark_means} reports the offline planning performance of SI-MPPI and the three kinodynamic baselines over five trials per planner on each of the four benchmark maps, with Figs.~\ref{fig:offline_shortest_random}--\ref{fig:offline_shortest_moon_gate} visualizing the trajectories and explored trees from the trial yielding the shortest path in each case.\footnote{The IK-RRT$^{*}$ informed-sampling ellipse, shown in yellow in Fig.~\ref{fig:offline_shortest_random_kino}, is omitted from the subsequent IK-RRT$^{*}$ panels for visual clarity.}
 In these trials, all planners successfully generate collision-free, goal-reaching trajectories through the cluttered environments (i.e., $P_{\mathrm{succ}}=100\%$) under the specified benchmark settings; accordingly, the comparison focuses on search complexity and trajectory quality. Across all maps, SI-MPPI achieves the lowest angular jitter $\sigma_{\omega}$ and RMS angular acceleration $a_{\omega,\mathrm{rms}}$, while maintaining the smallest search trees in terms of $N_{\mathrm{tree}}$. In particular, its $a_{\omega,\mathrm{rms}}$ is lower by a factor of approximately $4.8$--$7.9$ relative to MCTS and $22$--$37$ relative to the RRT$^{*}$-based baselines, owing to the combined effects of cost-weighted MPPI control updates, the control-effort penalty, and Savitzky--Golay filtering applied to the optimized control sequences. Moreover, SI-MPPI generates smaller search trees than all three baselines on every map, frequently by more than an order of magnitude. This compactness reflects both the top-$k$ admission rule, which retains at most $k_b=3$ candidates per expansion, and the representation of each full-horizon, MPPI-optimized motion segment by a single child node. In contrast, no planner consistently achieves the lowest planning time $T_{\mathrm{plan}}$ and path length $\ell_{\Gamma}$ across all maps. The remainder of this analysis therefore examines these map-dependent trade-offs.

In terms of planning time $T_{\mathrm{plan}}$, SI-MPPI achieves the lowest average planning time on \mapname{Cave} ($T_{\mathrm{plan}}=\SI{5.48}{\second}$) and ranks second on \mapname{Rand} and \mapname{4Moons}, whereas MCTS leads on the remaining three maps, benefiting from its relatively inexpensive discrete macro-action expansions. Notably, on \mapname{4Moons}, SI-MPPI and MCTS achieve comparable planning times (\SI{3.95}{\second} and \SI{3.37}{\second}, respectively), yet SI-MPPI yields the shortest mean path length among all planners ($\ell_{\Gamma}=\SI{111.69}{\metre}$), approximately $47\%$ shorter than that of MCTS ($\ell_{\Gamma}=\SI{209.79}{\metre}$).
On \mapname{MGate}, however, SI-MPPI requires its longest average planning time ($T_{\mathrm{plan}}=\SI{19.98}{\second}$), as the initially unfavorable heading and narrow gated passages lead to extensive exploration around the left C-shaped obstacle, as illustrated in Fig.~\ref{fig:offline_shortest_moon_gate_simppi} and reflected in a larger search tree ($N_{\mathrm{tree}}=1290.40$ nodes on average), approximately four to six times that on the other maps. 
Apart from IK-RRT$^{*}$ on \mapname{MGate}, both RRT$^{*}$-based baselines require considerably longer planning times than our proposed kinodynamic framework, as they continue sampling, generating kinodynamically feasible connections, and rewiring their search trees until their iteration budgets are exhausted rather than terminating upon finding a feasible goal-reaching plan, while exploring substantial regions beyond the final solution paths (see Figs.~\ref{fig:offline_shortest_random}--\ref{fig:offline_shortest_moon_gate}).
The convergence time $T_{\mathrm{conv}}$ provides complementary insight into the RRT$^{*}$-based baselines. On \mapname{Rand} and \mapname{4Moons}, SI-MPPI's average $T_{\mathrm{plan}}$ is lower than the average $T_{\mathrm{conv}}$ of both RRT$^{*}$-based baselines. By contrast, LQR-RRT$^{*}$ reaches its final best cost on \mapname{Cave} after only \SI{1.33}{\second} on average, but yields an average path length $\ell_{\Gamma}$ approximately $23\%$ longer than that of SI-MPPI. On \mapname{MGate}, IK-RRT$^{*}$ reaches its final best cost after \SI{5.69}{\second} on average, yielding the shortest mean path length among all planners. Moreover, for LQR-RRT$^{*}$, the ratio of average $T_{\mathrm{conv}}$ to average $T_{\mathrm{plan}}$ remains below $9\%$ on \mapname{Rand}, \mapname{Cave}, and \mapname{4Moons}, indicating that most of its total planning time elapses after the last observed improvement in solution cost. Nevertheless, since the RRT$^{*}$-based planners cannot know during execution whether their best solution will improve further, $T_{\mathrm{conv}}$ can only be determined after the search terminates; thus, $T_{\mathrm{plan}}$ remains the appropriate measure of their actual planning time.

\begin{figure}[!t]
    \centering
    \begin{subfigure}[t]{0.48\columnwidth}
        \centering
        \includegraphics[width=\linewidth]{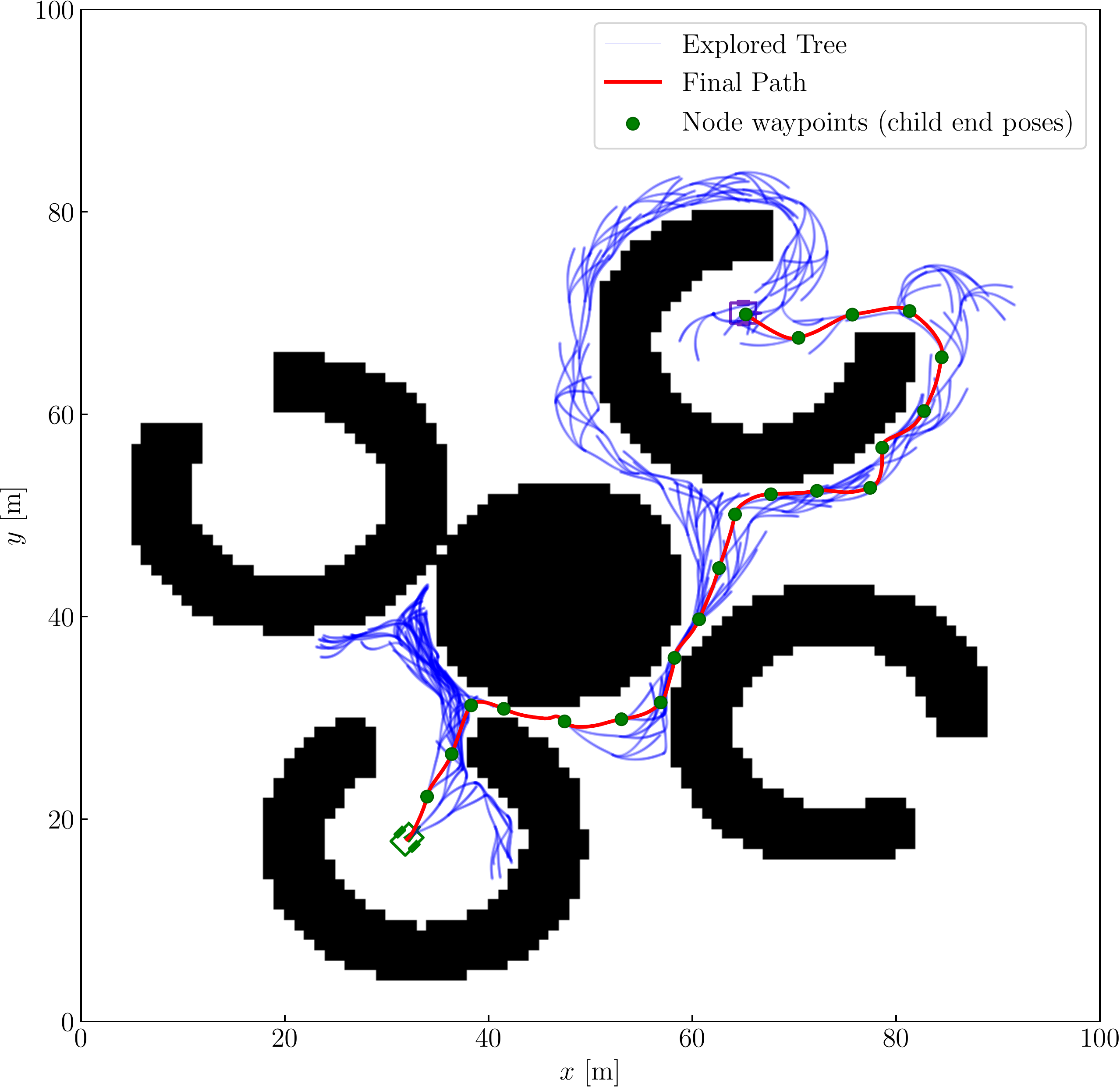}
        \vspace*{-6pt}
        \caption{\mapname{4Moons}, $\ell_{\Gamma}=\SI{107.78}{\metre}$}
        \label{fig:offline_simppi_four_moons_homotopy}
    \end{subfigure}%
    \hfill
    \begin{subfigure}[t]{0.48\columnwidth}
        \centering
        \includegraphics[width=\linewidth]{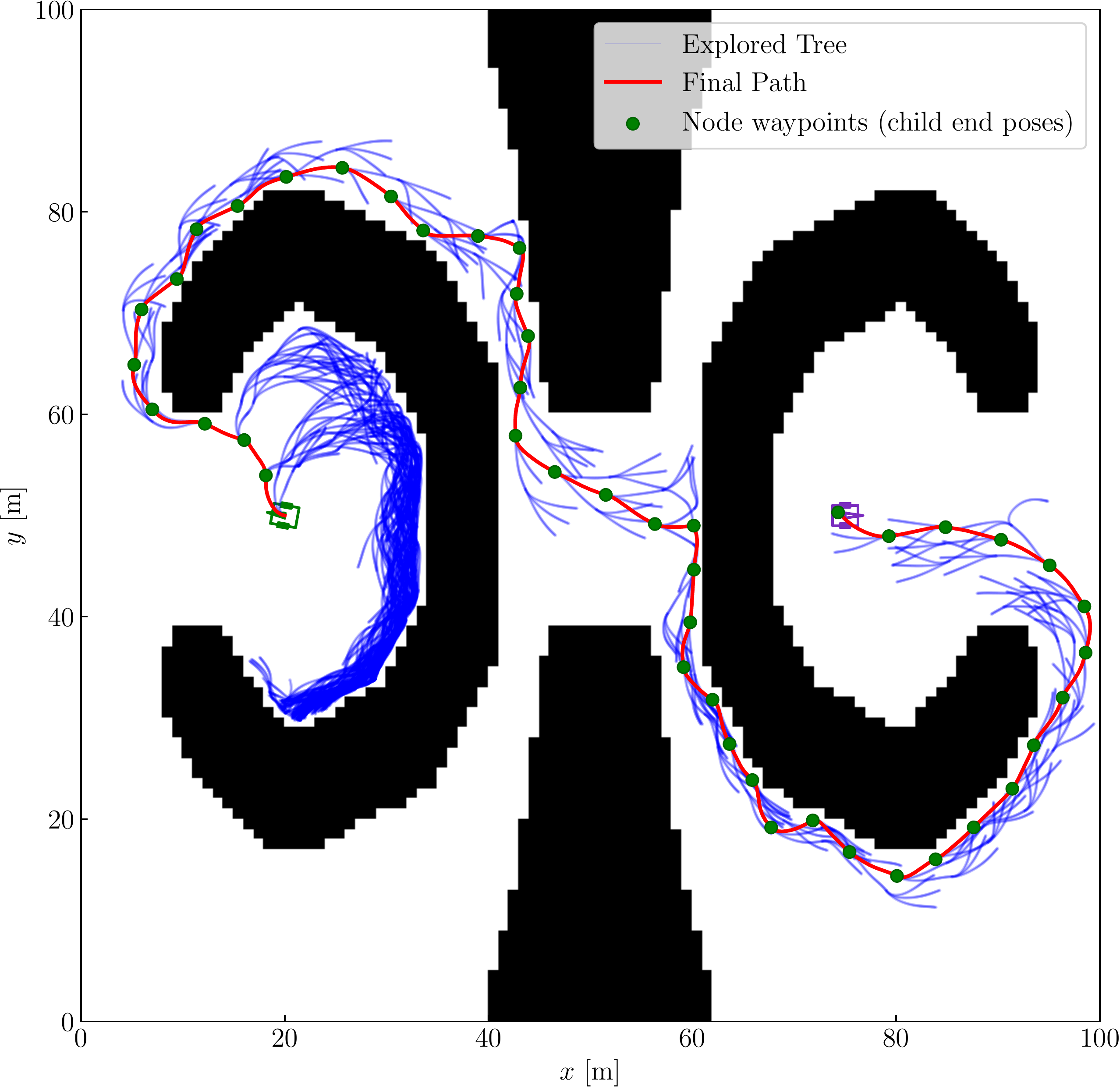}
        \vspace*{-6pt}
        \caption{\mapname{MGate}, $\ell_{\Gamma}=\SI{224.75}{\metre}$}
        \label{fig:offline_simppi_moon_gate_seed1}
    \end{subfigure}
    \vspace*{-6pt}
   \caption{Selected SI-MPPI trials illustrating (a) exploration of alternative branches and selection of a shorter interior route on \mapname{4Moons}, and (b) a more direct route on \mapname{MGate} than that shown in Fig.~\ref{fig:offline_shortest_moon_gate_simppi}.}
    \label{fig:offline_simppi_homotopy}
\end{figure}

Turning to path length $\ell_{\Gamma}$, all four planners achieve relatively comparable results on \mapname{Rand}, whereas the more structured obstacle configurations in the remaining maps reveal greater differences in route selection and path efficiency. In particular, SI-MPPI achieves the shortest average path length $\ell_{\Gamma}$ and the lowest standard deviation on both \mapname{Cave} and \mapname{4Moons}, outperforming the next-best planner, IK-RRT$^{*}$, marginally on the former and by approximately $19\%$ on the latter. 
On the other hand, IK-RRT$^{*}$ returns the shortest average $\ell_{\Gamma}$ on \mapname{MGate}, benefiting from its direct minimization of path length through informed sampling and rewiring, while SI-MPPI yields the longest average $\ell_{\Gamma}$, approximately $18\%$ longer than that of IK-RRT$^{*}$. As described in Section~\ref{subsec:expansion}, the forward-oriented baits guide local exploration in SI-MPPI, while the terminal distance-to-goal penalty in~\eqref{eq:term_cost} favors candidate motion segments that terminate closer to the goal.
This behavior is evident on \mapname{4Moons}, where SI-MPPI consistently selects the shorter interior passage across all five trials, as illustrated in Fig.~\ref{fig:offline_shortest_four_moons_simppi}, whereas the three baselines more frequently follow longer detours around the C-shaped obstacles, consistent with their larger standard deviations. Fig.~\ref{fig:offline_simppi_four_moons_homotopy} further illustrates how SI-MPPI explores alternative branches around the upper C-shaped obstacle while selecting the shorter interior route.
However, this goal-directed expansion mechanism does not explicitly minimize the length of the complete path, as the resulting route also depends on stochastic MPPI sampling and subsequent tree-search decisions. On \mapname{MGate}, for example, the SI-MPPI solution shown in Fig.~\ref{fig:offline_shortest_moon_gate_simppi} follows a gradual detour around the left C-shaped obstacle, highlighting the path-length cost of this route selection. Nevertheless, an additional run with winner-cost backpropagation as defined in~\eqref{eq:backpropagation_cost} and random seed $1$ yields a substantially shorter path ($\ell_{\Gamma}=\SI{224.75}{\metre}$), as shown in Fig.~\ref{fig:offline_simppi_moon_gate_seed1}, demonstrating that SI-MPPI can also discover more direct routes under alternative search configurations.

Despite the $100\%$ success rates in the reported benchmark trials, additional tests reveal sensitivity of some baselines to changes in planning conditions.\footnote{It is worth noting that classical MPPI~\cite{williams2017model} was additionally evaluated but succeeded only on \mapname{Rand} under the evaluated settings, highlighting the benefit of embedding MPPI within a persistent global search tree.} Specifically, MCTS occasionally fails to return feasible plans or becomes trapped in local minima on \mapname{Cave}, with similar failures observed on \mapname{4Moons} when the initial heading is changed from \SI{45}{\degree} to \SI{-45}{\degree} or \SI{-90}{\degree}, potentially due to the limited turning angles of its discrete steering macro-actions. Likewise, the iteration budget for LQR-RRT$^{*}$ was increased from $I_{\max}=5000$ to $I_{\max}=8000$ to reliably obtain feasible plans on \mapname{MGate}. In contrast, SI-MPPI employs a common parameter configuration across all four maps. To complement these observations on planning robustness, we further aggregate the five-trial means in Table~\ref{tab:offline_benchmark_means} across the four benchmark maps, assigning equal weight to each environment. On this basis, SI-MPPI reduces the average tree size $N_{\mathrm{tree}}$ from $8032.90$ to $521.35$ nodes ($15.4\times$) and the average path length $\ell_{\Gamma}$ from $\SI{177.57}{\metre}$ to $\SI{138.90}{\metre}$ ($21.8\%$) relative to kinodynamic MCTS, owing to the combined effects of top-$k$ candidate admission, which limits tree growth, and bait-guided MPPI optimization, which promotes goal-directed, low-cost motion generation; these gains come with an additional $\SI{2.41}{\second}$ in average planning time $T_{\mathrm{plan}}$ ($\SI{8.38}{\second}$ vs. $\SI{5.97}{\second}$). Compared with LQR-RRT$^{*}$ and IK-RRT$^{*}$, the corresponding reductions in average $T_{\mathrm{plan}}$ are $26.8\times$ and $9.7\times$, respectively, while average $\ell_{\Gamma}$ is $14.7\%$ shorter than that of LQR-RRT$^{*}$ and only $2.7\%$ longer than that of IK-RRT$^{*}$. Taken together, these results demonstrate that our proposed framework achieves compact global search and competitive path quality without the substantial planning-time overhead of the evaluated RRT$^{*}$-based baselines.

\subsubsection{Ablation Analysis}
\label{subsubsec:ablation_analysis}

To further investigate how the planning performance of SI-MPPI is affected by its algorithmic components and parameter settings, we conduct five ablation tests across the four benchmark maps by systematically varying the default configuration specified in Section~\ref{subsubsec:planning_setup}. Specifically, in \textit{Test~\#1}, we examine the two settings of the backpropagation mode $\mathsf{mode}_{\mathrm{bp}}$ in~\eqref{eq:backpropagation_cost}, namely $\mathrm{winner}$ (\texttt{W}) and $\mathrm{expected}$ (\texttt{E}), with the optional heading-alignment setting enabled (\texttt{H}) or disabled (\texttt{NH}).
Unlike the default configuration, in which $\operatorname{Prog}(c^{(j)})$ in~\eqref{eq:accepted_set} relies solely on the normalized-displacement criterion in~\eqref{eq:candidate_progress}, enabling this setting allows $\operatorname{Prog}(c^{(j)})$ to be satisfied by either the displacement criterion or a reduction of at least $\SI{10}{\degree}$ in the absolute heading error relative to $\theta_{\mathrm{b}}^{(j)}$, measured from the selected node $n_{\mathrm{s}}$ to the candidate endpoint $\mathbf{x}_{N}^{(j)}$.
Consequently, a candidate $c^{(j)}$ with $\rho^{(j)}<\eta_{\mathrm{dist}}$ may still be admitted to $\mathcal{C}_{\mathrm{Acc}}$ if it satisfies the heading-alignment criterion and $\operatorname{Safe}(\tau^{(j)})$ holds. 
The resulting four configurations (namely, \texttt{W+H}, \texttt{W+NH}, \texttt{E+H}, and \texttt{E+NH}) are each evaluated over ten trials per map, compared with five trials per map for \textit{Tests~\#2--\,\#5}. In \textit{Test~\#2}, we disable the ESS-guided switching mechanism in~\eqref{eq:proposal_update_mode} by fixing $\mathsf{mode}=\mathrm{moment}$; thus, the nominal control mean is updated exclusively through the moment-matching regime in~\eqref{eq:moment_mean_update}, without invoking the winner-step regime in~\eqref{eq:winner_mean_update}. In \textit{Test~\#3}, we replace the default GPU-batched MPPI optimization described in Section~\ref{sec:GPU-Batched-MPPI-Optimization} with sequential per-bait optimization, while disabling the adaptive bait-handling procedure in~\eqref{eq:bait_validity}--\eqref{eq:bait_retraction}; as a result, geometric bait screening and retraction are omitted, whereas candidate trajectories remain subject to the collision-safety condition in~\eqref{eq:candidate_collision_safety}. In \textit{Test~\#4}, we investigate the optional cruise-speed cost in~\eqref{eq:cruise_cost} by setting $w_c=1$, instead of the default $w_c=0$, thereby activating this term in both MPPI rollout  optimization and deterministic candidate reevaluation in~\eqref{eq:bait_cost}.
Finally, \textit{Test~\#5} examines the effect of restricting the branching limit by reducing $k_b$ from its default value of $3$ to $1$ in~\eqref{eq:effective_topk}--\eqref{eq:topk_set}; accordingly, at most one candidate, corresponding to the lowest cost among the accepted candidates, is retained from each tree expansion.

    \begin{table}[!t]
        \caption{Influence of algorithmic components and parameter settings on SI-MPPI planning performance relative to the default results in Table~\ref{tab:offline_benchmark_means}, with mean~$\pm$~std computed from the four per-map means.}
        \label{tab:ablation_overall_summary}
        \centering
        \footnotesize
        \addtolength{\tabcolsep}{-4.5pt}
        \setlength{\extrarowheight}{1pt}
        \begin{tabular}{|c||c|c|c|c|}
        \hline
        Setup & $P_{\mathrm{succ}}$~[\%]\;$\uparrow$  & $T_{\mathrm{plan}}$~[\si{\second}]\;$\downarrow$ & $\ell_{\Gamma}$~[\si{\metre}]\;$\downarrow$ & $N_{\mathrm{tree}}$\;$\downarrow$ \\
        \hline\hline
        Default
          & 100
          & $8.38 \pm 7.76$
          & $138.90 \pm 87.75$
          & $521.35 \pm 515.42$ \\
        \hline\hline
        \multicolumn{5}{|c|}{\textit{\textbf{Test~\#1:} Backpropagation modes and heading-alignment settings}} \\
        \hline
        \texttt{W+H}
          & \cellcolor{gray!30}100
          & $10.74 \pm 7.48$
          & $157.94 \pm 88.44$
          & $659.53 \pm 540.78$ \\
        \texttt{W+NH}
          & 97.5
          & $9.66 \pm 10.35$
          & $151.78 \pm 87.83$
          & $595.20 \pm 666.47$ \\
        \texttt{E+H}
          & \cellcolor{gray!30}100
          & $9.00 \pm 9.43$
          & $146.58 \pm 86.68$
          & $550.15 \pm 602.27$ \\
        \texttt{E+NH}
          & \cellcolor{gray!30}100
          & \cellcolor{gray!30}$8.50 \pm 9.03$
          & \cellcolor{gray!30}$143.18 \pm 88.35$
          & \cellcolor{gray!30}$521.48 \pm 593.80$ \\
        \hline\hline
        \multicolumn{5}{|c|}{\textit{\textbf{Tests \#2--\#5:} \texttt{moment}, Sequential Optimization, $w_c=1$, and $k_b=1$}} \\
        \hline
        \textit{Test~\#2}
          & 100
          & $8.00 \pm 6.82$
          & $159.40 \pm 83.34$
          & $485.50 \pm 468.52$ \\
          \textit{Test~\#3}
          & 100
          & $18.53 \pm 12.38$
          & $142.65 \pm 87.66$
          & $440.95 \pm 287.51$ \\
          \textit{Test~\#4}
          & 100
          & $8.99 \pm 5.76$
          & $164.69 \pm 82.71$
          & $566.15 \pm 407.00$ \\
          \textit{Test~\#5}
          & 80
          & $1.57 \pm 0.80$
          & $137.86 \pm 86.67$
          & $28.32 \pm 16.67$ \\
        \hline
        \end{tabular}
    \end{table}
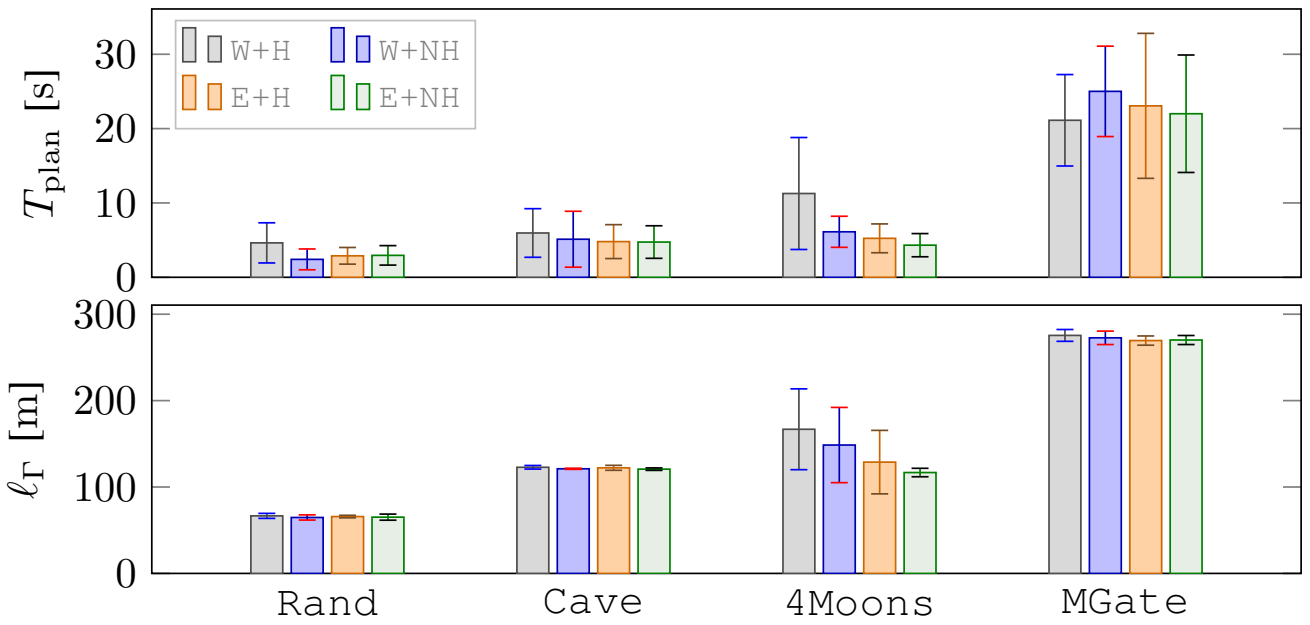
\begin{figure}[!t]
    \centering
    \input{figures/ablation_backprop_blend_bars.tikz}
    \vspace*{-5pt}
\caption{Effect of backpropagation modes and heading alignment on planning time $T_{\mathrm{plan}}$ (top) and path length $\ell_{\Gamma}$ (bottom) across the four benchmark maps, based on ten trials per configuration per map.}
    \label{fig:ablation_backprop_blend_bars}
\end{figure} 

\begin{figure}[!t]
    \centering
    \begin{subfigure}[t]{\columnwidth}
        \centering
        \input{figures/ablation_expected_ell.tikz}
        \caption{Path length $\ell_{\Gamma}$ for \textit{Tests~\#2 and~\#4}}
        \label{fig:ablation_expected_ell}
    \end{subfigure}
    \begin{subfigure}[t]{\columnwidth}
        \centering
        \input{figures/ablation_expected_tplan.tikz}
        \caption{Planning time $T_{\mathrm{plan}}$ for \textit{Tests~\#3 and~\#5}}
        \label{fig:ablation_expected_tplan}
    \end{subfigure}
    \begin{subfigure}[t]{\columnwidth}
        \centering
        \input{figures/ablation_expected_angacc.tikz}
        \caption{RMS angular acceleration $a_{\omega,\mathrm{rms}}$ for \textit{Test~\#3}}
        \label{fig:ablation_expected_angacc}
    \end{subfigure}
    \vspace*{-6pt}
    \caption{Performance comparison of the default SI-MPPI configuration, as reported in Table~\ref{tab:offline_benchmark_means}, and the ablation variants in \textit{Tests~\#2--\#5} across the four benchmark maps.}
    \label{fig:ablation_expected_bars}
\end{figure}
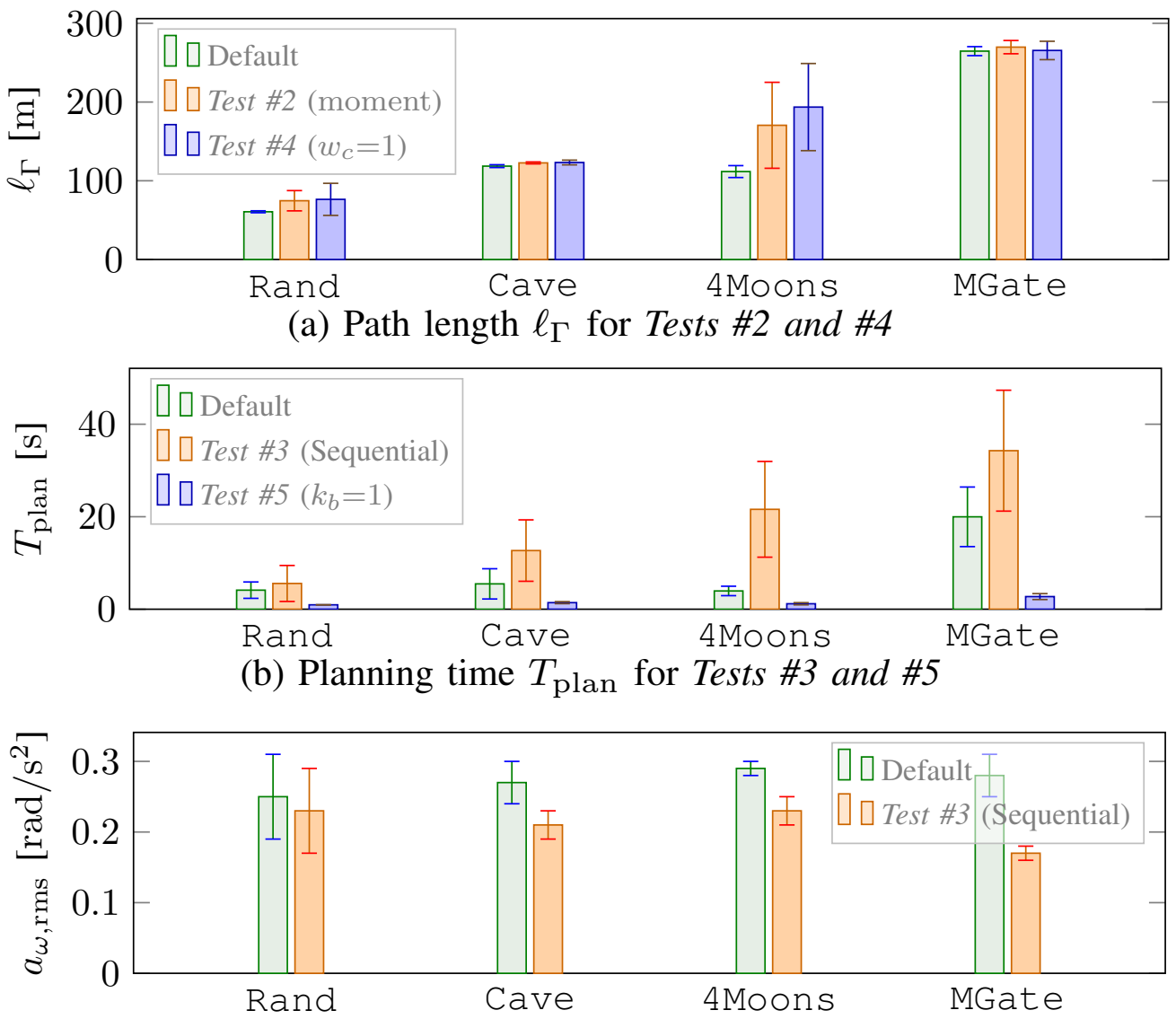

Table~\ref{tab:ablation_overall_summary} summarizes the overall results of the five ablation tests, while Figs.~\ref{fig:ablation_backprop_blend_bars} and~\ref{fig:ablation_expected_bars} highlight the principal performance differences across the four benchmark maps. As observed in \textit{Test~\#1}, the expected-cost (\texttt{E}) backpropagation mode generally outperforms the winner-cost (\texttt{W}) mode in terms of $T_{\mathrm{plan}}$ and $\ell_{\Gamma}$, while maintaining $P_{\mathrm{succ}}=100\%$ under both heading-alignment settings. In particular, \texttt{E+NH} achieves the lowest average $T_{\mathrm{plan}}$ ($8.50\,\si{\second}$) and shortest average $\ell_{\Gamma}$ ($143.18\,\si{\metre}$) among the four configurations, supporting its choice as the default SI-MPPI configuration in Table~\ref{tab:offline_benchmark_means}. This advantage is particularly evident on \mapname{4Moons}, a map characterized by narrow passages, where \texttt{E+NH} reduces $T_{\mathrm{plan}}$ from $11.27$ to $4.32\,\si{\second}$ and $\ell_{\Gamma}$ from $166.84$ to $116.77\,\si{\metre}$ compared with \texttt{W+H}, as shown in Fig.~\ref{fig:ablation_backprop_blend_bars}. On \mapname{MGate}, \texttt{W+NH} is the only configuration with an unsuccessful trial ($P_{\mathrm{succ}}=90\%$ on this map), whereas the remaining configurations achieve full success with relatively small differences in $\ell_{\Gamma}$. Taken together, these findings suggest that backpropagating the expected cost $\widehat{J}$ in~\eqref{eq:backpropagation_cost}, rather than the minimum retained-candidate cost $J_{\mathrm{win}}$, provides more effective guidance for UCB-based tree selection by accounting for the relative costs of all retained candidates. 
Turning to \textit{Test~\#2}, restricting adaptation of the MPPI nominal control mean to the moment-matching regime maintains $P_{\mathrm{succ}}=100\%$ and yields a planning time $T_{\mathrm{plan}}$ comparable to the default configuration ($8.00$ versus $8.38\,\si{\second}$); however, the average path length $\ell_{\Gamma}$ increases from $138.90$ to $159.40\,\si{\metre}$. The largest increase occurs on \mapname{4Moons}, where $\ell_{\Gamma}$ increases from $111.69$ to $170.48\,\si{\metre}$, as depicted in Fig.~\ref{fig:ablation_expected_ell}, while the remaining maps exhibit smaller differences. These results indicate that the ESS-guided switching mechanism in~\eqref{eq:proposal_update_mode} improves path quality without substantially increasing $T_{\mathrm{plan}}$ by activating the winner-step regime when a low-cost candidate dominates the weight distribution, thereby preserving its control pattern and helping the planner commit to shorter routes through narrow passages.

\begin{figure}[!t]
    \centering
    \begin{subfigure}[t]{0.48\columnwidth}
        \centering
        \includegraphics[width=\linewidth]{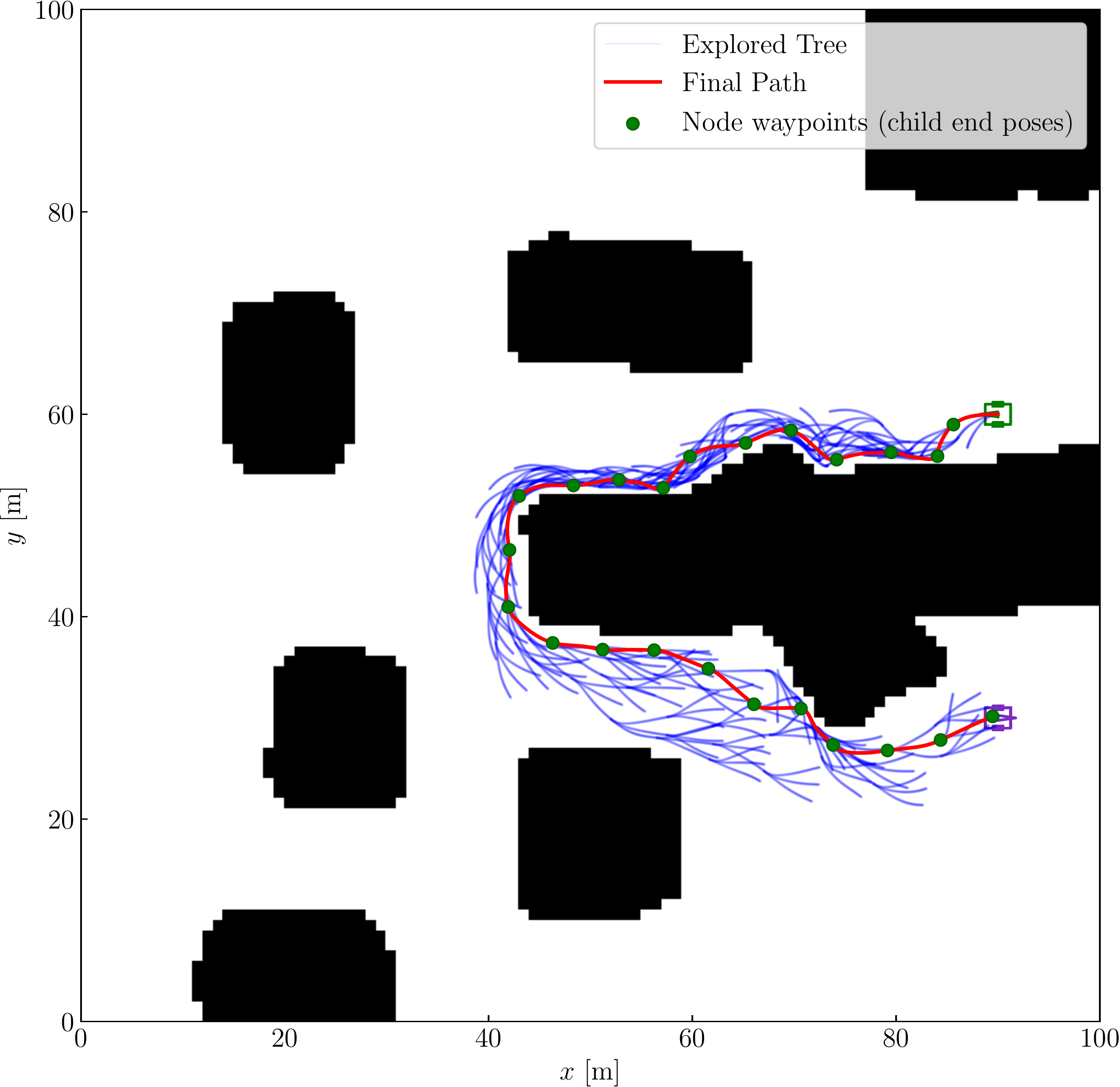}
        \vspace*{-6pt}
        \caption{Default: $\ell_{\Gamma}=\SI{117.59}{\metre}$}
        \label{fig:ablation_sequential_cave_default_tree}
    \end{subfigure}%
    \begin{subfigure}[t]{0.48\columnwidth}
        \centering
        \includegraphics[width=\linewidth]{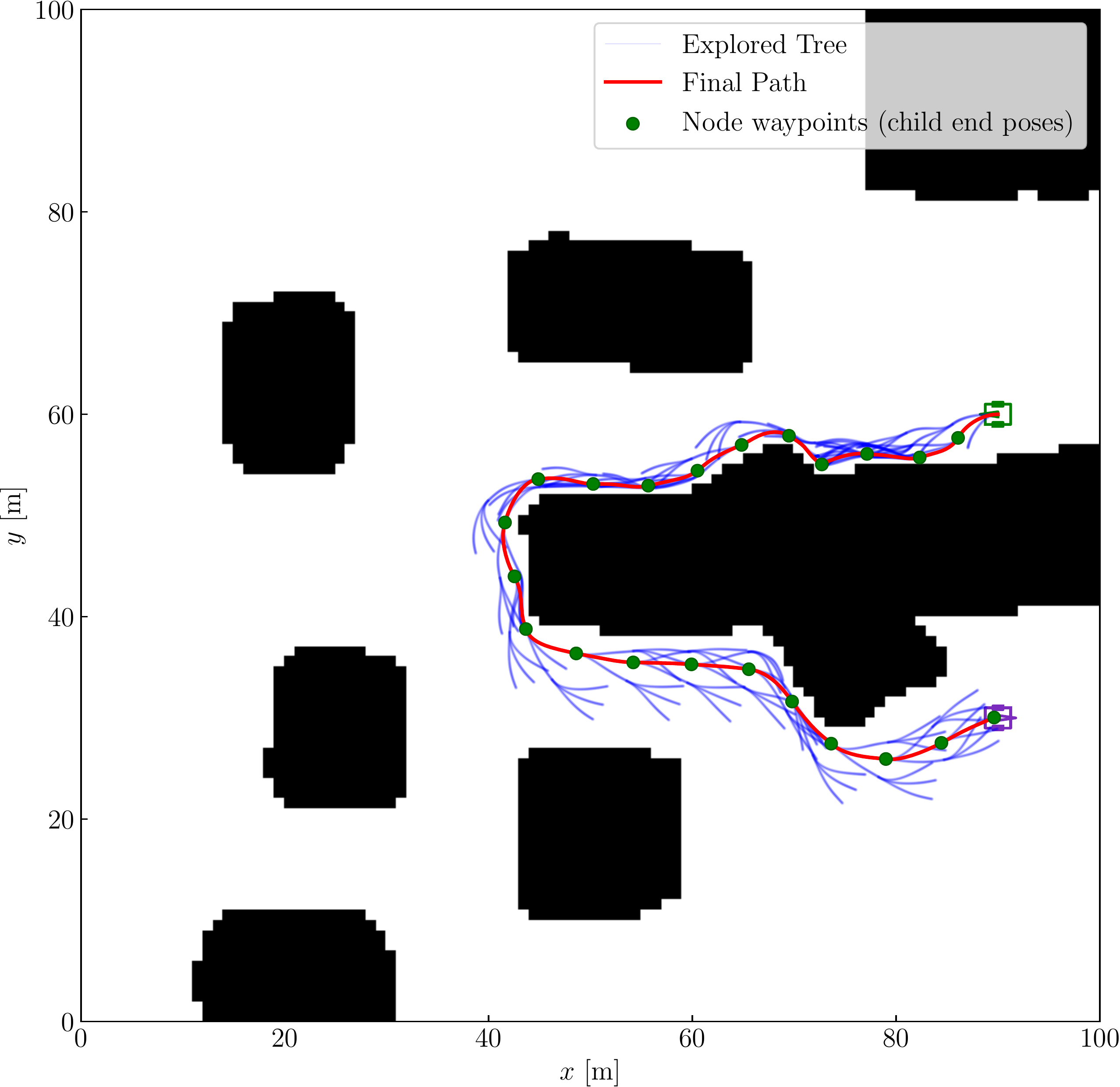}
        \vspace*{-6pt}
        \caption{Sequential: $\ell_{\Gamma}=\SI{117.17}{\metre}$}
        \label{fig:ablation_sequential_cave_seq_tree}
    \end{subfigure}
    \begin{subfigure}[t]{0.50\columnwidth}
        \centering
        \input{figures/ablation_sequential_cave/cave_default_r1_omega.tikz}
        \caption{Default: $a_{\omega,\mathrm{rms}}=\SI{0.30}{\radian/\second^2}$}
        \label{fig:ablation_sequential_cave_default_omega}
    \end{subfigure}%
    \begin{subfigure}[t]{0.50\columnwidth}
        \centering
        \hspace*{-3pt}\input{figures/ablation_sequential_cave/cave_sequential_r4_omega.tikz}
        \caption{Sequential: $a_{\omega,\mathrm{rms}}=\SI{0.18}{\radian/\second^2}$}
        \label{fig:ablation_sequential_cave_seq_omega}
    \end{subfigure}
    \vspace*{-6pt}
    \caption{Comparison of the default GPU-batched and sequential MPPI configurations on \mapname{Cave}, using trials with nearly equal path lengths: (a)--(b) search trees and returned plans; (c)--(d) corresponding angular-velocity $\omega$ profiles along the returned trajectories.}
    \label{fig:ablation_sequential_cave}
\end{figure}

In \textit{Test~\#3}, GPU-batched MPPI expansion and adaptive bait handling are disabled, so baits are optimized sequentially. 
As anticipated, the primary effect is a substantial reduction in planning efficiency, as $T_{\mathrm{plan}}$ increases from $8.38$ to $18.53\,\si{\second}$, more than doubling overall, with the largest absolute increases on \mapname{4Moons} and \mapname{MGate}, as depicted in Fig.~\ref{fig:ablation_expected_tplan}. Despite this computational overhead, $\ell_{\Gamma}$ remains comparable to the default configuration ($142.65$ versus $138.90\,\si{\metre}$), while the average RMS angular acceleration $a_{\omega,\mathrm{rms}}$ decreases from $0.27$ to $0.21\,\si{\radian\per\second\squared}$, with reductions observed on all four maps (see Fig.~\ref{fig:ablation_expected_angacc}). This improvement in control smoothness is further illustrated in Fig.~\ref{fig:ablation_sequential_cave}, where two nearly equal-length paths ($117.59$ versus $117.17\,\si{\metre}$) on \mapname{Cave} exhibit noticeably different angular-velocity profiles, with the sequential configuration producing smoother control variations ($a_{\omega,\mathrm{rms}}=0.18$ versus $0.30\,\si{\radian\per\second\squared}$ for the default configuration). By contrast, the default GPU-batched configuration applies the bait-directed yaw-rate initialization in~\eqref{eq:bait_yaw_initialization}, introducing an additional steering bias that can increase angular-velocity variations.
For \textit{Test~\#4}, the dominant effect of enabling the cruise-speed cost $q_{\mathrm{cruise}}$ in~\eqref{eq:cruise_cost} with $w_c=1$ is a deterioration in path quality, as $\ell_{\Gamma}$ increases from $138.90$ to $164.69\,\si{\metre}$, largely driven by the increase on \mapname{4Moons} (Fig.~\ref{fig:ablation_expected_ell}), while $T_{\mathrm{plan}}$ remains comparable to the default configuration ($8.99$ versus $8.38\,\si{\second}$). Notably, the empirical results show that the average linear speed $v$ along the successful trajectories across the four benchmark maps increases slightly from $0.429$ to $0.457\,\si{\metre\per\second}$ (approximately $6.5\%$), remaining well below the maximum allowable speed $v_{\max}=\SI{2}{\metre\per\second}$, which supports the choice of $w_c=0$ in the default SI-MPPI configuration.

\begin{figure*}[!ht]
    \centering
    \begin{subfigure}[t]{0.245\textwidth}
        \centering
        \includegraphics[width=\linewidth,height=0.975\linewidth]{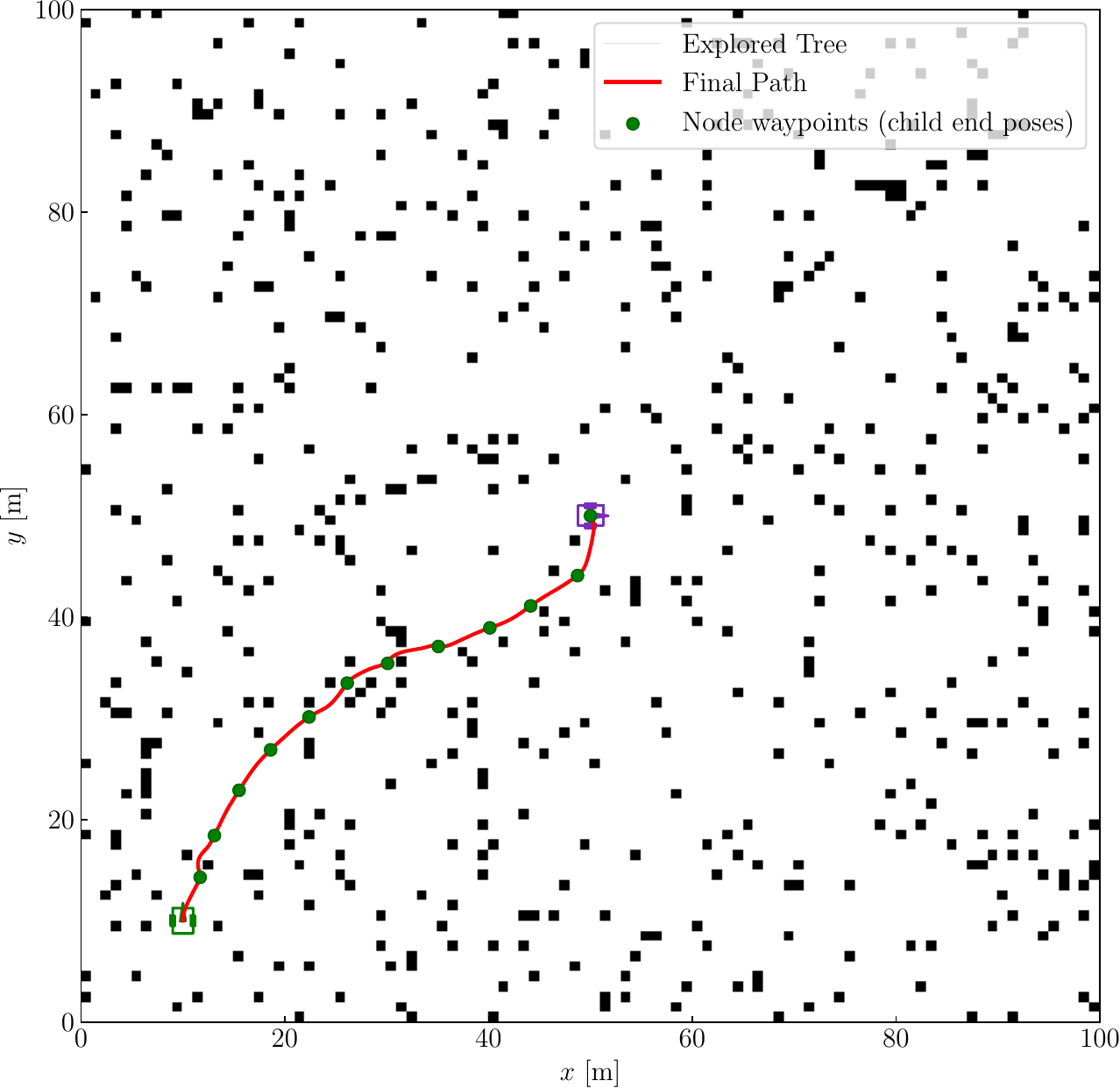}
        \vspace*{-6pt}
        \caption{\mapname{Rand}, $\ell_{\Gamma}=\SI{61.38}{\metre}$} 
        \label{fig:ablation_topk1_shortest_random}
    \end{subfigure}%
    \begin{subfigure}[t]{0.245\textwidth}
        \centering
        \includegraphics[width=\linewidth,height=0.975\linewidth]{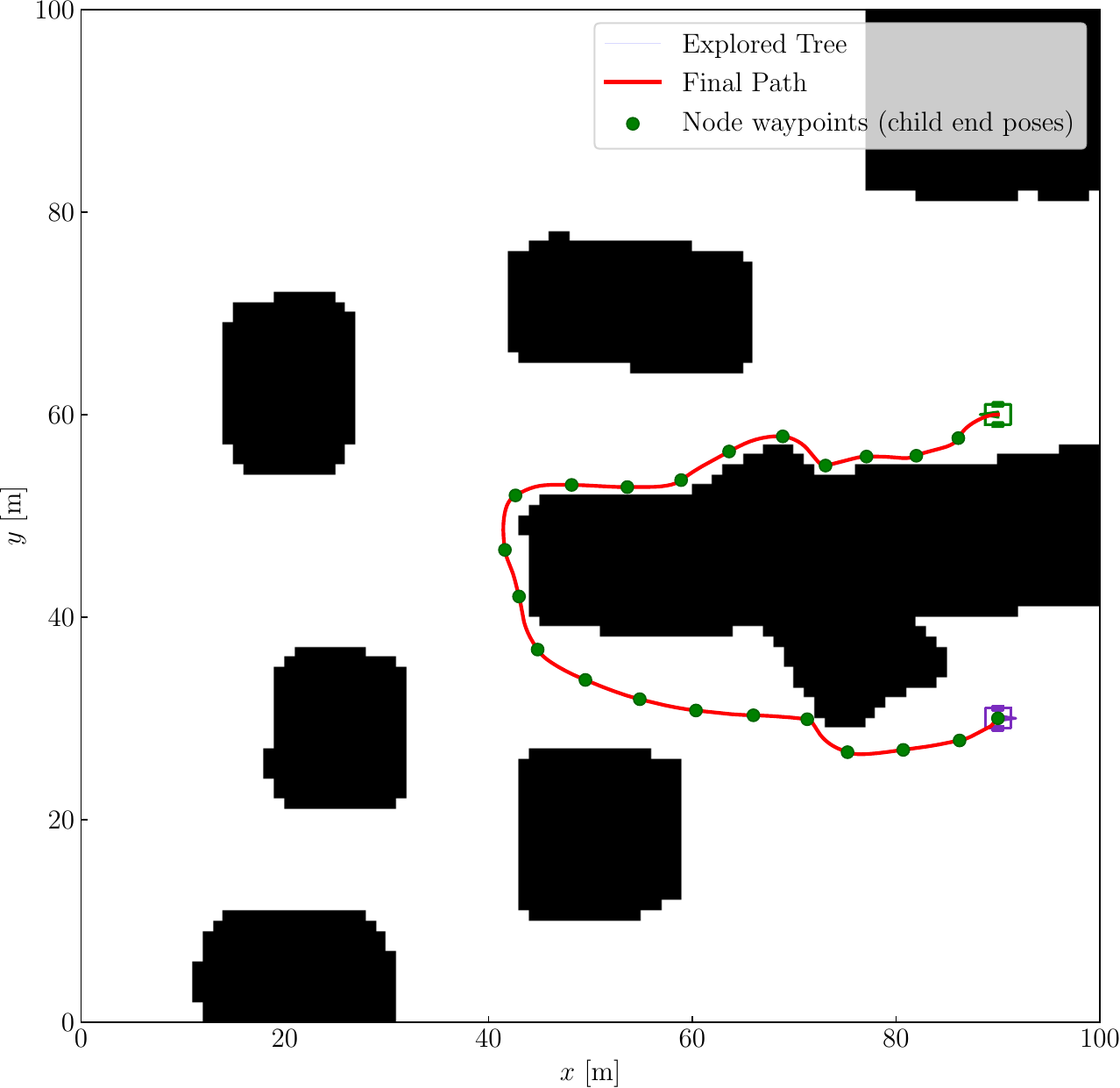}
        \vspace*{-6pt}
        \caption{\mapname{Cave}, $\ell_{\Gamma}=\SI{116.21}{\metre}$} 
        \label{fig:ablation_topk1_shortest_cave}
    \end{subfigure}
    \begin{subfigure}[t]{0.245\textwidth}
        \centering
        \includegraphics[width=\linewidth,height=0.975\linewidth]{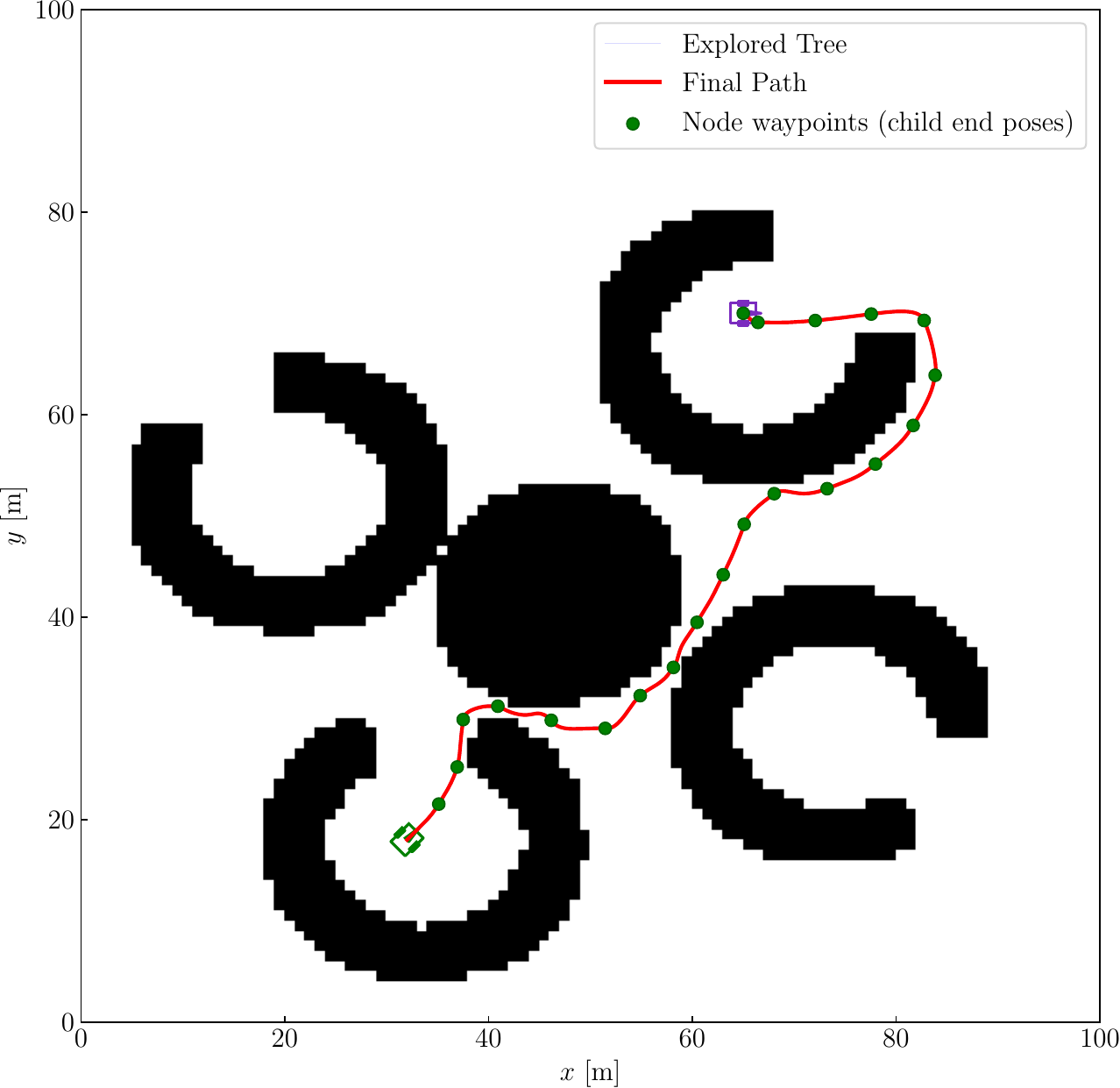}
        \vspace*{-6pt}
        \caption{\mapname{4Moons}, $\ell_{\Gamma}=\SI{103.52}{\metre}$} 
        \label{fig:ablation_topk1_shortest_four_moons}
    \end{subfigure}
    \begin{subfigure}[t]{0.245\textwidth}
        \centering
        \includegraphics[width=\linewidth,height=0.975\linewidth]{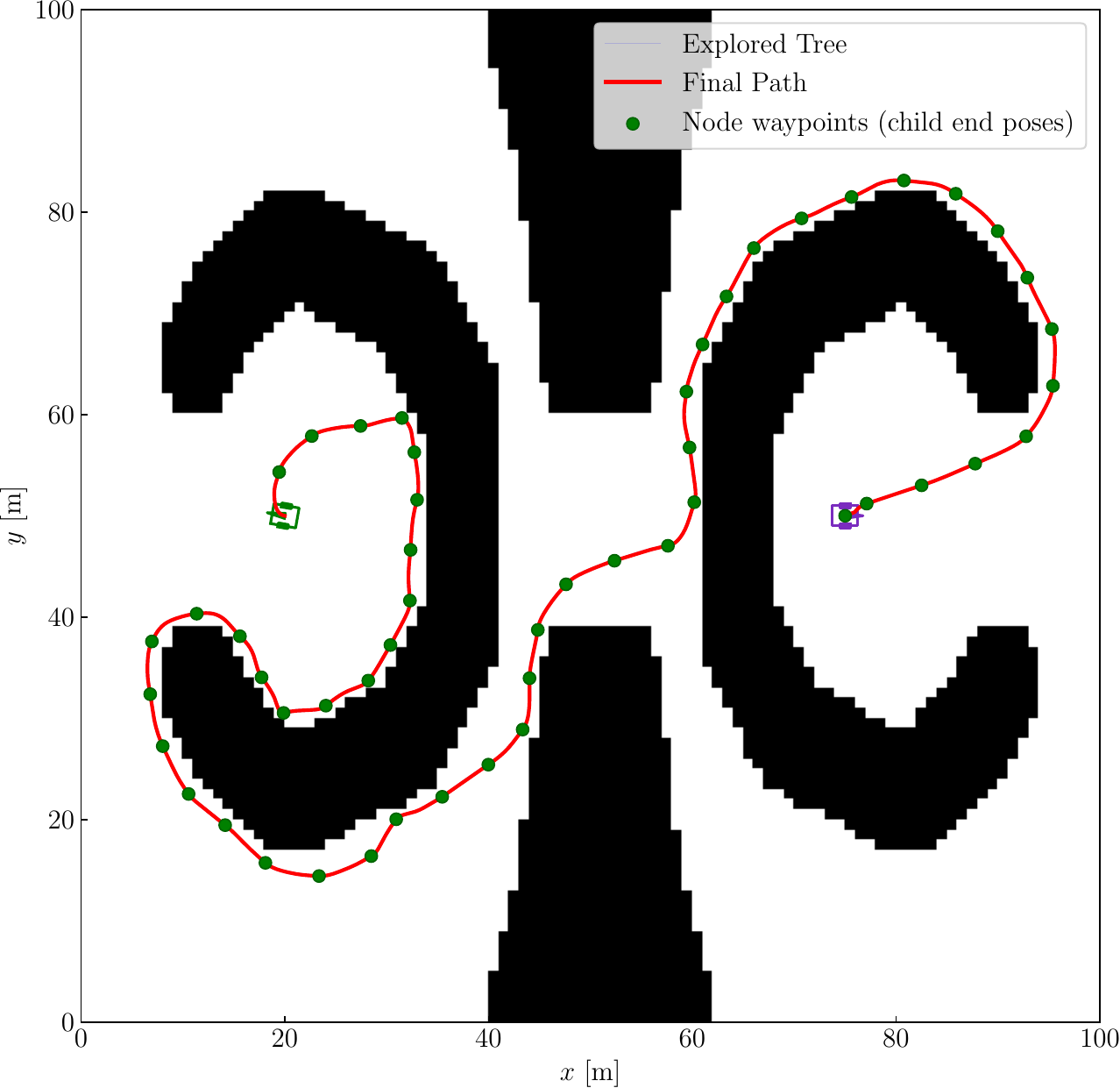}
        \vspace*{-6pt}
        \caption{\mapname{MGate}, $\ell_{\Gamma}=\SI{259.60}{\metre}$} 
        \label{fig:ablation_topk1_shortest_moon_gate}
    \end{subfigure}
    \vspace*{-6pt}
    \caption{Shortest successful trajectories obtained with $k_b=1$ (\textit{Test~\#5}) over five trials on each benchmark map; under single-branch retention, each search tree forms a single chain that coincides with the returned path.}
    \label{fig:ablation_topk1_shortest}
\end{figure*}

\needspace{2\baselineskip} 
Finally, \textit{Test~\#5} reveals the strongest trade-off between computational efficiency and planning robustness. In particular, reducing the branching limit to $k_b=1$ substantially reduces search effort, as the average $T_{\mathrm{plan}}$ decreases from $8.38$ to $1.57\,\si{\second}$ (see Fig.~\ref{fig:ablation_expected_tplan} for the per-map reductions) and tree size $N_{\mathrm{tree}}$ from $521.35$ to $28.32$ nodes. These computational savings, however, come at the expense of planning robustness, as the overall success rate $P_{\mathrm{succ}}$ decreases from $100\%$ to $80\%$, marking the lowest success rate among all ablation tests. At the individual-map level, all trials succeed on \mapname{Rand} and \mapname{Cave}, whereas $P_{\mathrm{succ}}$ falls to $40\%$ on \mapname{4Moons} and $80\%$ on \mapname{MGate}. 
This difference in success rates reflects the distinction between local motion exploration and persistent tree branching. With $k_b=1$, each expansion admits at most one child, reducing $\mathcal{T}$ to a single chain and effectively turning SI-MPPI into a greedy search without backtracking. Nevertheless, each expansion still optimizes control sequences toward the $B_{\mathrm{ret}}$ retained baits spanning different forward directions and selects the lowest-cost candidate satisfying the collision-safety and progress criteria. Guided by collision-avoidance and goal-progress costs, this local exploration can produce successful plans even without retaining alternative branches, as observed on \mapname{Rand} and \mapname{Cave}. However, when the chain reaches a dead end, no sibling branches remain for further exploration, limiting robustness in environments with concave obstacles and narrow passages, as reflected in the lower success rates on \mapname{4Moons} and \mapname{MGate}. To illustrate the resulting search structure, Fig.~\ref{fig:ablation_topk1_shortest} presents the shortest successful path obtained with $k_b=1$ among five trials on each map, together with the corresponding single-chain search trees, which coincide with their returned paths and contain only $13$, $23$, $22$, and $52$ nodes, respectively.

\subsection{Discussion}
\label{subsec:discussion}

The comparative results demonstrate the effectiveness of SI-MPPI for kinodynamic motion planning in large-scale, cluttered environments, achieving the smoothest angular control profiles (as measured by $\sigma_{\omega}$ and $a_{\omega,\mathrm{rms}}$) and the most compact search trees ($N_{\mathrm{tree}}$) among the evaluated planners, while maintaining competitive planning times and path lengths; these outcomes reflect the complementary roles of UCB-guided tree selection and bait-guided MPPI expansions. The ablation results further show that GPU-batched MPPI optimization combined with the adaptive bait-handling rule substantially reduces planning time $T_{\mathrm{plan}}$, whereas retaining multiple candidates per expansion (i.e., $k_b>1$) is important for planning robustness in geometrically constrained environments. 
Beyond GPU acceleration, another central contributor to this computational efficiency is the role assigned to MPPI during tree expansion. As discussed in Section~\ref{subsubsec:control_update}, we employ MPPI as an open-loop local trajectory optimizer rather than a receding-horizon controller: a finite-horizon control sequence is optimized toward each retained bait and subsequently applied in forward simulation of the robot dynamics to generate a candidate motion segment without repeated feedback-based steering.
This differs from LQR-RRT$^{*}$, in which each local connection is obtained by simulating LQR feedback control, with additional steering computations during parent selection and rewiring. Together with selective tree growth and termination upon finding a feasible plan, this open-loop expansion formulation provides insight into why SI-MPPI achieves competitive planning times, even when local MPPI optimizations are performed sequentially. 

On the other hand, such an open-loop expansion strategy yields planned trajectories with relatively low linear velocities compared with the robot's maximum allowable linear velocity $v_{\max}$, even when the cruise-speed cost is enabled, as observed in \textit{Test~\#4}. This motivates a two-stage use of MPPI: SI-MPPI first identifies a feasible global route through open-loop tree expansion, after which a receding-horizon MPPI tracker follows the extracted waypoints using the same underlying state-cost formulation in~\eqref{eq:task_state_cost}, achieving average linear velocities closer to $v_{\max}$, as demonstrated in the supplementary video. This approach confines the additional closed-loop optimization to the selected route rather than the entire search tree, allowing efficient global planning and higher-speed execution to be addressed separately.

\section{Conclusion and Future Work}
\label{sec:conclusion}

In this paper, we proposed SI-MPPI, a kinodynamic planning framework that integrates MCTS-based tree search with GPU-batched MPPI trajectory optimization, enabling efficient long-horizon exploration through dynamically feasible local motion generation. By fusing the expansion and default-policy simulation stages of MCTS into a single bait-guided MPPI expansion stage, SI-MPPI constructs a sparse and asymmetric search tree whose edges represent dynamically feasible motion segments without requiring exact BVP steering, local linearization, or fixed motion primitives. 
Adaptive bait handling, feasibility gating, and top-$k$ selection further confine tree growth to feasible, low-cost branches, while ESS-guided control-mean adaptation and cost-based backpropagation exploit candidate-cost information to warm-start subsequent MPPI optimizations and guide UCB-based node selection, respectively. Comparative planning simulations across the benchmark maps demonstrated that SI-MPPI achieved an effective balance among search complexity, trajectory quality, and planning efficiency. Compared with kinodynamic MCTS, SI-MPPI reduced average tree size by approximately $15\times$ and average path length by $22\%$, highlighting the benefits of the proposed bait-guided MPPI expansion mechanism. Relative to LQR-RRT$^{*}$ and IK-RRT$^{*}$, SI-MPPI reduced average planning time by approximately $26.8\times$ and $9.7\times$, respectively, while achieving an average path length $14.7\%$ shorter than that of LQR-RRT$^{*}$ and only $2.7\%$ longer than that of IK-RRT$^{*}$.
Our future plan involves extending SI-MPPI to higher-dimensional robotic systems and developing speed-aware multi-objective planning strategies to improve traversal speeds within the existing framework.

\ifanonymous
\else
\section*{Acknowledgment}
The authors thank Weizhe Chen for insightful discussions and generous support throughout this work.
\fi
\bibliographystyle{IEEEtran}
\bibliography{references}

\end{document}

%% file: figures/ess_update_regimes_a.tikz
\begin{tikzpicture}[scale=1,
  font=\scriptsize,
  >=Latex,
  axis/.style={draw=black!75, line width=0.55pt, ->},
  bar/.style={draw=blue!55!black, fill=blue!30, line width=0.4pt},
  ticklab/.style={font=\scriptsize},
  vallab/.style={font=\scriptsize, anchor=south, inner sep=0.8pt},
]
  \def\ys{1.3}
  \def\bw{0.17}
  \def\x0{0.55}
  \def\bsep{0.8}
  \def\xmax{3.35}
  \def\vlabgap{0.08}  

  \pgfmathsetmacro{\xA}{\x0}
  \pgfmathsetmacro{\xB}{\x0+\bsep}
  \pgfmathsetmacro{\xC}{\x0+2*\bsep}
  \pgfmathsetmacro{\xD}{\x0+3*\bsep}

  \foreach \x/\h in {
    \xA/0.28,
    \xB/0.24,
    \xC/0.26,
    \xD/0.22} {
    \pgfmathsetmacro{\yy}{\h*\ys}
    \filldraw[bar] (\x-\bw,0) rectangle (\x+\bw,\yy);
    \node[vallab] at (\x, {\yy+\vlabgap}) {\h};
  }

  \draw[axis] (0,0) -- (0,\ys+0.28);
  \node[anchor=south, font=\scriptsize] at (0,\ys+0.30) {$w^{(j)}$};
  \draw[axis] (0,0) -- (\xmax+0.22,0);
  \node[anchor=west, font=\scriptsize] at (\xmax+0.24,0) {$j$};

  \foreach \h in {0.5,1} {
    \pgfmathsetmacro{\yy}{\h*\ys}
    \draw[black!75, line width=0.55pt] (0,\yy) -- ++(0.06,0);
    \node[ticklab, anchor=east, inner sep=1.2pt] at (-0.05,\yy) {\h};
  }

  \foreach \x/\lab in {\xA/1, \xB/2, \xC/3, \xD/4} {
    \draw[black!75, line width=0.55pt] (\x,0) -- ++(0,0.06);
    \node[ticklab, anchor=north, inner sep=1.2pt] at (\x,-0.05) {\lab};
  }
\end{tikzpicture}

%% file: figures/ess_update_regimes_b.tikz
\begin{tikzpicture}[scale=1,
  font=\scriptsize,
  >=Latex,
  axis/.style={draw=black!75, line width=0.55pt, ->},
  bar win/.style={draw=orange!70!black, fill=orange!55, line width=0.4pt},
  bar weak/.style={draw=black!45, fill=black!15, line width=0.4pt},
  ticklab/.style={font=\scriptsize},
  vallab/.style={font=\scriptsize, anchor=south, inner sep=0.8pt},
]
  \def\ys{1.3}
  \def\ybr{0.42}
  \def\wbr{0.10}
  \def\bw{0.17}
  \def\x0{0.55}
  \def\bsep{0.8}
  \def\xmax{3.35}
  \def\vlabgap{0.08}  

  \pgfmathsetmacro{\xA}{\x0}
  \pgfmathsetmacro{\xB}{\x0+\bsep}
  \pgfmathsetmacro{\xC}{\x0+2*\bsep}
  \pgfmathsetmacro{\xD}{\x0+3*\bsep}

  \pgfmathsetmacro{\yy}{\ybr + ((0.82-\wbr)/(1-\wbr))*(\ys-\ybr)}
  \filldraw[bar win] (\xA-\bw,0) rectangle (\xA+\bw,\yy);
  \node[vallab] at (\xA, {\yy+\vlabgap}) {0.82};

  \foreach \x/\h in {
    \xB/0.08,
    \xC/0.05,
    \xD/0.05} {
    \pgfmathsetmacro{\yy}{(\h/\wbr)*\ybr}
    \filldraw[bar weak] (\x-\bw,0) rectangle (\x+\bw,\yy);
    \node[vallab] at (\x, {\yy+\vlabgap}) {\h};
  }

  \draw[axis] (0,0) -- (0,\ys+0.28);
  \node[anchor=south, font=\scriptsize] at (0,\ys+0.30) {$w^{(j)}$};
  \draw[axis] (0,0) -- (\xmax+0.22,0);
  \node[anchor=west, font=\scriptsize] at (\xmax+0.24,0) {$j$};

  \draw[black!75, line width=0.55pt]
    (-0.06,\ybr-0.04) -- (0.06,\ybr+0.02)
    (-0.06,\ybr-0.01) -- (0.06,\ybr+0.05);

  \foreach \h in {0.5,1} {
    \pgfmathsetmacro{\yy}{\ybr + ((\h-\wbr)/(1-\wbr))*(\ys-\ybr)}
    \draw[black!75, line width=0.55pt] (0,\yy) -- ++(0.06,0);
    \node[ticklab, anchor=east, inner sep=1.2pt] at (-0.05,\yy) {\h};
  }
  \draw[black!75, line width=0.55pt] (0,\ybr) -- ++(0.06,0);
  \node[ticklab, anchor=east, inner sep=1.2pt] at (-0.05,\ybr) {\wbr};

  \foreach \x/\lab in {\xA/1, \xB/2, \xC/3, \xD/4} {
    \draw[black!75, line width=0.55pt] (\x,0) -- ++(0,0.06);
    \node[ticklab, anchor=north, inner sep=1.2pt] at (\x,-0.05) {\lab};
  }
\end{tikzpicture}

%% file: figures/ablation_backprop_blend_bars.tikz
\begin{tikzpicture}[scale=0.80]
\begin{groupplot}[
    group style={
        group size=1 by 2,
        vertical sep=0.23cm,
        x descriptions at=edge bottom,
        y descriptions at=edge left,
    },
    width=1.22\columnwidth,
    height=3.8cm,
    ybar,
    enlarge x limits=0.22,
    symbolic x coords={Rand,Cave,4Moons,MGate},
    xtick=data,
    xtick style={draw=none}, 
    xticklabel style={font=\ttfamily}, 
    tick align=inside,
    tick style={thin},
    ymin=0,
    every axis plot/.append style={
        /pgf/bar width=7.5pt,
        error bars/y dir=both,
        error bars/y explicit,
    },
]
\nextgroupplot[
    ylabel={$T_{\mathrm{plan}}$~[\si{\second}]},
    legend style={
        at={(0.02,0.96)},
        anchor=north west,
        legend columns=2,
        font=\footnotesize\color{gray},
        /tikz/every even column/.append style={column sep=0.25cm},
        draw=black!25,
        fill=white,
        fill opacity=0.92,
        text opacity=1,
        inner sep=1.5pt,
    },
    legend cell align=left,
]
\addplot+[draw=black!70, fill=black!15] coordinates {
    (Rand,4.63) +- (0,2.70)
    (Cave,5.96) +- (0,3.27)
    (4Moons,11.27) +- (0,7.53)
    (MGate,21.12) +- (0,6.15)
};
\addlegendentry{\texttt{W+H}}
\addplot+[draw=blue!70!black, fill=blue!25] coordinates {
    (Rand,2.41) +- (0,1.40)
    (Cave,5.12) +- (0,3.76)
    (4Moons,6.12) +- (0,2.09)
    (MGate,25.01) +- (0,6.08)
};
\addlegendentry{\texttt{W+NH}}
\addplot+[draw=orange!80!black, fill=orange!35] coordinates {
    (Rand,2.89) +- (0,1.12)
    (Cave,4.80) +- (0,2.28)
    (4Moons,5.24) +- (0,1.94)
    (MGate,23.06) +- (0,9.75)
};
\addlegendentry{\texttt{E+H}}
\addplot+[draw=green!50!black, fill=green!35!black!10] coordinates {
    (Rand,2.95) +- (0,1.31)
    (Cave,4.74) +- (0,2.19)
    (4Moons,4.32) +- (0,1.56)
    (MGate,22.00) +- (0,7.90)
};
\addlegendentry{\texttt{E+NH}}
\nextgroupplot[
    ylabel={$\ell_{\Gamma}$~[\si{\metre}]},
]
\addplot+[draw=black!70, fill=black!15] coordinates {
    (Rand,66.60) +- (0,2.89)
    (Cave,122.85) +- (0,2.08)
    (4Moons,166.84) +- (0,46.77)
    (MGate,275.47) +- (0,6.84)
};
\addplot+[draw=blue!70!black, fill=blue!25] coordinates {
    (Rand,64.75) +- (0,3.00)
    (Cave,121.10) +- (0,0.64)
    (4Moons,148.59) +- (0,43.54)
    (MGate,272.68) +- (0,7.78)
};
\addplot+[draw=orange!80!black, fill=orange!35] coordinates {
    (Rand,65.81) +- (0,1.47)
    (Cave,122.22) +- (0,2.87)
    (4Moons,128.82) +- (0,36.73)
    (MGate,269.50) +- (0,5.34)
};
\addplot+[draw=green!50!black, fill=green!35!black!10] coordinates {
    (Rand,65.13) +- (0,3.54)
    (Cave,120.69) +- (0,1.43)
    (4Moons,116.77) +- (0,4.90)
    (MGate,270.14) +- (0,5.24)
};
\end{groupplot}
\end{tikzpicture}

%% file: figures/ablation_expected_ell.tikz
\begin{tikzpicture}[scale=0.80]
\begin{axis}[
    width=1.22\columnwidth,
    height=3.8cm,
    ybar,
    enlarge x limits=0.22,
    symbolic x coords={Rand,Cave,4Moons,MGate},
    xtick=data,
    xtick style={draw=none}, 
    xticklabel style={font=\ttfamily}, 
    ylabel={$\ell_{\Gamma}$~[\si{\metre}]},
    tick align=inside,
    tick style={thin},
    legend style={
        at={(0.02,0.96)},
        anchor=north west,
        legend columns=1,
        font=\footnotesize\color{gray},
        draw=black!25,
        fill=white,
        fill opacity=0.55,
        text opacity=1,
        inner sep=1.5pt,
    },
    legend cell align=left,
    ymin=0,
    every axis plot/.append style={
        /pgf/bar width=7.5pt,
        error bars/y dir=both,
        error bars/y explicit,
    },
]
\addplot+[draw=green!50!black, fill=green!35!black!10] coordinates {
    (Rand,60.62) +- (0,1.23)
    (Cave,118.60) +- (0,1.80)
    (4Moons,111.69) +- (0,7.57)
    (MGate,264.68) +- (0,5.71)
};
\addlegendentry{Default}
\addplot+[draw=orange!80!black, fill=orange!35] coordinates {
    (Rand,74.64) +- (0,12.87)
    (Cave,122.70) +- (0,1.25)
    (4Moons,170.48) +- (0,54.61)
    (MGate,269.77) +- (0,8.43)
};
\addlegendentry{\textit{Test~\#2} ($\mathrm{moment}$)}
\addplot+[draw=blue!70!black, fill=blue!25] coordinates {
    (Rand,76.38) +- (0,20.38)
    (Cave,123.21) +- (0,3.00)
    (4Moons,193.61) +- (0,55.35)
    (MGate,265.54) +- (0,11.67)
};
\addlegendentry{\textit{Test~\#4} ($w_c{=}1$)}
\end{axis}
\end{tikzpicture}

%% file: figures/ablation_expected_tplan.tikz
\begin{tikzpicture}[scale=0.80]
\begin{axis}[
    width=1.22\columnwidth,
    height=3.8cm,
    ybar,
    enlarge x limits=0.22,
    symbolic x coords={Rand,Cave,4Moons,MGate},
    xtick=data,
    xtick style={draw=none}, 
    xticklabel style={font=\ttfamily}, 
    ylabel={$T_{\mathrm{plan}}$~[\si{\second}]},
    tick align=inside,
    tick style={thin},
    legend style={
        at={(0.02,0.96)},
        anchor=north west,
        legend columns=1,
        font=\footnotesize\color{gray},
        draw=black!25,
        fill=white,
        fill opacity=0.55,
        text opacity=1,
        inner sep=1.5pt,
    },
    legend cell align=left,
    ymin=0,
    every axis plot/.append style={
        /pgf/bar width=7.5pt,
        error bars/y dir=both,
        error bars/y explicit,
    },
]
\addplot+[draw=green!50!black, fill=green!35!black!10] coordinates {
    (Rand,4.12) +- (0,1.77)
    (Cave,5.48) +- (0,3.27)
    (4Moons,3.95) +- (0,1.02)
    (MGate,19.98) +- (0,6.44)
};
\addlegendentry{Default}
\addplot+[draw=orange!80!black, fill=orange!35] coordinates {
    (Rand,5.57) +- (0,3.89)
    (Cave,12.68) +- (0,6.64)
    (4Moons,21.60) +- (0,10.35)
    (MGate,34.28) +- (0,13.07)
};
\addlegendentry{\textit{Test~\#3} (Sequential)}
\addplot+[draw=blue!70!black, fill=blue!25] coordinates {
    (Rand,0.95) +- (0,0.02)
    (Cave,1.42) +- (0,0.22)
    (4Moons,1.18) +- (0,0.24)
    (MGate,2.73) +- (0,0.64)
};
\addlegendentry{\textit{Test~\#5} ($k_b{=}1$)}
\end{axis}
\end{tikzpicture}

%% file: figures/ablation_expected_angacc.tikz
\begin{tikzpicture}[scale=0.80]
\begin{axis}[
    width=1.22\columnwidth,
    height=3.8cm,
    ybar,
    enlarge x limits=0.22,
    symbolic x coords={Rand,Cave,4Moons,MGate},
    xtick=data,
    xtick style={draw=none}, 
    xticklabel style={font=\ttfamily}, 
    ylabel={$a_{\omega,\mathrm{rms}}$~[\si{\radian/\second^2}]},
    ylabel shift={-2pt},
    tick align=inside,
    tick style={thin},
    legend style={
        at={(0.98,0.96)},
        anchor=north east,
        legend columns=1,
        font=\footnotesize\color{gray},
        draw=black!25,
        fill=white,
        fill opacity=0.55,
        text opacity=1,
        inner sep=1.5pt,
    },
    legend cell align=left,
    ymin=0,
    every axis plot/.append style={
        /pgf/bar width=7.5pt,
        error bars/y dir=both,
        error bars/y explicit,
    },
]
\addplot+[draw=green!50!black, fill=green!35!black!10] coordinates {
    (Rand,0.25) +- (0,0.06)
    (Cave,0.27) +- (0,0.03)
    (4Moons,0.29) +- (0,0.01)
    (MGate,0.28) +- (0,0.03)
};
\addlegendentry{Default}
\addplot+[draw=orange!80!black, fill=orange!35] coordinates {
    (Rand,0.23) +- (0,0.06)
    (Cave,0.21) +- (0,0.02)
    (4Moons,0.23) +- (0,0.02)
    (MGate,0.17) +- (0,0.01)
};
\addlegendentry{\textit{Test~\#3} (Sequential)}
\end{axis}
\end{tikzpicture}

%% file: figures/ablation_sequential_cave/cave_default_r1_omega.tikz
\begin{tikzpicture}[scale=0.8]
\begin{axis}[
    width=4cm,
    height=2.1cm,
    scale only axis,
    xlabel={Time~[\si{\second}]},
    ylabel={$\omega$~[\si{\radian/\second}]},
    ylabel style={yshift=-3pt},
    tick align=inside,
    tick style={thin},
    enlarge x limits=false,
    ymin=-0.7, ymax=0.5,
    ytick={-0.6,-0.3,0.0,0.3},
]
\addplot+[thick, mark=none, solid, color=blue!70!black] coordinates {
    (0.00,0.0176)
    (1.10,0.0263)
    (2.20,0.0408)
    (3.30,0.0430)
    (4.40,0.0513)
    (5.50,0.0651)
    (6.60,0.0819)
    (7.70,0.0835)
    (8.80,0.0779)
    (9.90,0.0671)
    (11.00,0.0580)
    (12.10,0.1660)
    (13.20,0.1301)
    (14.30,0.0808)
    (15.40,0.0487)
    (16.50,0.0440)
    (17.60,0.0510)
    (18.70,0.0613)
    (19.80,0.0680)
    (20.90,0.0619)
    (22.00,0.0509)
    (23.10,0.0433)
    (24.20,-0.6364)
    (25.30,-0.5244)
    (26.40,-0.3291)
    (27.50,-0.1385)
    (28.60,-0.0290)
    (29.70,-0.0073)
    (30.80,-0.0591)
    (31.90,-0.0819)
    (33.00,-0.0263)
    (34.10,0.0976)
    (35.20,0.1942)
    (36.30,-0.0081)
    (37.40,0.0027)
    (38.50,0.0197)
    (39.60,0.0340)
    (40.70,0.0432)
    (41.80,0.0429)
    (42.90,0.0457)
    (44.00,0.0362)
    (45.10,0.0153)
    (46.20,-0.0038)
    (47.30,-0.0126)
    (48.40,-0.5270)
    (49.50,-0.3451)
    (50.60,-0.1249)
    (51.70,-0.0009)
    (52.80,0.0303)
    (53.90,0.0263)
    (55.00,0.0287)
    (56.10,0.0555)
    (57.20,0.0876)
    (58.30,0.1002)
    (59.40,0.0932)
    (60.50,0.2973)
    (61.60,0.1786)
    (62.70,0.0627)
    (63.80,0.0316)
    (64.90,0.0519)
    (66.00,0.0538)
    (67.10,0.0232)
    (68.20,-0.0067)
    (69.30,-0.0188)
    (70.40,-0.0344)
    (71.50,-0.0448)
    (72.60,-0.1612)
    (73.70,-0.0837)
    (74.80,-0.0070)
    (75.90,0.0343)
    (77.00,0.0585)
    (78.10,0.0764)
    (79.20,0.0831)
    (80.30,0.0771)
    (81.40,0.0616)
    (82.50,0.0509)
    (83.60,0.0461)
    (84.70,0.1389)
    (85.80,0.0881)
    (86.90,0.0469)
    (88.00,0.0332)
    (89.10,0.0259)
    (90.20,0.0097)
    (91.30,-0.0155)
    (92.40,-0.0231)
    (93.50,-0.0179)
    (94.60,-0.0023)
    (95.70,-0.0003)
    (96.80,-0.6338)
    (97.90,-0.3450)
    (99.00,-0.0474)
    (100.10,0.1153)
    (101.20,0.1251)
    (102.30,0.0520)
    (103.40,-0.0125)
    (104.50,-0.0287)
    (105.60,-0.0037)
    (106.70,0.0259)
    (107.80,0.0391)
    (108.90,0.1408)
    (110.00,0.0713)
    (111.10,0.0180)
    (112.20,-0.0106)
    (113.30,-0.0102)
    (114.40,-0.0145)
    (115.50,-0.0195)
    (116.60,-0.0298)
    (117.70,-0.0384)
    (118.80,-0.0444)
    (119.90,0.0000)
    (121.00,-0.0200)
    (122.10,0.0130)
    (123.20,0.0423)
    (124.30,0.0656)
    (125.40,0.0827)
    (126.50,0.0929)
    (127.60,0.0954)
    (128.70,0.0887)
    (129.80,0.0631)
    (130.90,0.0318)
    (132.00,0.3647)
    (133.10,0.2840)
    (134.20,0.1533)
    (135.30,0.0731)
    (136.40,0.0594)
    (137.50,0.0595)
    (138.60,0.0608)
    (139.70,0.0574)
    (140.80,0.0455)
    (141.90,0.0193)
    (143.00,-0.0013)
    (144.10,-0.2016)
    (145.20,-0.1391)
    (146.30,-0.0454)
    (147.40,0.0220)
    (148.50,0.0417)
    (149.60,0.0490)
    (150.70,0.0494)
    (151.80,0.0577)
    (152.90,0.0642)
    (154.00,0.0644)
    (155.10,0.0540)
    (156.20,0.2879)
    (157.30,0.2029)
    (158.40,0.0956)
    (159.50,0.0413)
    (160.60,0.0340)
    (161.70,0.0368)
    (162.80,0.0500)
    (163.90,0.0550)
    (165.00,0.0629)
    (166.10,0.0586)
    (167.20,0.0592)
    (168.30,0.1551)
    (169.40,0.0912)
    (170.50,0.0131)
    (171.60,-0.0238)
    (172.70,-0.0260)
    (173.80,-0.0202)
    (174.90,-0.0251)
    (176.00,-0.0225)
    (177.10,-0.0110)
    (178.20,0.0056)
    (179.30,0.0140)
    (180.40,0.1252)
    (181.50,0.0695)
    (182.60,0.0104)
    (183.70,-0.0131)
    (184.80,-0.0137)
    (185.90,-0.0177)
    (187.00,-0.0221)
    (188.10,-0.0128)
    (189.20,0.0051)
    (190.30,0.0181)
    (191.40,0.0251)
    (192.50,-0.1968)
    (193.60,-0.1188)
    (194.70,-0.0410)
    (195.80,-0.0025)
    (196.90,0.0064)
    (198.00,0.0004)
    (199.10,0.0007)
    (200.20,-0.0002)
    (201.30,0.0023)
    (202.40,0.0009)
    (203.50,0.0060)
    (204.60,-0.1718)
    (205.70,-0.1011)
    (206.80,-0.0345)
    (207.90,-0.0021)
    (209.00,0.0060)
    (210.10,0.0085)
    (211.20,0.0137)
    (212.30,0.0103)
    (213.40,0.0065)
    (214.50,-0.0015)
    (215.60,-0.0017)
    (216.70,0.3166)
    (217.80,0.1869)
    (218.90,0.0739)
    (220.00,0.0217)
    (221.10,0.0057)
    (222.20,-0.0094)
    (223.30,-0.0257)
    (224.40,-0.0395)
    (225.50,-0.0409)
    (226.60,-0.0357)
    (227.70,-0.0325)
    (228.80,-0.3970)
    (229.90,-0.2612)
    (231.00,-0.1068)
    (232.10,0.0191)
    (233.20,0.0981)
    (234.30,0.1162)
    (235.40,0.0961)
    (236.50,0.0485)
    (237.60,-0.0166)
    (238.70,-0.0889)
    (239.80,-0.1227)
    (240.90,0.3325)
    (242.00,0.1911)
    (243.10,0.0853)
    (244.20,0.0450)
    (245.30,0.0372)
    (246.40,0.0345)
    (247.50,0.0358)
    (248.60,0.0245)
    (249.70,0.0058)
    (250.80,-0.0163)
    (251.90,0.0000)
    (253.00,-0.0034)
    (254.10,0.0002)
    (255.20,0.0113)
    (256.30,0.0225)
    (257.40,0.0287)
    (258.50,0.0404)
    (259.60,0.0460)
    (260.70,0.0478)
    (261.80,0.0516)
    (262.90,0.0583)
    (264.00,-0.0135)
    (265.10,-0.0074)
    (266.20,-0.0067)
    (267.30,-0.0063)
    (268.40,-0.0117)
    (269.50,-0.0141)
    (270.60,-0.0070)
    (271.70,-0.0007)
    (272.80,0.0061)
    (273.90,0.0055)
    (275.00,0.0066)
};
\end{axis}
\end{tikzpicture}

%% file: figures/ablation_sequential_cave/cave_sequential_r4_omega.tikz
\begin{tikzpicture}[scale=0.80]
\begin{axis}[
    width=4cm,
    height=2.1cm,
    scale only axis,
    xlabel={Time~[\si{\second}]},
    ylabel={$\omega$~[\si{\radian/\second}]},
    ylabel style={yshift=-3pt},
    tick align=inside,
    tick style={thin},
    enlarge x limits=false,
    ymin=-0.7, ymax=0.5,
    ytick={-0.6,-0.3,0.0,0.3},
]
\addplot+[thick, mark=none, solid, color=blue!70!black] coordinates {
    (0.00,0.1842)
    (1.00,0.1609)
    (2.00,0.1210)
    (3.00,0.0923)
    (4.00,0.0846)
    (5.00,0.0795)
    (6.00,0.0731)
    (7.00,0.0679)
    (8.00,0.0660)
    (9.00,0.0595)
    (10.00,0.0560)
    (11.00,0.0631)
    (12.00,-0.1843)
    (13.00,-0.1818)
    (14.00,-0.1670)
    (15.00,-0.1409)
    (16.00,-0.0971)
    (17.00,-0.0553)
    (18.00,-0.0307)
    (19.00,-0.0093)
    (20.00,0.0018)
    (21.00,0.0101)
    (22.00,0.0051)
    (23.00,-0.0065)
    (24.00,-0.2266)
    (25.00,-0.1881)
    (26.00,-0.1052)
    (27.00,-0.0256)
    (28.00,0.0155)
    (29.00,0.0229)
    (30.00,0.0151)
    (31.00,0.0165)
    (32.00,0.0225)
    (33.00,0.0114)
    (34.00,-0.0147)
    (35.00,-0.0414)
    (36.00,0.1377)
    (37.00,0.1169)
    (38.00,0.0815)
    (39.00,0.0593)
    (40.00,0.0491)
    (41.00,0.0370)
    (42.00,0.0285)
    (43.00,0.0312)
    (44.00,0.0394)
    (45.00,0.0388)
    (46.00,0.0330)
    (47.00,0.0256)
    (48.00,-0.4852)
    (49.00,-0.4267)
    (50.00,-0.3130)
    (51.00,-0.2055)
    (52.00,-0.1405)
    (53.00,-0.0717)
    (54.00,0.0413)
    (55.00,0.1180)
    (56.00,0.1038)
    (57.00,0.0565)
    (58.00,0.0218)
    (59.00,0.0209)
    (60.00,0.2631)
    (61.00,0.2324)
    (62.00,0.1834)
    (63.00,0.1452)
    (64.00,0.1189)
    (65.00,0.0957)
    (66.00,0.0690)
    (67.00,0.0555)
    (68.00,0.0373)
    (69.00,0.0084)
    (70.00,-0.0381)
    (71.00,-0.0730)
    (72.00,0.0034)
    (73.00,0.0149)
    (74.00,0.0232)
    (75.00,0.0325)
    (76.00,0.0321)
    (77.00,0.0441)
    (78.00,0.0462)
    (79.00,0.0484)
    (80.00,0.0453)
    (81.00,0.0264)
    (82.00,-0.0044)
    (83.00,-0.0403)
    (84.00,-0.1673)
    (85.00,-0.1465)
    (86.00,-0.1040)
    (87.00,-0.0582)
    (88.00,-0.0226)
    (89.00,-0.0052)
    (90.00,0.0024)
    (91.00,-0.0026)
    (92.00,-0.0107)
    (93.00,-0.0110)
    (94.00,0.0069)
    (95.00,0.0298)
    (96.00,-0.1766)
    (97.00,-0.1437)
    (98.00,-0.0917)
    (99.00,-0.0424)
    (100.00,-0.0059)
    (101.00,0.0208)
    (102.00,0.0366)
    (103.00,0.0301)
    (104.00,0.0247)
    (105.00,0.0189)
    (106.00,0.0109)
    (107.00,0.0023)
    (108.00,-0.1466)
    (109.00,-0.1135)
    (110.00,-0.0458)
    (111.00,0.0093)
    (112.00,0.0412)
    (113.00,0.0537)
    (114.00,0.0543)
    (115.00,0.0480)
    (116.00,0.0399)
    (117.00,0.0402)
    (118.00,0.0363)
    (119.00,0.0340)
    (120.00,0.3242)
    (121.00,0.2752)
    (122.00,0.1932)
    (123.00,0.1215)
    (124.00,0.0840)
    (125.00,0.0683)
    (126.00,0.0515)
    (127.00,0.0467)
    (128.00,0.0401)
    (129.00,0.0392)
    (130.00,0.0295)
    (131.00,0.0228)
    (132.00,0.1474)
    (133.00,0.1306)
    (134.00,0.1046)
    (135.00,0.0841)
    (136.00,0.0701)
    (137.00,0.0549)
    (138.00,0.0407)
    (139.00,0.0391)
    (140.00,0.0497)
    (141.00,0.0520)
    (142.00,0.0436)
    (143.00,0.0273)
    (144.00,-0.1405)
    (145.00,-0.1221)
    (146.00,-0.0922)
    (147.00,-0.0749)
    (148.00,-0.0567)
    (149.00,-0.0222)
    (150.00,0.0110)
    (151.00,0.0470)
    (152.00,0.0653)
    (153.00,0.0667)
    (154.00,0.0511)
    (155.00,0.0326)
    (156.00,0.3152)
    (157.00,0.2783)
    (158.00,0.1967)
    (159.00,0.1330)
    (160.00,0.0853)
    (161.00,0.0603)
    (162.00,0.0347)
    (163.00,0.0156)
    (164.00,0.0125)
    (165.00,0.0138)
    (166.00,0.0149)
    (167.00,0.0115)
    (168.00,-0.0008)
    (169.00,0.0006)
    (170.00,0.0014)
    (171.00,-0.0016)
    (172.00,-0.0007)
    (173.00,0.0093)
    (174.00,0.0138)
    (175.00,0.0221)
    (176.00,0.0243)
    (177.00,0.0285)
    (178.00,0.0296)
    (179.00,0.0342)
    (180.00,-0.0016)
    (181.00,0.0011)
    (182.00,0.0027)
    (183.00,0.0008)
    (184.00,-0.0050)
    (185.00,-0.0108)
    (186.00,-0.0115)
    (187.00,-0.0065)
    (188.00,-0.0039)
    (189.00,0.0008)
    (190.00,-0.0022)
    (191.00,-0.0076)
    (192.00,0.0021)
    (193.00,-0.0011)
    (194.00,-0.0017)
    (195.00,-0.0034)
    (196.00,-0.0042)
    (197.00,-0.0069)
    (198.00,-0.0082)
    (199.00,-0.0071)
    (200.00,-0.0063)
    (201.00,-0.0104)
    (202.00,-0.0108)
    (203.00,-0.0061)
    (204.00,-0.1540)
    (205.00,-0.1438)
    (206.00,-0.1191)
    (207.00,-0.0996)
    (208.00,-0.0835)
    (209.00,-0.0716)
    (210.00,-0.0590)
    (211.00,-0.0453)
    (212.00,-0.0336)
    (213.00,-0.0268)
    (214.00,-0.0267)
    (215.00,-0.0364)
    (216.00,0.0395)
    (217.00,0.0328)
    (218.00,0.0307)
    (219.00,0.0187)
    (220.00,0.0206)
    (221.00,0.0264)
    (222.00,0.0341)
    (223.00,0.0378)
    (224.00,0.0317)
    (225.00,0.0158)
    (226.00,0.0003)
    (227.00,-0.0100)
    (228.00,0.1765)
    (229.00,0.1550)
    (230.00,0.1076)
    (231.00,0.0719)
    (232.00,0.0542)
    (233.00,0.0453)
    (234.00,0.0385)
    (235.00,0.0308)
    (236.00,0.0240)
    (237.00,0.0092)
    (238.00,-0.0051)
    (239.00,-0.0177)
    (240.00,0.1624)
    (241.00,0.1408)
    (242.00,0.0976)
    (243.00,0.0623)
    (244.00,0.0398)
    (245.00,0.0325)
    (246.00,0.0289)
    (247.00,0.0246)
    (248.00,0.0185)
    (249.00,0.0227)
    (250.00,0.0252)
    (251.00,0.0247)
    (252.00,-0.0031)
    (253.00,-0.0037)
    (254.00,-0.0057)
    (255.00,-0.0119)
    (256.00,-0.0164)
    (257.00,-0.0208)
    (258.00,-0.0197)
    (259.00,-0.0110)
    (260.00,-0.0002)
    (261.00,0.0028)
    (262.00,-0.0002)
    (263.00,-0.0006)
};
\end{axis}
\end{tikzpicture}